\documentclass{article}

\usepackage[preprint]{neurips_2026}
\usepackage[caption=false]{subfig}
\usepackage{graphicx}
\usepackage{placeins}
\usepackage{amsmath}
\usepackage{amssymb}
\usepackage{amsfonts}
\usepackage{pgffor}

\usepackage[utf8]{inputenc} 
\usepackage[T1]{fontenc}    
\usepackage{hyperref}       
\usepackage{url}            
\usepackage{booktabs}       
\usepackage{amsfonts}       
\usepackage{nicefrac}       
\usepackage{microtype}      
\usepackage{xcolor}         
\usepackage{booktabs}
\usepackage{enumitem}

\title{Lost or Hidden? A Concept-Level Forgetting in Supervised Continual Learning}

\author{%
  Katarzyna Filus \\
  Institute of Theoretical and Applied Informatics\\
  Polish Academy of Sciences, Gliwice, Poland\\
  \texttt{kfilus@iitis.pl} \\
  \And
  Kamil Faber \\
  AGH University of Krakow \\
  Krakow, Poland \\
  \texttt{kfaber@agh.edu.pl} \\
  \AND
  Roberto Corizzo \\
  American University \\
  Washington DC, USA \\
  \texttt{rcorizzo@american.edu} \\
  \And
  Christopher Kanan \\
  University of Rochester \\
  Rochester,  New York, USA \\
  \texttt{ckanan@cs.rochester.edu} \\
}

\begin{document}

\maketitle

\begin{abstract}
Continual learning studies how models can adapt to new tasks while retaining previously acquired knowledge. Although a broad spectrum of methods has been proposed to mitigate catastrophic forgetting, the field remains predominantly performance-driven, with limited insight into what forgetting actually corresponds to within the vision model's representation space. Prior work has primarily analyzed forgetting through task-level performance or coarse measures of representational drift, without disentangling output-level accessibility from changes in finer-grained internal structure. To this end, we propose a diagnostic framework that leverages Sparse Autoencoders (SAEs) to define a task-anchored latent feature space, enabling analysis of how task-specific information evolves at a finer granularity, where individual SAE latents are treated as concept proxies for recurring and relatively disentangled visual patterns in the model’s internal computations. Within this framework, we decompose forgetting into apparent concept deletion, recoverability, and decodability. We show that a large portion of seemingly lost concept-level information can often be recovered under linearity assumption, with concept decodability degrading as more tasks are introduced. Overall, our findings suggest that a significant part of concept-level forgetting can be attributed to changes in the representational accessibility rather than complete information erasure.
\end{abstract}

\section{Introduction}

The degradation of previously learned knowledge during sequential training, known as catastrophic forgetting, remains a central challenge in continual learning \cite{wang2024comprehensive}. Various forgetting mitigation strategies have been proposed, including rehearsal-based methods \cite{rolnick2019experience}, regularization and distillation \cite{li2017lwf}, as well as hybrid methods \cite{buzzega2020der}. 
Despite this progress, evaluation remains largely performance-driven, equating forgetting to decrease in task accuracy after learning new tasks, while leaving unresolved the fundamental questions of how forgetting manifests within models~\cite{masip2026putting}. 
Recent works notice this gap and attempt to shift from the output-level evaluation to representations \cite{hu2025continual,masip2026putting,davari2022probing,gu2023preserving}. Previous studies measure representational drift using mostly similarity measures \cite{kornblith2019similarity,hu2025continual,masip2026putting}. Other works \cite{davari2022probing,gu2023preserving} show that much apparent performance loss can be recovered via retrained classifiers or representation constraints.
However, these analyses remain global and tied to task-level performance, and thus do not capture how fine-grained information is organized, transformed, or lost.

\begin{figure*}[t]
    \centering
                \includegraphics[width=\textwidth]{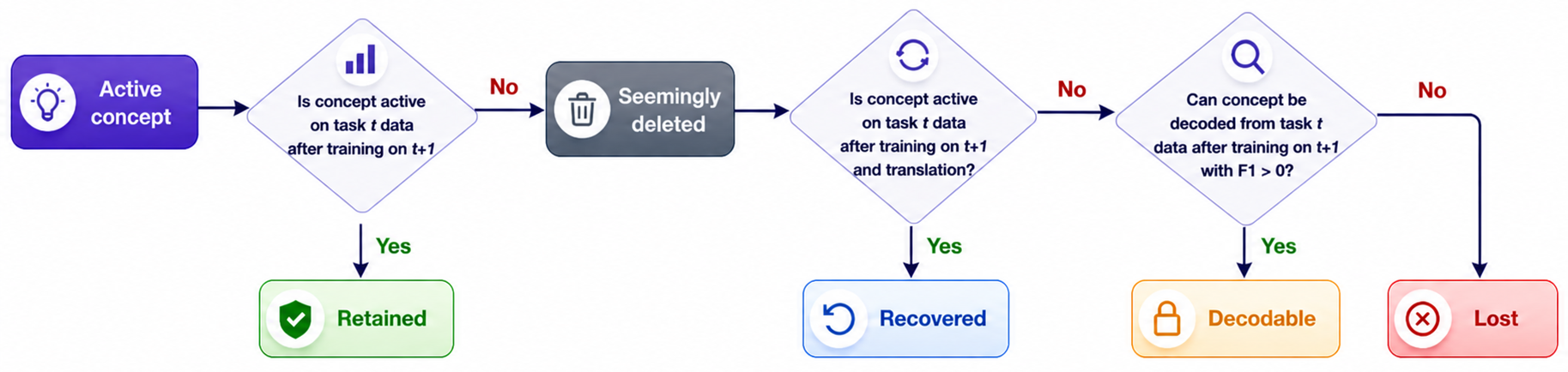}
    \caption{The taxonomy of concept transitions for task $t$ data after task $t+1$ training.}
    \label{fig:conceps_taxonomy}
\end{figure*}

Similarly to how humans recognize objects by combining recurring concepts, recent progress in mechanistic interpretability enables the study of neural representations in terms of more disentangled and interpretable features used to perform learned tasks \cite{bereska2024mechanistic}. It is particularly important because deep networks are considered to rely on superposition, where multiple concepts -- understood as recurring patterns within the model computation -- are encoded in overlapping neural directions, making feature-level analysis difficult \cite{bereska2024mechanistic}. Methods such as Sparse Autoencoders (SAEs) \cite{huben2023sparse} and transcoders \cite{dunefsky2024transcoders} map representations into higher-dimensional sparse latent spaces that can isolate more atomic and interpretable features. 
Although these latent dimensions do not correspond to the ground-truth semantic concepts, they often capture recurring and relatively disentangled patterns in model computations, serving as useful \emph{concept proxies} that often align with abstract motifs such as furry animals or red objects~\cite{bereska2024mechanistic}. In the following, we refer to them as \emph{concepts} and leverage them as a structured basis for analyzing how fine-grained information evolves during continual learning.

Studying forgetting at a fine-grained level should be as important as in terms of accuracy degradation, indicating whether the model retains knowledge and supports stable interpretations rather than  shifts representations. However, despite the potential of mechanistic interpretability, it has been scarcely applied to this problem. The only work \cite{masip2026putting} uses transcoders to analyze forgetting as geometric transformations of features in a shared latent space. 
However, it focuses on \emph{how} features transform, rather than analyzing task-specific information accessibility. In particular, it does not quantify concept-level activation dynamics, the decodability of individual features, or the extent to which seemingly lost information can be recovered under standard readout assumptions. As a result, the relationship between representational change and functional forgetting remains insufficiently understood.

In this work, we address this gap by proposing a concept-level framework for analyzing forgetting through the accessibility of task-specific information. Using Sparse Autoencoders (SAEs), we define a fixed, task-anchored latent space in which individual features act as  concept proxies, enabling fine-grained tracking of how information evolves during continual training. Within this framework, we decompose forgetting into concept deletion, recoverability, and decodability, distinguishing between information that is removed, misaligned, or still present but not directly accessible. Our results show that much of the seemingly lost concept-level information remains linearly recoverable, indicating that forgetting often reflects reduced accessibility to the underlying computational patterns rather than true erasure. At the same time, some concepts become less linearly decodable over time, making loss of accessibility functionally equivalent to forgetting under standard linear readouts. 
Moreover, continual strategies differ in how well they preserve the fine-grained information. Our findings provide a unified view of forgetting that links representational drift, readout limitations, and concept-level information dynamics.
Our contributions can be summarized as follows:
\begin{itemize}[topsep=0pt, itemsep=1pt, parsep=0pt, partopsep=0pt]
    \item We introduce a \emph{concept-level} framework for analyzing forgetting in continual learning, leveraging SAEs to define a task-anchored latent space whose features act as concept proxies. It enables studying task-specific knowledge preservation and linear recoverability through concept activation dynamics and task- and concept-level decodability.
    
    \item We define a taxonomy of concept transitions after learning subsequent tasks in continual learning: retained, seemingly deleted, recovered, decodable, and lost concepts (see Fig. \ref{fig:conceps_taxonomy}).
    
    \item We show that a substantial portion of task-specific information remains linearly recoverable at the level of individual concepts, and that continual learning strategies differ in how well they preserve the fine-grained knowledge and its recoverability.
    
\end{itemize}



\section{Related work}
Most work in continual learning  quantifies forgetting through degradation in the model’s classifier performance \cite{verwimp2024continual,soutif2023comprehensive,kemker2018measuring, kirkpatrick2017overcoming}. Most works focused on developing mitigation strategies, including replay-based methods, regularization and distillation, architectural, and hybrid variants~\cite{wang2024comprehensive,parisi2019continual,202601.1931}, with forgetting typically evaluated through accuracy-based metrics. Although this line of work has established effective methods for reducing catastrophic forgetting, it offers limited insight into how forgetting manifests within internal representations.
More recent studies relate forgetting to representational drift and changes in feature geometry \cite{davari2022probing,gu2023preserving,masip2026putting,hu2025continual}. The authors of \cite{davari2022probing} shows that classifier-level forgetting can overestimate degradation of the underlying representations. In contrast, we move beyond task-level probing on raw features by using an SAE-defined, task-anchored concept space to analyze whether more fine-grained task-relevant information remains preserved and linearly decodable at the concept level. 
The authors of \cite{gu2023preserving} enforces linear separability across tasks through backward feature projection. In contrast, we use linear mappings not as a training constraint, but as a diagnostic tool to measure how much past information remains linearly recoverable under standard continual learning strategies, and where this recoverability breaks down.
In terms of understanding forgetting, mechanistic interpretability offers a complementary perspective by decomposing neural representations into more atomic and interpretable components, e.g., by using Sparse Autoencoders (SAEs) and transcoders \cite{huben2023sparse,dunefsky2024transcoders,bereska2024mechanistic}.
Such methods construct overcomplete, sparse latent spaces that expose recurring disentangled patterns, offering a stronger basis for feature-level analysis. Yet, mechanistic interpretability remains rarely used to study forgetting in continual learning.
The closest work to ours is~\cite{masip2026putting}, in which  transcoders are used to map representations from different tasks into a shared space and describes forgetting in terms of geometric transformations such as rotations and scaling.
In contrast, our goal is not to characterize how features transform but to study and quantify whether task-specific information remains accessible at the concept level. To this end, we use task-anchored SAE latent spaces and analyze concept activation dynamics, linear recoverability, and concept-level decodability. This allows us to distinguish between information that is preserved but misaligned, implicitly encoded, or no longer accessible under linear readouts.


\section{Forgetting as Loss of Accessibility: A Concept-Level Framework}
\label{sec:methodology}




In this section, we formulate our concept-level analysis of forgetting as a loss of accessibility to fine-grained task-specific information rather than complete information loss. The central question is whether information that no longer supports the original task readout after continual training has been erased, or it remains present in the representation but is misaligned, less active, or harder to decode. Our framework is diagnostic, analyzing how past task information evolves within the model.

\paragraph{Problem setup.}
Let $\mathcal{D}^t = \{(x_i^t, y_i^t)\}_{i=1}^{N_t}$ denote the dataset of task $t$. Let $f_{\theta_t}$ denote the continual model's feature extractor after training on task $t$, and $f_{\theta_{t+s}}$ the same extractor after further training on tasks $t+1, \dots, t+s$. We study representations of task $t$ data under these two models:
\begin{equation}
    h_t(x) = f_{\theta_t}(x), \quad h_{t+s}(x) = f_{\theta_{t+s}}(x), \quad x \in \mathcal{D}^t.
\end{equation}
The goal is to analyze how the representation of task $t$ evolves after subsequent training and to what extent task-relevant information remains accessible.

\paragraph{Linear translation of representations.}
Continual training may introduce geometric shifts in the representation space without necessarily removing the information. To study this behavior, we introduce a linear translation mapping:
\begin{equation}
    T_{t+s \rightarrow t}(h) = W_{t+s} h + b_{t+s},
\end{equation}
trained to align $h_{t+s}(x)$ with $h_t(x)$ using task $t$ training data:
\begin{equation}
    \min_{W,b} \ \mathbb{E}_{x \sim \mathcal{D}^t_{\text{train}}} \left\| W h_{t+s}(x) + b - h_t(x) \right\|_2^2.
\end{equation}
The learned mapping $T_{t+s \rightarrow t}$ is then applied to $t$ test data. We use linear mappings to match standard readouts in continual learning, enabling recovery analysis under typical decision heads. This choice is supported by prior work showing that forgetting can be mitigated by retraining linear heads \cite{davari2022probing}, that task representations may remain linearly related \cite{gu2023preserving}, and aligns with the linear representation hypothesis \cite{elhage2022toy,bereska2024mechanistic,masip2026putting}. If task-$t$ information can be recovered via a linear mapping, forgetting may reflect representational change rather than complete erasure.

\paragraph{Concept proxies via sparse autoencoders.}
To analyze forgetting with finer granularity than task accuracy or global representation similarity, for each task $t$, we train an SAE on the train split of $h_t(x)$, and use the trained SAE's encoder to obtain different test representations:
\begin{equation}
    z_t(x) = \text{SAE}_t(h_t(x)), \quad z_{t+s}(x) = \text{SAE}_t(h_{t+s}(x)), \quad z_{T}(x) = \text{SAE}_t(T_{t+s \rightarrow t}(h_{t+s}(x))),
\end{equation}
where $z(x) \in \mathbb{R}^K$ denotes latent activations. The SAE trained on task $t$ defines a fixed coordinate system -- an anchor for concept-level analysis to track how specific representational components evolve across training. 
Individual task-anchored latent features serve as \textbf{concept proxies}. We refer to them as \textbf{concepts}: not ground-truth semantic entities, but more disentangled, fine-grained, recurring features in model computation (see Appendix \ref{app:concept} for a detailed definition), in contrast to the highly entangled features of standard continual models that hinder feature-level analysis.

\paragraph{Identifying active concepts through binarization of activations.}
Following recent work showing that binarization of SAE features can provide an effective alternative representation \cite{gallifant2025sparse,aswal2025llmsymguard}, and to focus on salient and reusable features, we introduce a frequency-based binarization rule to assess whether the concept is \textbf{active}. For each latent dimension $k$, we compute its mean activation over task $t$ training data $\mu_k$ and use it to define a mean-based binarization rule \cite{aswal2025llmsymguard}:
\begin{equation}
    a_k(x) = \mathbb{I}[z^k(x) > \mu_k].
\end{equation}
Then, a concept $k$ is considered \textbf{active} for task $t$ and a frequency threshold $\tau$ if
\begin{equation}
    \frac{1}{|\mathcal{D}^t_{\text{train}}|} \sum_{x} a_k(x) \geq \tau,
\end{equation}
as consistently and strongly activated, corresponding to a likely informative and reusable feature. Therefore, we consider: raw features $h(x)$, continuous $z(x)$ and binarized $a(x)$ latent activations.


\paragraph{Analysis modules.}
\hfill \\
\textbf{(1) Binary concept activation analysis.}
We analyze the structure of the binary activation space by measuring:
i) the number of \emph{active concepts},
ii) the \emph{deletion ratio} (fraction of concepts active at $t$ but inactive at $t+s$; by subtracting this value from 1, we obtain the ratio of retained concepts),
iii) \emph{regained concept count} after applying $T_{t+s \rightarrow t}$ and \emph{regained activation mass} (the non-negative recovery of latent activation magnitude in deleted features provided by translation $T$ relative to $z_{t+s}(x)$, normalized by the total activation loss incurred between $z_t(x)$ and $z_{t+s}(x)$).
These metrics quantify how concept-level information is preserved, potentially lost, or recoverable. \\
\textbf{(2) Concept prediction from raw features.}
We assess concept-level decodability by predicting binary concept activations from raw features. For each concept $k$, we define a binary label:
\begin{equation}
    y_k(x) = \mathbb{I}[z_t^k(x) > 0],
\end{equation}
and train a classifier to predict $y_k(x)$ from $h_{t+s}(x)$. We use a less strict threshold than in the activity analysis, since the selected latent features have already been identified as salient concept proxies at the dataset level and to reduce sparse space class imbalance. We evaluate performance using balanced accuracy and F1 score on the $t+s$ test splits. High values indicate that concept-level information remains decodable from the representation. Mean scores reflect overall concept preservation, while their distribution reflects that concepts can have different decodability levels. \\
\textbf{(3) Task prediction via linear probes.}
We train linear classifiers on task $t$ data after task $t$ training to predict task labels using:
(i) raw features $h_t(x)$,
(ii) concept activations $z_t(x)$.
We evaluate these trained probes on test $t$ data at $t$, $t+s$, and $T_{t+s \rightarrow t}$ using accuracy. This measures how well task-level information remains linearly decodable when using raw features and concept-level features, and whether translation can fix some of the linear readout issues.


\paragraph{The concept taxonomy.} We distinguish five possible outcomes for a task-$t$ concept proxy after training on task $t+s$: \textbf{\emph{Retained}}, when the concept is active at $t$ and remains active in the raw $t+s$ representation; \textbf{\emph{Seemingly deleted}}, when it was active at $t$ but is absent in raw $t+s$ under the binarization rule; \textbf{\emph{Recovered}}, when it is absent in raw $t+s$ but reappears after translation; \textbf{\emph{Decodable}}, when it does not reappear as an active concept, but its presence can be predicted from the representation ($\mathrm{F1}>0$); and \textbf{\emph{Lost}}, when it is neither recovered by translation nor decodable ($\mathrm{F1}=0$). \textbf{\emph{Decodable}} concepts do not reactivate but remain predictable (non-zero performance), whereas \textbf{\emph{Lost}} concepts are neither recoverable nor decodable. Among decodable concepts, balanced accuracy and F1 quantify recoverability, with higher values indicating greater accessibility (see Fig.~\ref{fig:conceps_taxonomy}).

\section{Experimental setup}\label{sec:setup}

Our experimental design aims to evaluate the accessibility-based formulation of forgetting introduced in Section~\ref{sec:methodology}.  
%
Our \textbf{continual learning setting} includes evaluation of our framework on 3 task-incremental benchmarks: i) \textbf{2seq-CIFAR10}, splitting CIFAR-10 \cite{krizhevsky2009learning} into 2 tasks, 5 classes each; ii) \textbf{2seq-tiny-ImageNet}, using the first 40 classes of tiny-ImageNet \cite{deng2009imagenet}, divided into 2 tasks with 20 classes each, enabling a coarse-grained comparison with 2seq-CIFAR10; iii) \textbf{10seq-tiny-ImageNet}, using a split of tiny-ImageNet into 10 tasks with 20 classes each.
We leverage four representative continual learning strategies: i) \textbf{SGD}, corresponding to naive fine-tuning without knowledge retention, serves as a baseline for natural continual-learning behavior; ii) Learning without Forgetting (\textbf{LWF}) \cite{li2017lwf}, a distillation-based strategy; iii) Elastic Weight Consolidation (\textbf{EWC}), a regularization-based strategy  \cite{kirkpatrick2017overcoming}, and iv) Dark Experience Replay ++ (\textbf{DER++}) \cite{buzzega2020der}, a hybrid strategy combining experience replay with knowledge distillation and regularization. 
As the backbone, we use Resnet18 \cite{he2016deep}, one of the most popular models in continual image classification \cite{bian2024make,masana2022class}.
To ensure reproducibility, we use implementations, data splits and hyperparameters from \textit{Mammoth}\footnote{\url{https://github.com/aimagelab/mammoth}} \cite{boschini2022class}.
Appendix~\ref{app:accuracy_continual} reports Task-Incremental accuracies: all strategies achieve relatively strong performance on new tasks, while exhibiting different forgetting levels, making them suitable for our analysis.


In our \textbf{concept-level analysis setup}, 
for each task, we train an SAE with \texttt{BatchTopKSAE} regime via \texttt{overcomplete} \footnote{\url{https://github.com/KempnerInstitute/overcomplete}} on the same Mammoth's train/test splits. We use Mean Squared Error (MSE) with an \texttt{overcomplete}'s dead-neuron reactivation loss weighted by $10^{-2}$. Across experiments, manual inspection showed reconstruction $R^2 > 0.6$ and close to 0\% dead-neuron rate, indicating meaningful and non-degenerate SAEs. We also examined top 9 activating images for selected SAE latents and found consistent shared visual motifs, indicating coherent patterns rather than arbitrary feature mixtures (App. \ref{app:concept_viz}). App.~\ref{app:ms_analysis} shows that active SAE latents can achieve substantially higher monosemanticity scores (MS) \cite{pach2025sparse} than randomized baselines, quantitatively supporting their quality and coherence as concept proxies.
We use a single linear layer for translation, and Logistic Regression for task-level and concept-level probes.
Though nonlinear mappings are less constrained and relevant under standard readouts, we additionally evaluate a simple nonlinear translator in App.~\ref{app:nonlinear_translation}, which yields no significant gains. We verified the analysis robustness across 10 runs (App.~\ref{app:stability_across_runs}), 6 frequency-based binarization thresholds (App.~\ref{app:stability_binarization}), 5 SAE batch sizes (App.~\ref{app:stability_sae_batch}), and 4 SAE $K$ values (App.~\ref{app:stability_sae_K}). The results show consistent behavior of key metrics
and support our conclusions' robustness. We validate that the selected active neurons retain most task-relevant information through SAE-based deletion experiments on non-active neurons under different parameters and binarization thresholds (App.~\ref{app:stability_binarization}, \ref{app:binarization_at_05}, \ref{app:stability_sae_K}, \ref{app:stability_sae_batch}).
We used  NVIDIA GH200 GPUs for continual training and Titan RTX GPU for SAE training and evaluation. Average execution time for an example task and 10 runs was $204.199\pm42.473$ s on the Titan GPU. We report hyperparameter values in App.~\ref{app:hyperparameters}. We provide our code as Supplementary Material and will release it in a public GitHub repository upon publication.

\section{Experimental analysis}\label{sec:experiments}

Our experiments follow the \emph{analysis modules} defined in Section \ref{sec:methodology}: (1) Binary concept activation analysis, (2) Concept prediction from raw features and (3) Task prediction via linear probes.

\subsection{Concept retention, deletion and recoverability: concept activation analysis}

\begin{figure}[t]
    \centering
    \subfloat[Active neurons]{
        \includegraphics[width=0.42\textwidth]{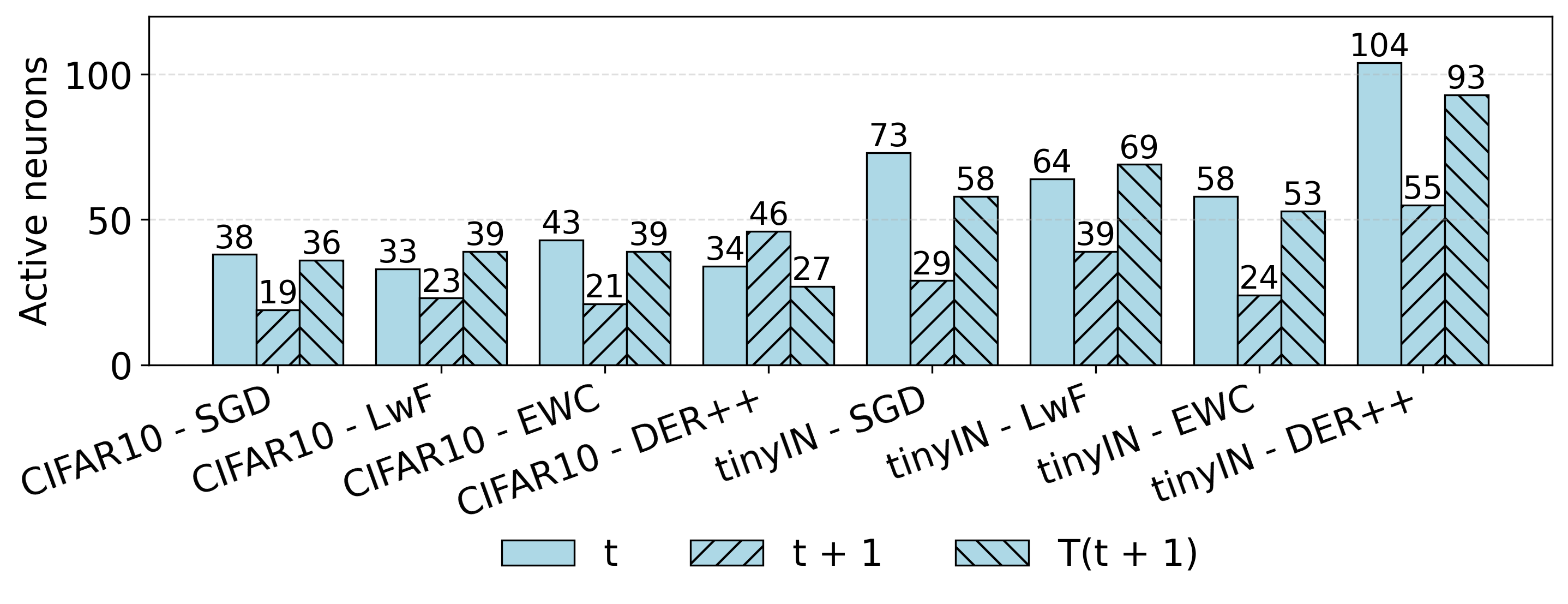}
        \label{fig:active_bar}}
    \hfill
    \subfloat[Deletion ratio]{
        \includegraphics[width=0.42\textwidth]{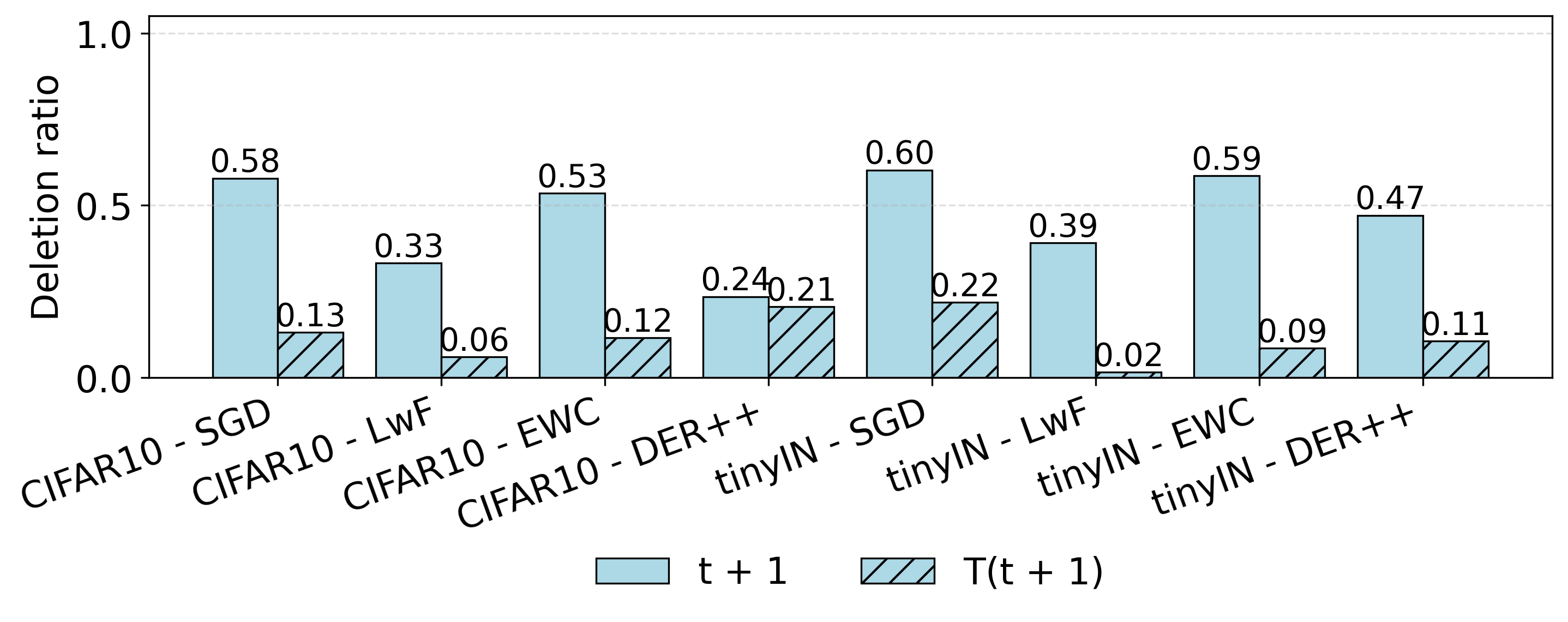}
        \label{fig:deletion_bar}}
\vspace{-3mm}
    \subfloat[Regained concept statistics]{
        \includegraphics[width=0.42\textwidth]{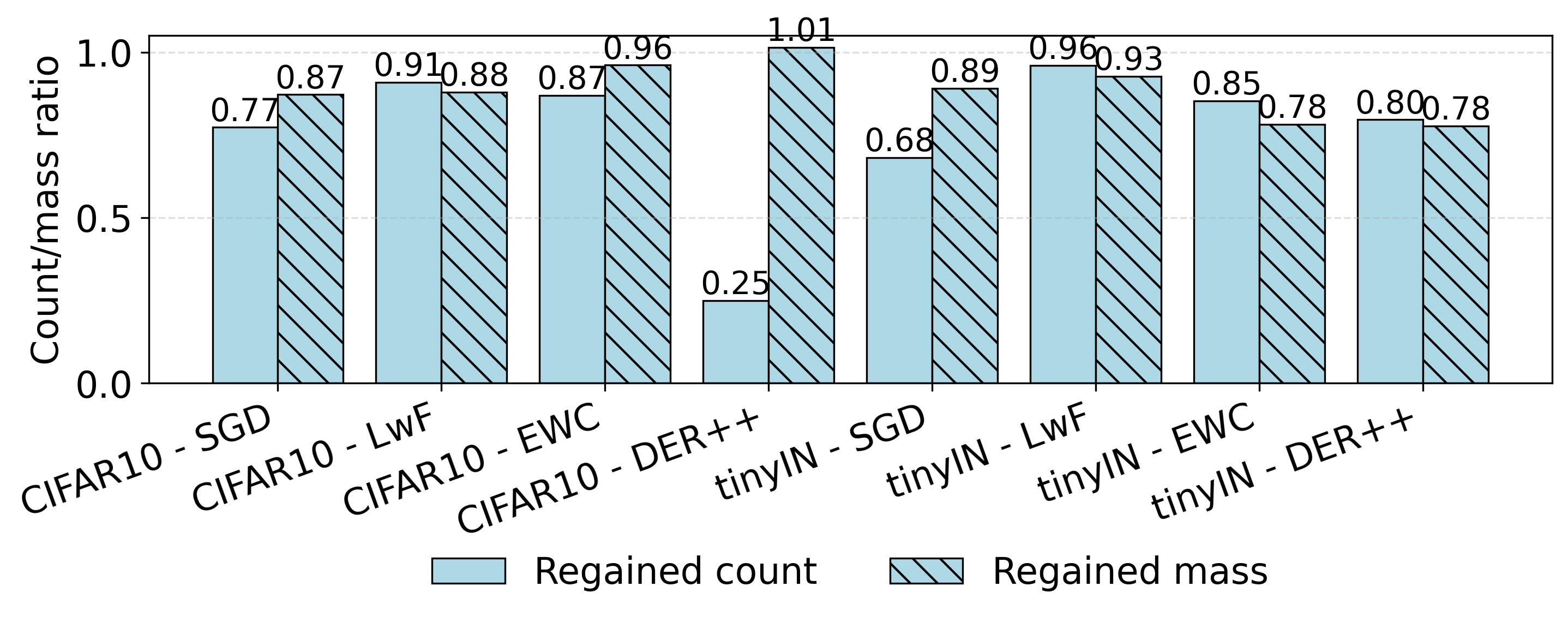}
        \label{fig:regained_count_mass_bar}}
        
    \caption{Concept activation analysis: \textbf{Active concept count}, (seeming) \textbf{deletion ratio} and \textbf{regained concept statistics} for 2seq-CIFAR10 and 2seq-tiny-ImageNet. }\label{fig:active_deletion_bar}
\end{figure}

Fig.~\ref{fig:active_bar} shows the number of active neurons for 2seq-CIFAR10 and 2seq-tiny-ImageNet. Under the same binarization rule, substantially more latent neurons are active for 2seq-tiny-ImageNet, likely reflecting its higher complexity. In all cases, only a small fraction of the SAE latent space is active, corresponding to salient, reusable features. The number of active neurons decreases when task-$t$ data are encoded after training on task $t+1$ for most setups, but a large fraction is recovered after linear translation. An exception is 2seq-CIFAR10 with DER++, where the number of active concepts increases at $t+1$, possibly due to replay-driven expansion of the active feature set. Fig.~\ref{fig:deletion_bar} reveals clear differences between continual strategies: deletion ratios are the highest for SGD and EWC, lower for DER++ and LwF. Linear translation noticeably improves deletion ratios for 2seq-tiny-ImageNet, especially for LwF, where deletion becomes almost zero. Fig.~\ref{fig:regained_count_mass_bar} shows that while the regained mass is similar across settings, the regained count ratio is the highest for LwF,  followed by EWC. It suggests that these strategies may preserve a more linearly recoverable fine-grained space. Overall, these results indicate that many apparently forgotten concepts are not irreversibly lost, but become less readable in the original task-$t$ coordinate system due to approximately linear representational drift. As a qualitative complement, Appendix \ref{app:concept_viz} compares the most activating images for example \textbf{retained} concepts at $t$ and $t+1$ (Figs. \ref{fig:neuron274_examples_retained}, \ref{fig:neuron370_examples_retained}, \ref{fig:neuron425_examples_retained}). Shared visual motifs indicate overlapping information across the two representations. We also show \textbf{recovered} concepts (Figs. \ref{fig:neuron687_examples_deleted_recovered}, \ref{fig:neuron358_examples_deleted_recovered}, \ref{fig:neuron146_examples_deleted_recovered}), whose common motifs further support that concept-level information can be linearly restored.

\begin{figure}[h]
    \centering
    \subfloat[SGD - Raw features]{
        \includegraphics[trim=0cm 4.02cm 0cm 0.1cm, clip, width=0.42\textwidth]{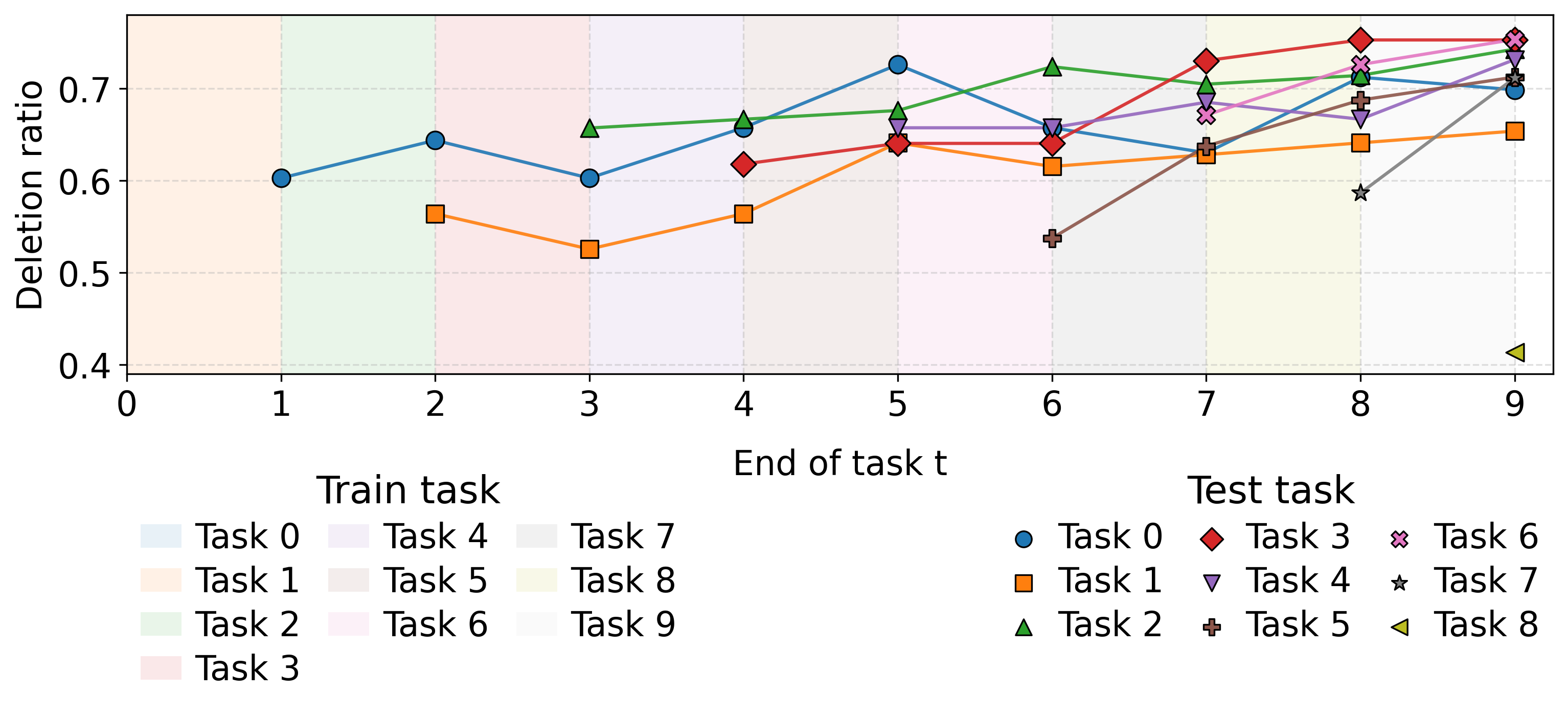}
        \label{fig:deletion_scatter_raw_sgd}}
    \hfill
    \subfloat[SGD - Linearly translated features]{
        \includegraphics[trim=0cm 4.02cm 0cm 0.1cm, clip, width=0.42\textwidth]{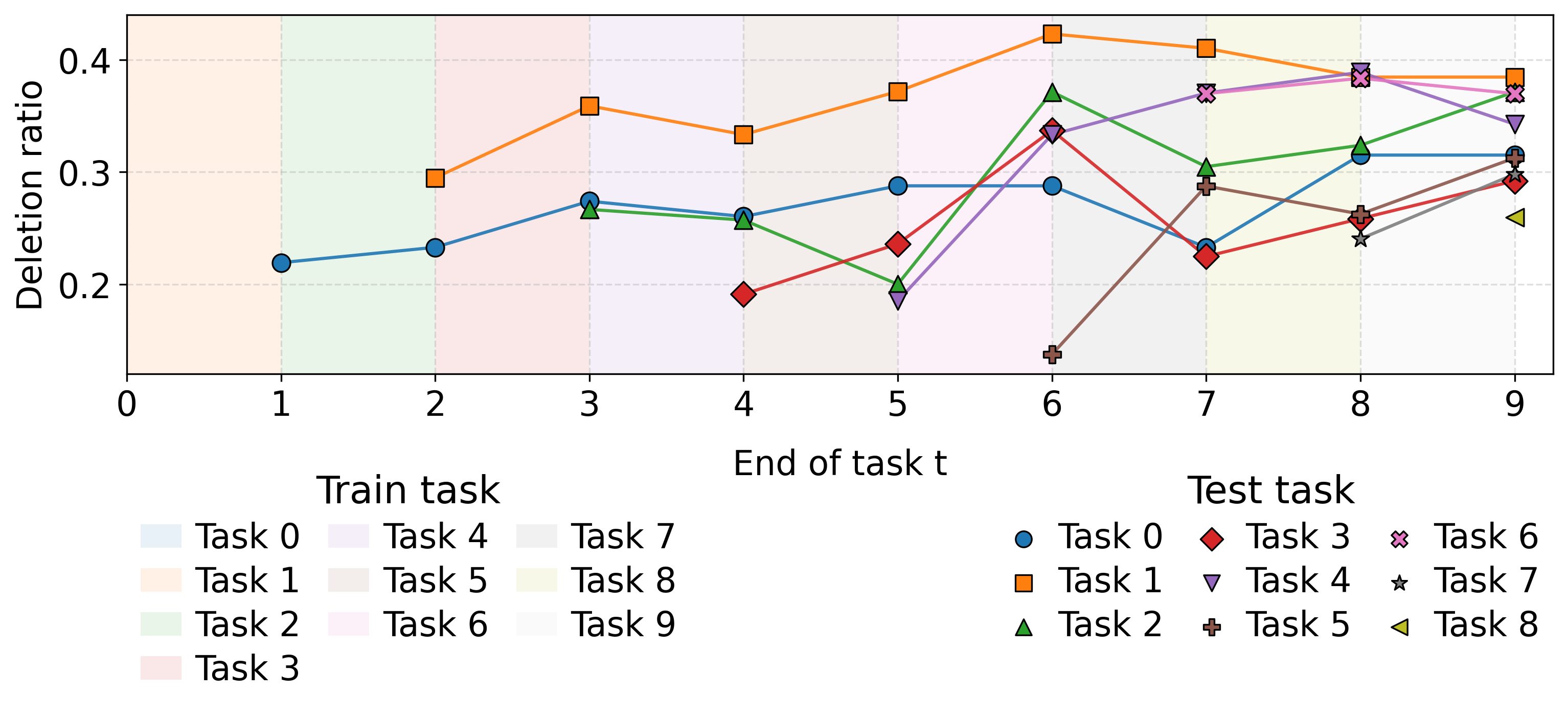}
        \label{fig:deletion_scatter_linear_sgd}}

    \vspace{-3mm}
    \subfloat[LwF - Raw features]{
        \includegraphics[trim=0cm 4.02cm 0cm 0.1cm, clip, width=0.42\textwidth]{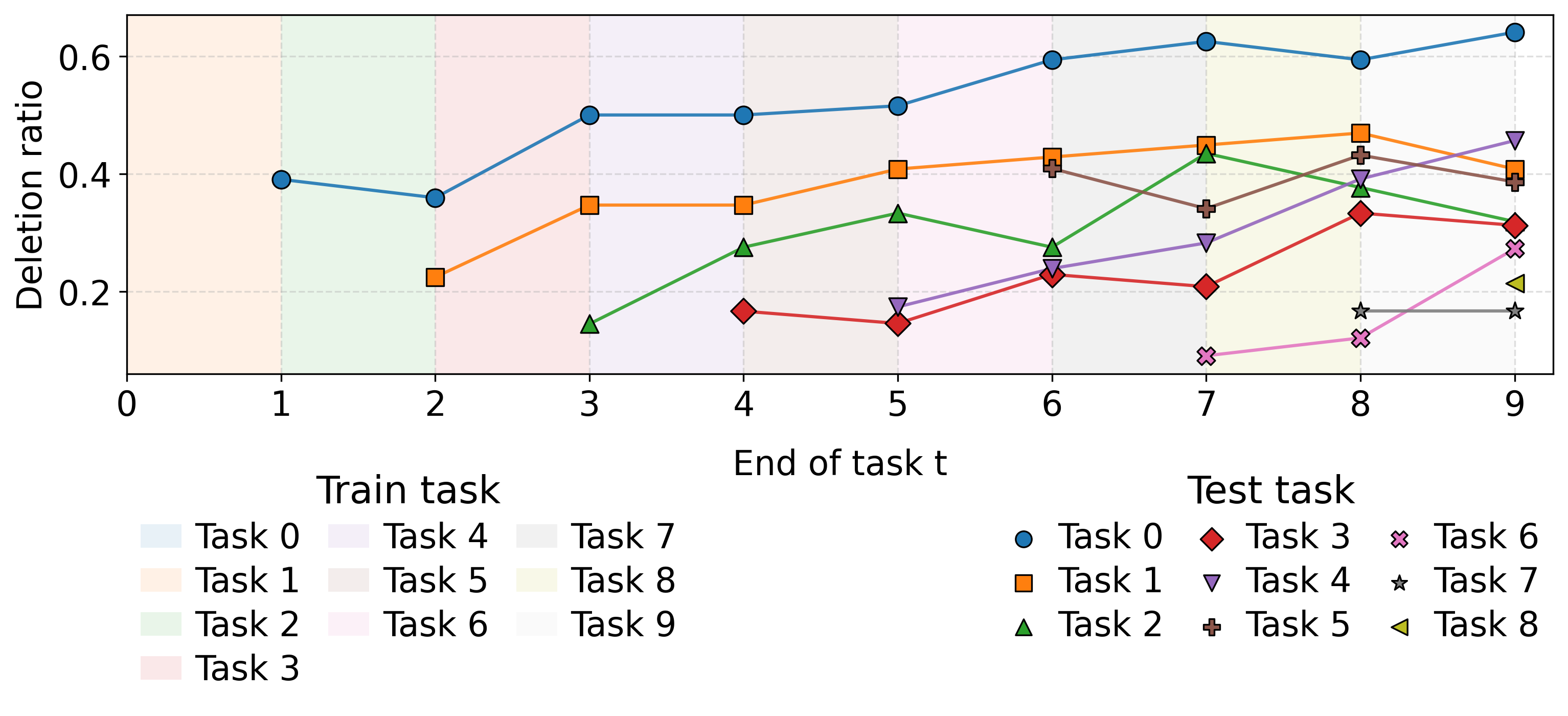}
        \label{fig:deletion_scatter_raw_lwf}}
    \hfill
    \subfloat[LwF - Linearly translated features]{
        \includegraphics[trim=0cm 4.02cm 0cm 0.1cm, clip, width=0.42\textwidth]{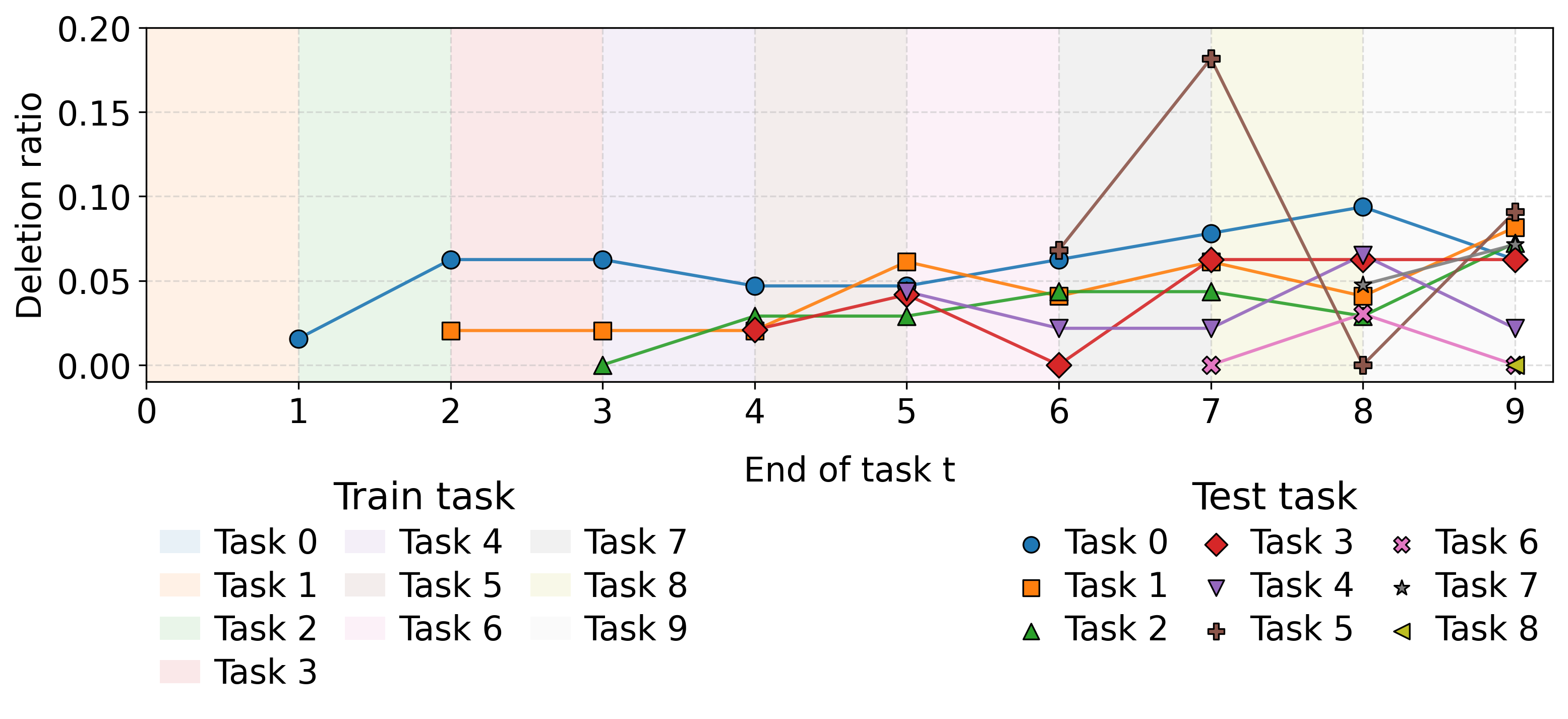}
        \label{fig:deletion_scatter_linear_lwf}}
    \vspace{-3mm}

    \subfloat[EWC - Raw features]{
        \includegraphics[trim=0cm 4.02cm 0cm 0.1cm, clip, width=0.42\textwidth]{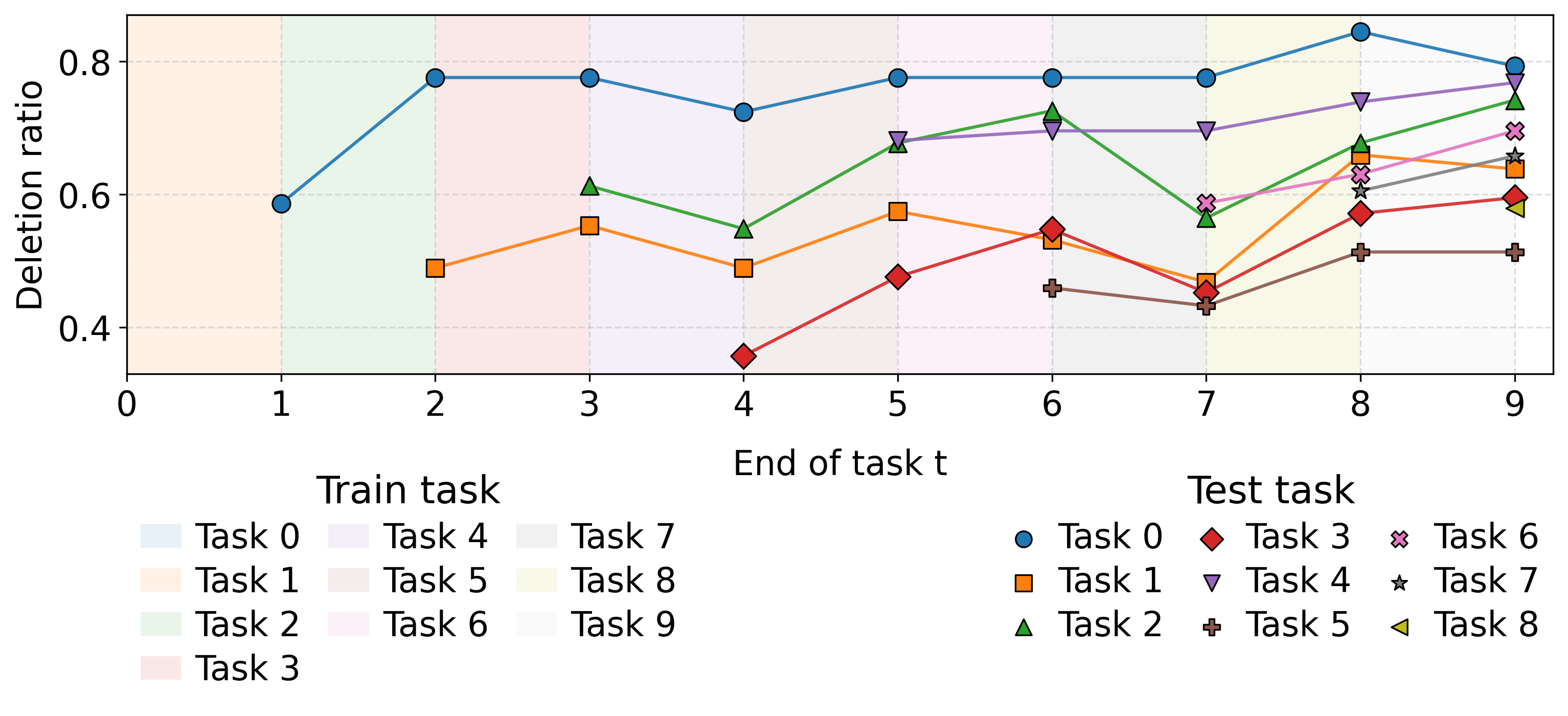}
        \label{fig:deletion_scatter_raw_ewc}}
    \hfill
    \subfloat[EWC - Linearly translated features]{
        \includegraphics[trim=0cm 4.02cm 0cm 0.1cm, clip, width=0.42\textwidth]{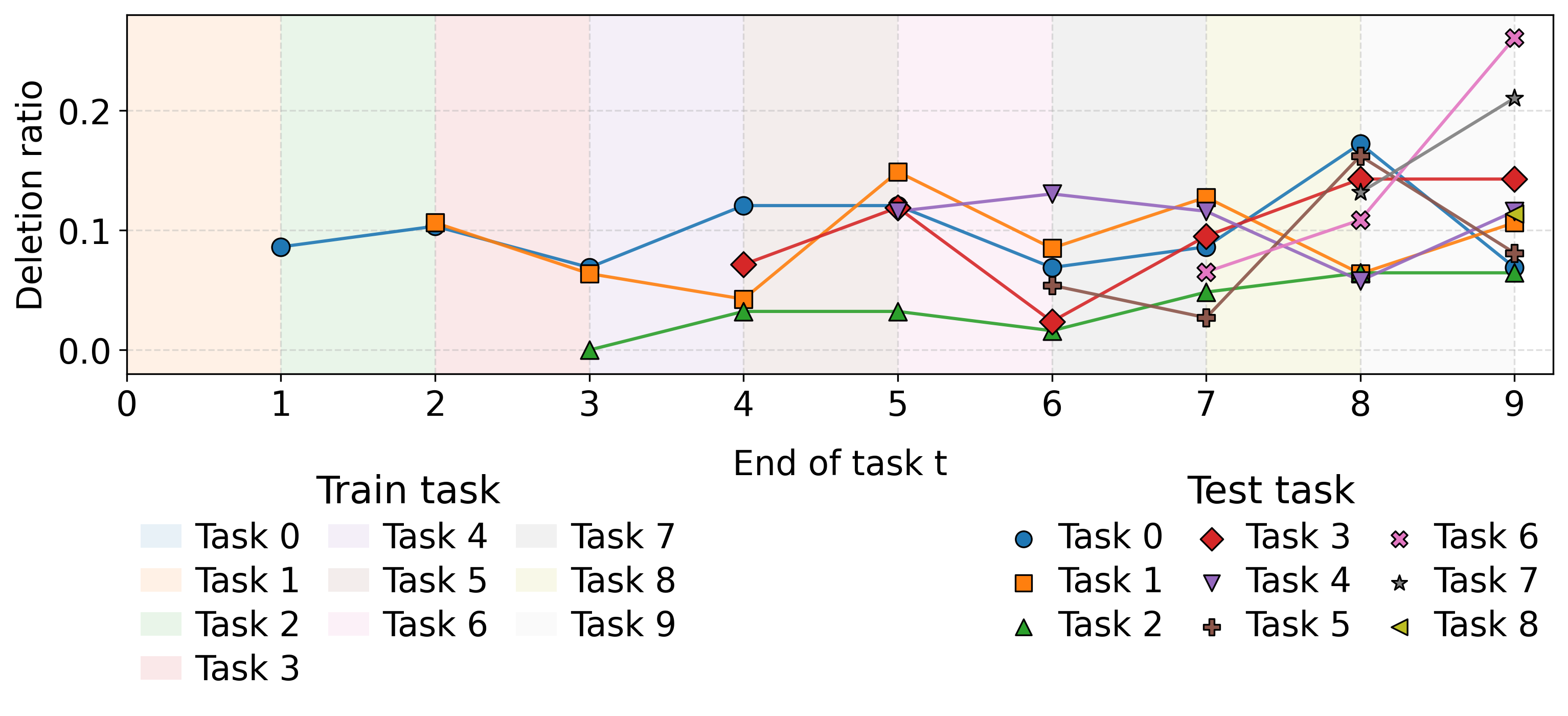}
        \label{fig:deletion_scatter_linear_ewc}}
    \vspace{-3mm}

    \subfloat[DER++ - Raw features]{
        \includegraphics[trim=0cm 4.02cm 0cm 0.1cm, clip, width=0.42\textwidth]{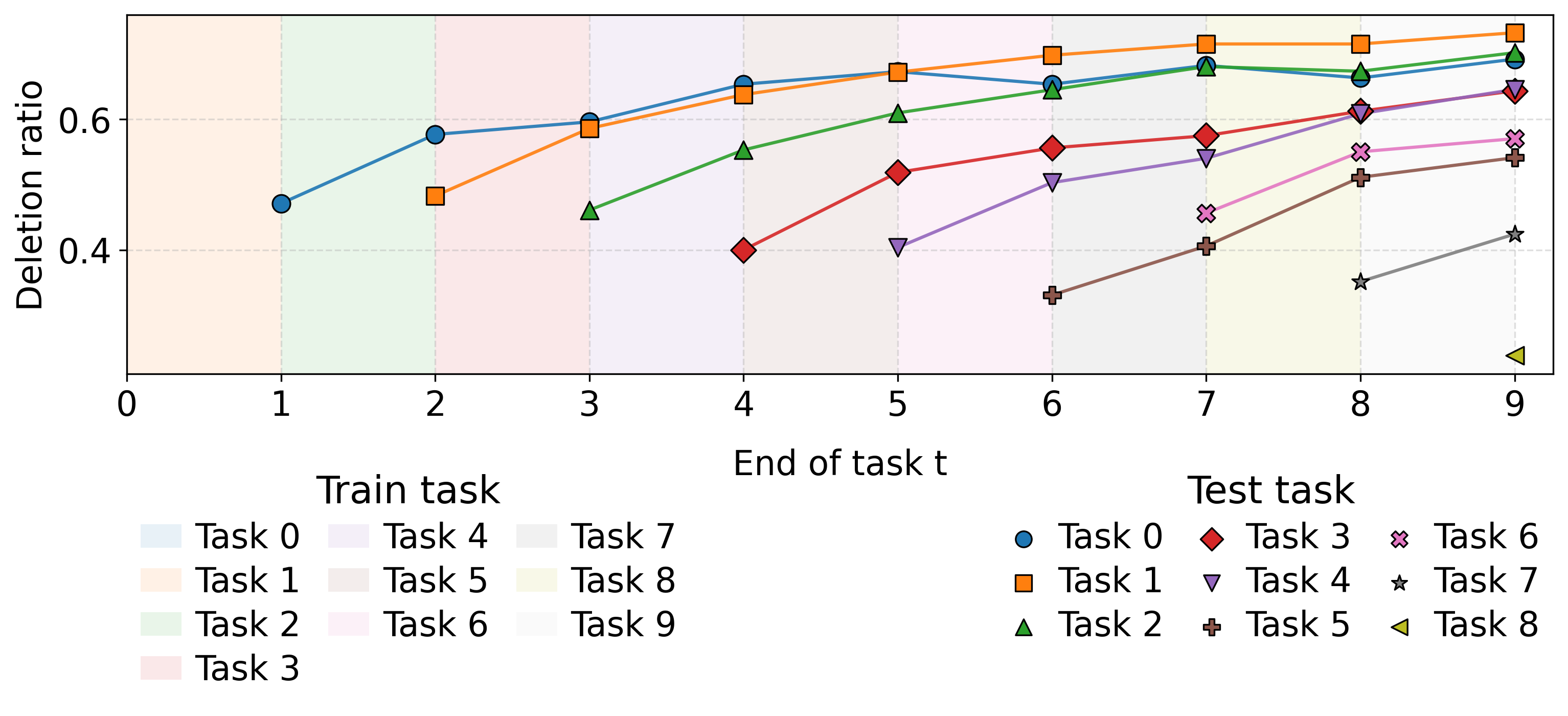}
        \label{fig:deletion_scatter_raw_derpp}}
    \hfill
    \subfloat[DER++ - Linearly translated features]{
        \includegraphics[trim=0cm 4.02cm 0cm 0.1cm, clip, width=0.42\textwidth]{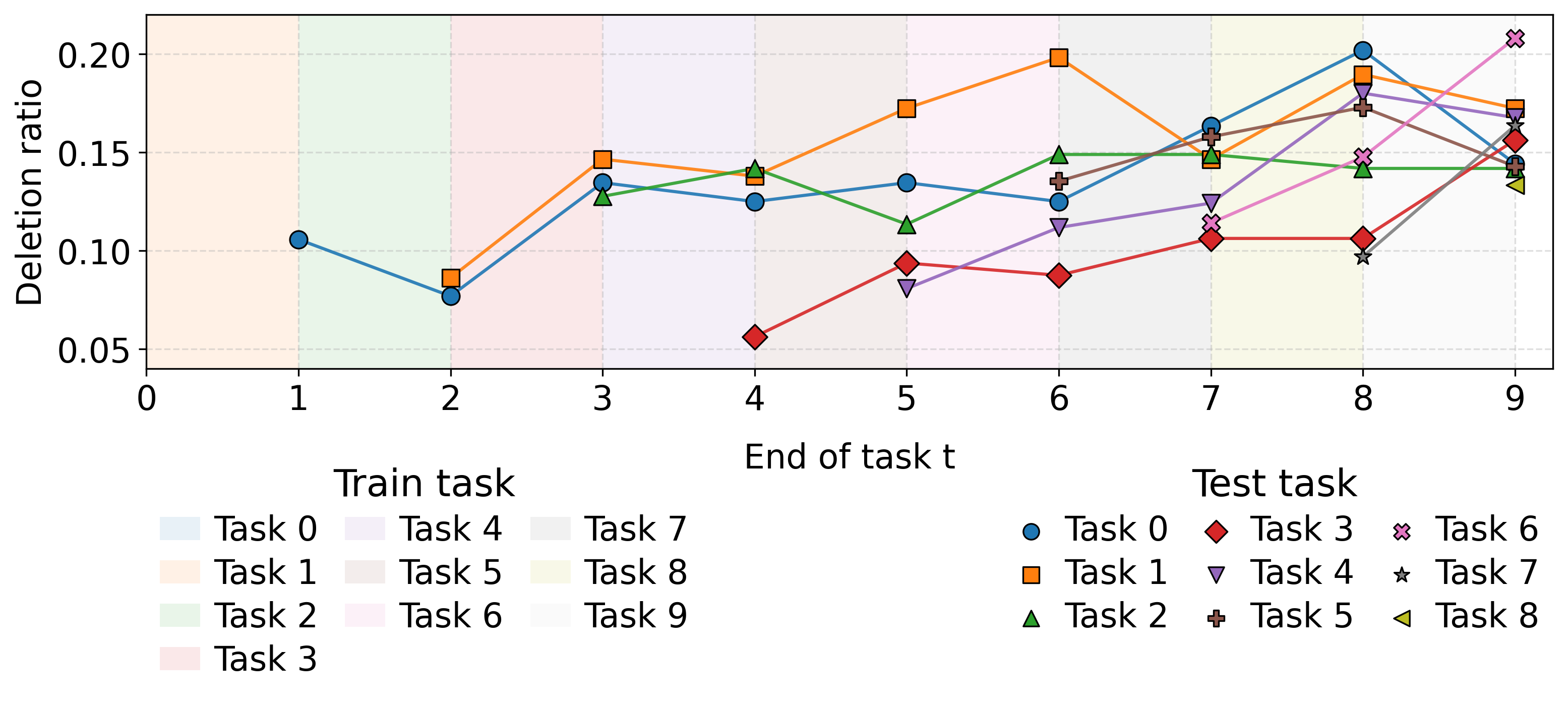}
        \label{fig:deletion_scatter_linear_derpp}}
    \vspace{-3mm}

    \subfloat[Common legend]{
        \includegraphics[trim=0cm 0cm 0cm 8.107cm, clip, width=0.5\textwidth]{FIGS/SCATTER/LWF/deletion_ratio_-_linear_t_tinyin-10_der.png}}
    \caption{\textbf{Deletion ratio} for all tasks of 10seq-tiny-ImageNet throughout the continual training.}\label{fig:deletion_scatter}
\end{figure}

\begin{figure}[h]
    \centering
    \subfloat[SGD]{
        \includegraphics[trim=0cm 4.02cm 0cm 0.1cm, clip, width=0.42\textwidth]{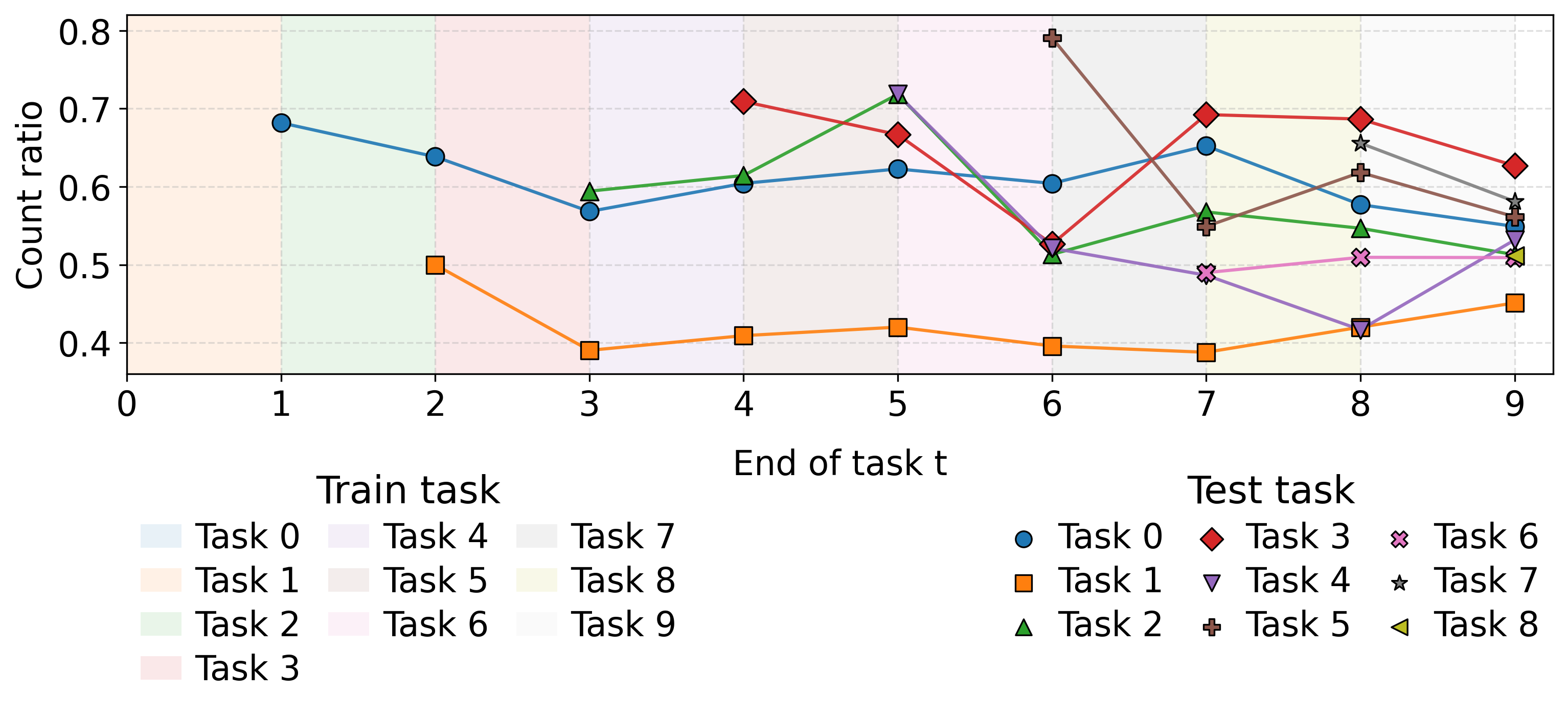}
    \label{fig:regained_activation_count_scatter_sgd}}
    \hfill
     \subfloat[LwF]{
        \includegraphics[trim=0cm 4.02cm 0cm 0.1cm, clip, width=0.42\textwidth]{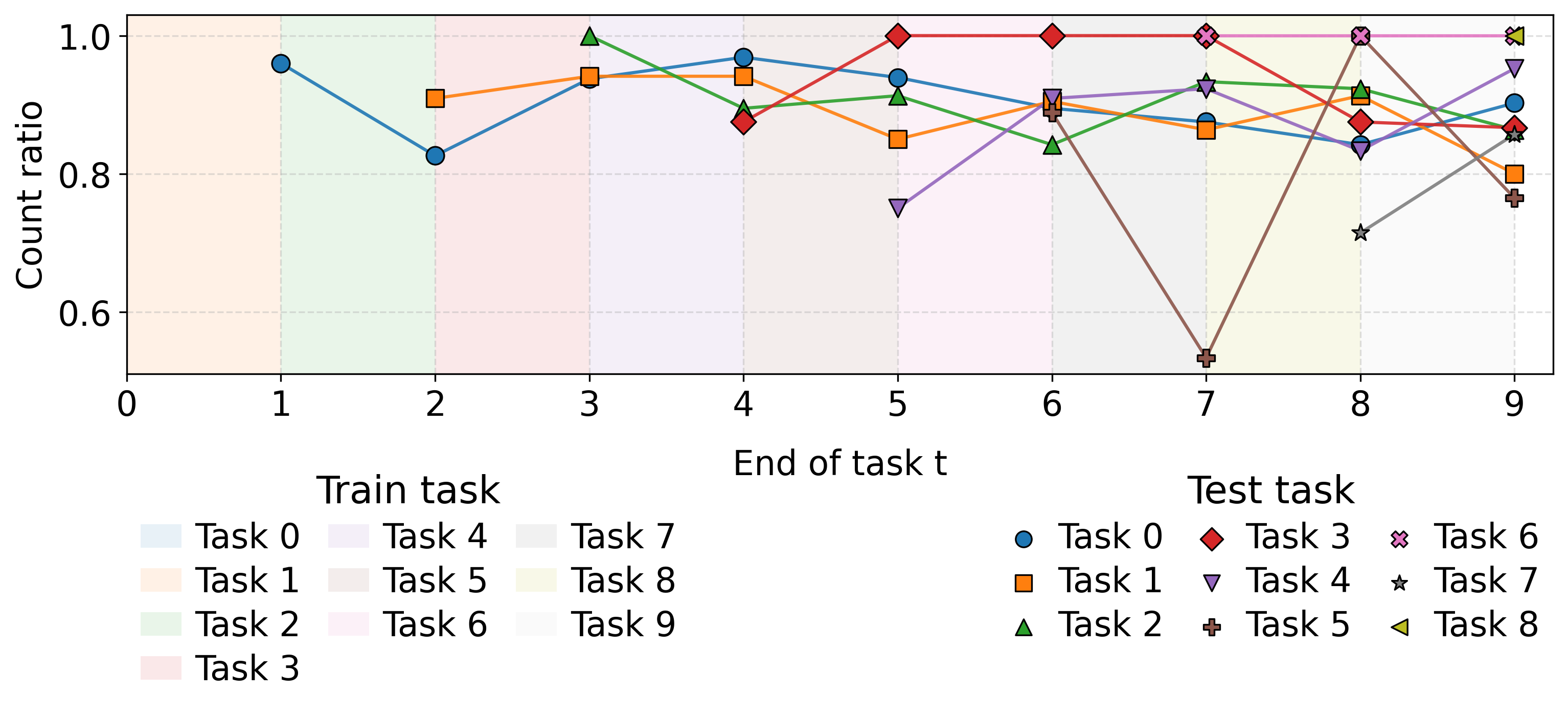}
        \label{fig:regained_activation_count_scatter_lwf}}
    \vspace{-3mm}

    \subfloat[EWC]{
        \includegraphics[trim=0cm 4.02cm 0cm 0.1cm, clip, width=0.42\textwidth]{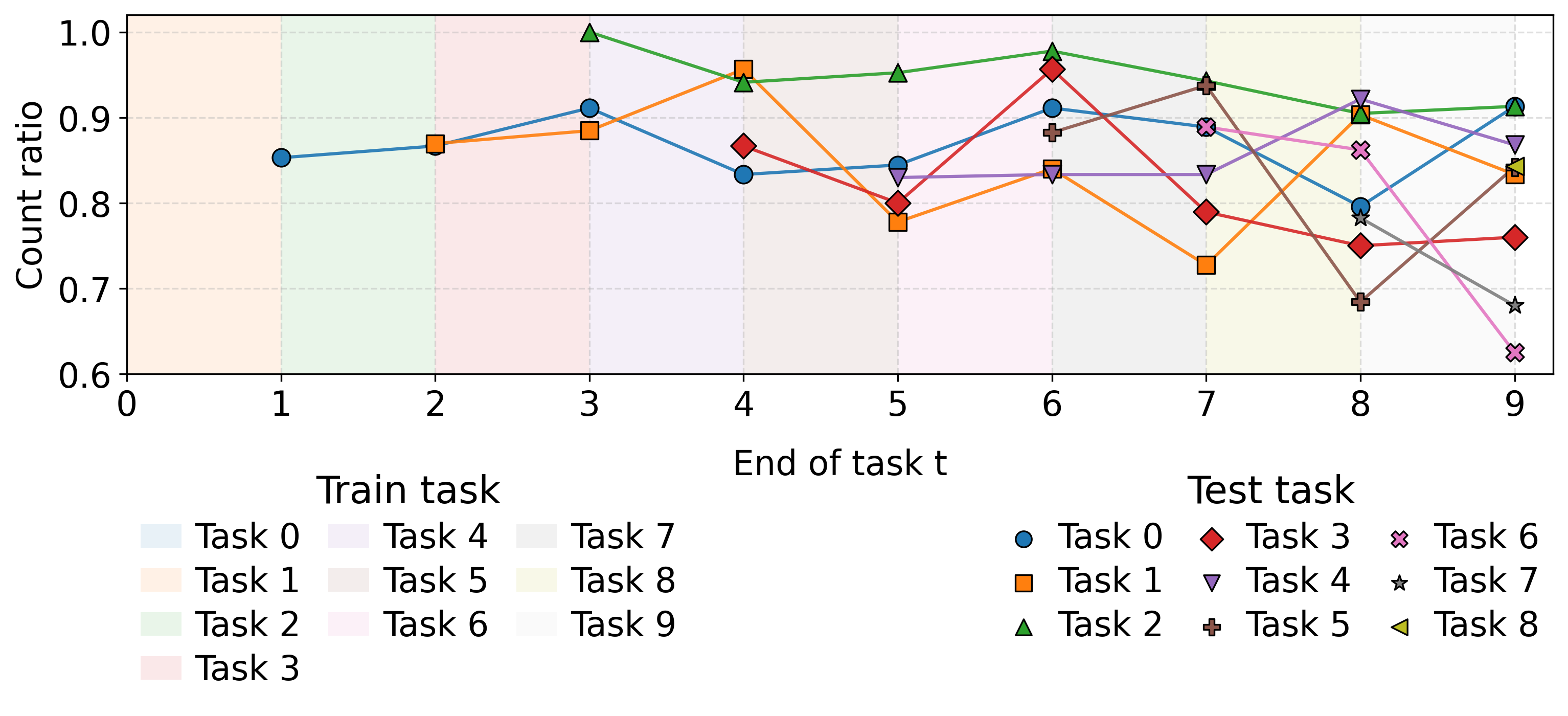}
        \label{fig:regained_activation_count_scatter_ewc}}
    \hfill    
    \subfloat[DER++]{
        \includegraphics[trim=0cm 4.02cm 0cm 0.1cm, clip, width=0.42\textwidth]{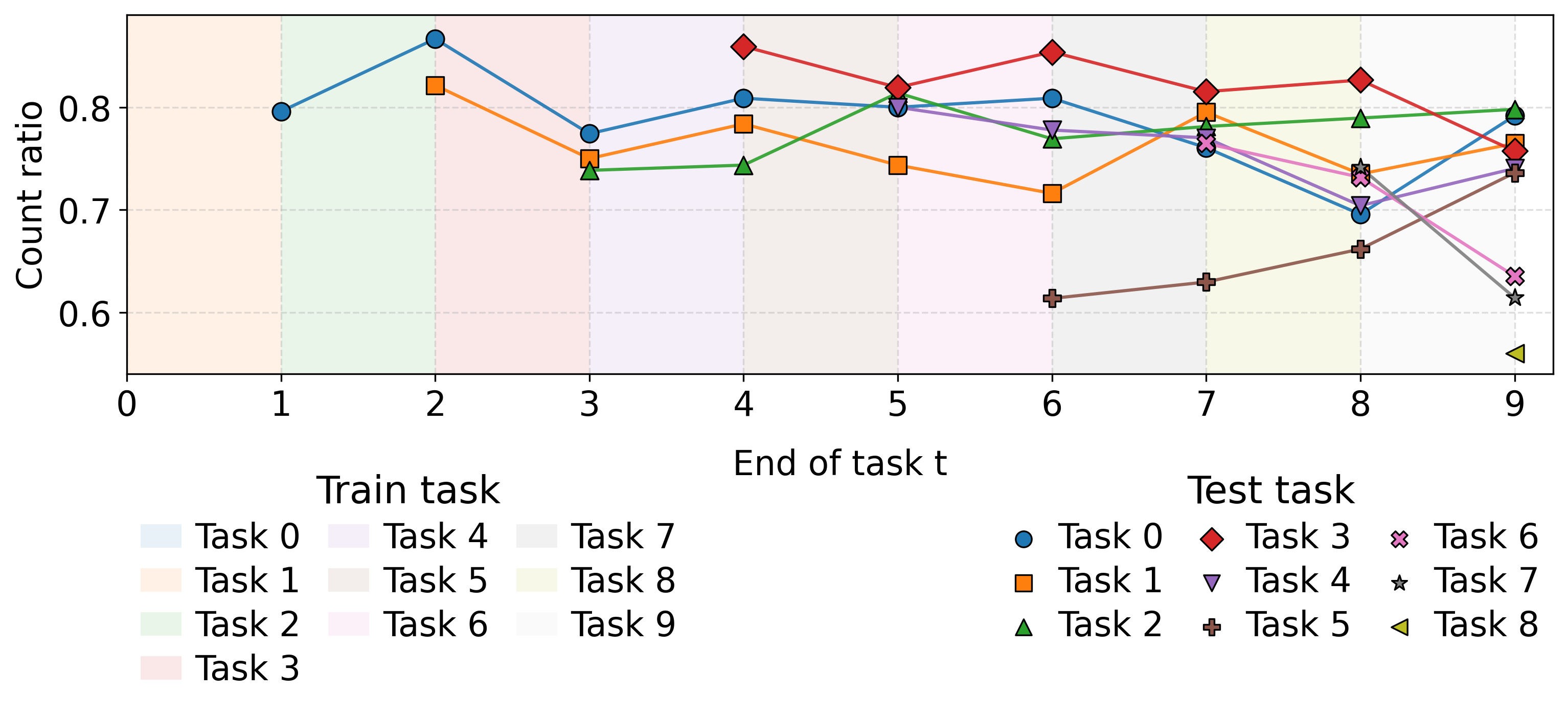}
        \label{fig:regained_activation_count_scatter_derpp}}
    \caption{\textbf{Regained concept count ratio} for all tasks of 10seq-tiny-ImageNet throughout the continual training after the linear translation. We use the same legend as in Fig. \ref{fig:deletion_scatter}.}\label{fig:regained_scatter}
\end{figure}

Figure \ref{fig:deletion_scatter} shows deletion ratios for each task $t$ after all subsequent tasks $t+s$ on 10seq-tiny-ImageNet. For raw representations, deletion ratios increase with $s$ under all strategies, with consistently higher values for SGD and EWC. Under LwF and DER++, later tasks start from lower deletion levels (at $t+1$), suggesting improved resistance to degradation. For DER++, deletion increases progressively with $s$, although with a decelerating trend over time (log-like). Across all tasks, linear alignment substantially reduces deletion ratios, indicating that much of the drift is partially reversible. The strength of this reduction is weaker for DER++ than for LwF. The regained count ratios in Fig. \ref{fig:regained_scatter} follow similar patterns as in the bar plots: they are the highest for LwF and EWC, comparable for SGD and DER++. Increases in deletion rates at higher $s$ and slight decreases in regained counts (SGD, DER++, EWC), suggest gradually reduced linear recoverability of concept activations.

\subsection{Concept decodability: concept prediction from raw features}

\begin{figure}[h]
    \centering
    \subfloat[Mean balanced accuracy and F1 score]{
        \includegraphics[width=0.42\textwidth]{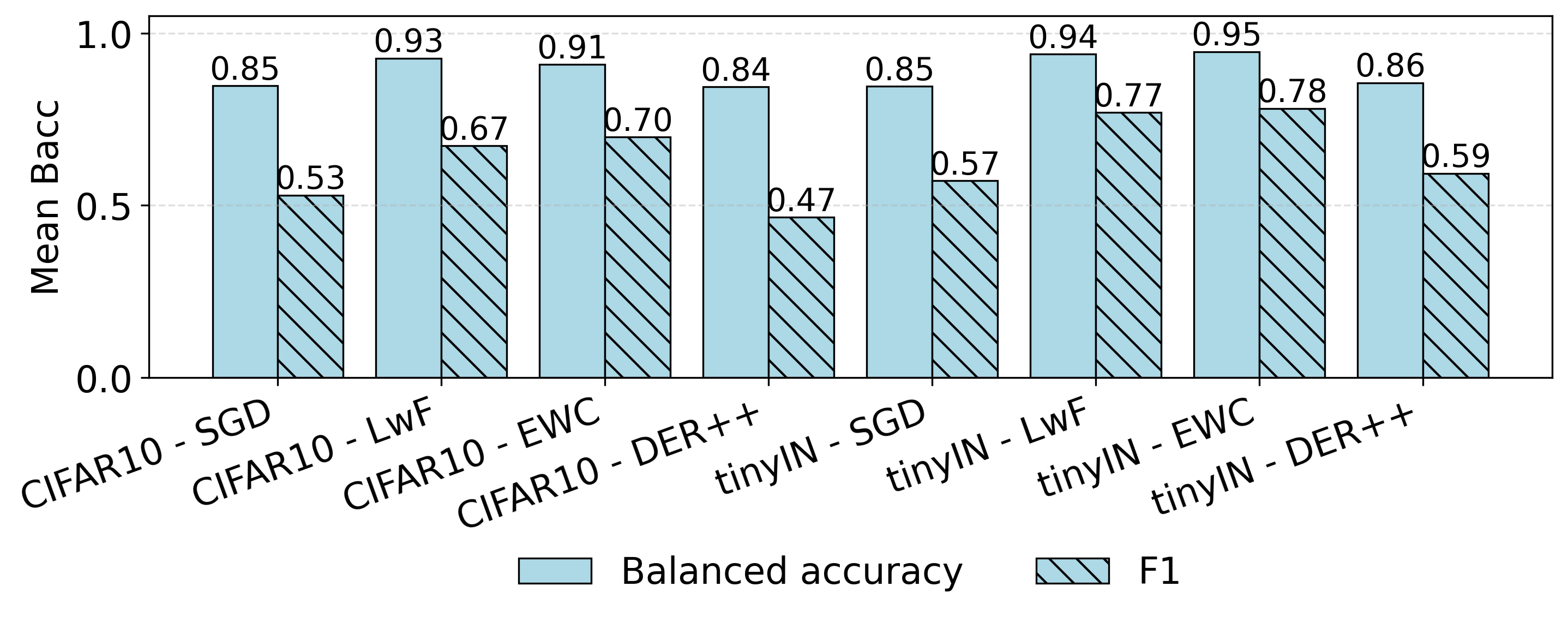}
        \label{fig:mean_bacc_f1}}
    \subfloat[Balanced accuracy distribution]{
        \includegraphics[width=0.42\textwidth]{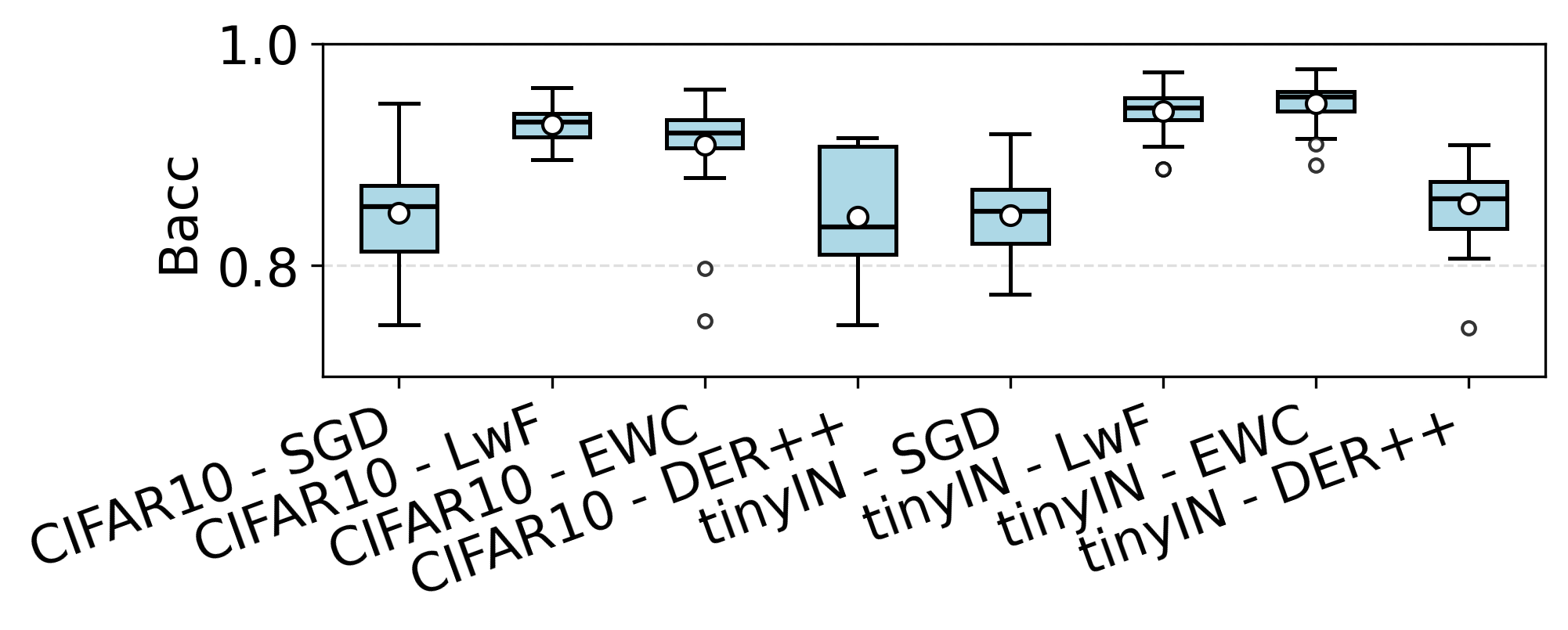}
        \label{fig:bacc_box}}
        \hfill
    \subfloat[F1 score distribution]{
        \includegraphics[width=0.42\textwidth]{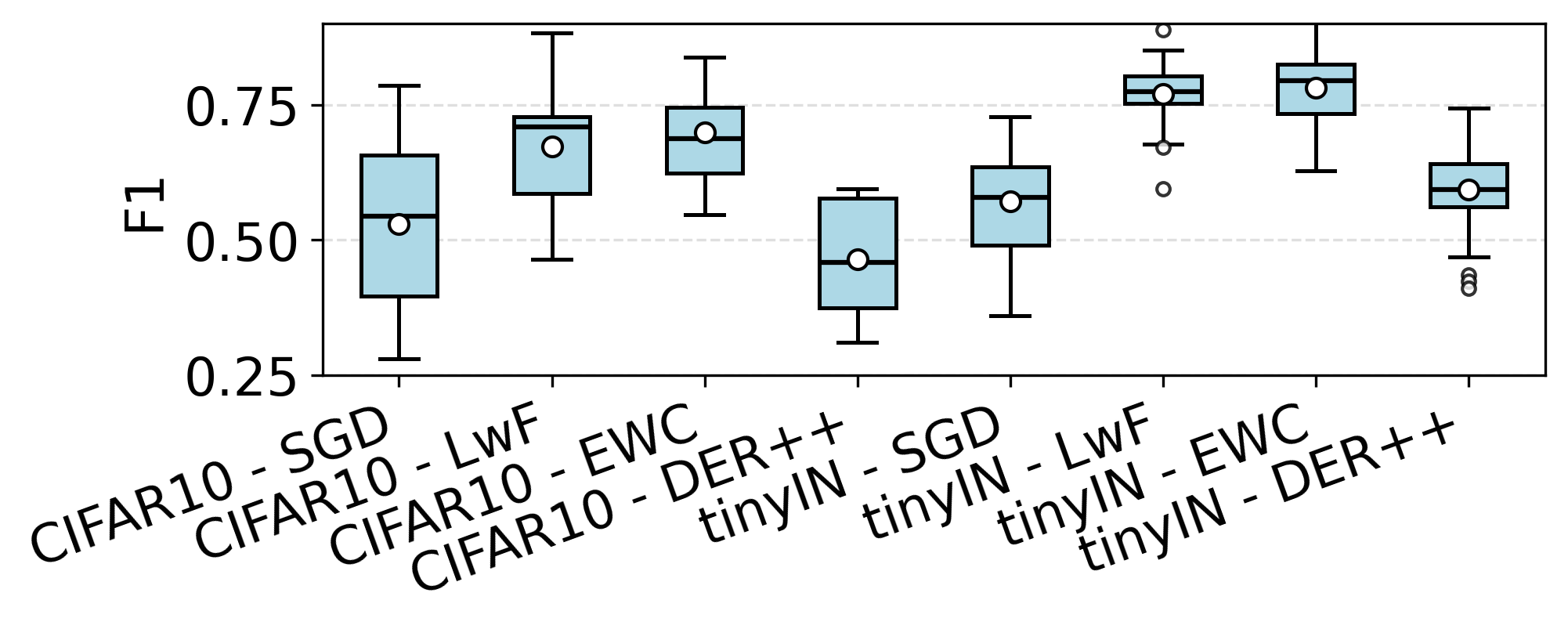}
        \label{fig:f1_box}}
    \caption{Mean \textbf{balanced accuracy} and \textbf{F1 score} along with their distributions for the concept prediction probe on 2seq-CIFAR10 and 2seq-tiny-ImageNet.}\label{fig:box_bacc_f1}
\end{figure}

Fig.~\ref{fig:mean_bacc_f1} reports the mean balanced accuracy and F1 of linear classifiers trained to predict concepts that become inactive in raw task-$t$ representations after learning task $t+1$, with the corresponding score distributions shown in Figs.~\ref{fig:f1_box} (F1) and~\ref{fig:bacc_box} (balanced accuracy). The results show that LwF and EWC preserve substantially higher concept decodability at $t+1$ than SGD and DER++, suggesting that, beyond coarse-grained class information~\cite{davari2022probing}, finer-grained concept-level information can also remain accessible, especially under forgetting mitigation with strategies differing in decodability preservation (LwF, EWC being superior to DER++). At the same time, the spread of balanced accuracy and F1 across concepts indicates that the recoverability is non-uniform: some concepts remain easily decodable, whereas others degrade strongly. For LwF and EWC, the distributions are shifted toward higher values and are more compact, further indicating their broader and more consistent preservation of information at the level of concept proxies. Overall, these results suggest that fine-grained forgetting is often selective and partial rather than uniform or complete.

\begin{figure}[h]
    \centering
    \subfloat[SGD - Mean balanced accuracy]{
        \includegraphics[trim=0cm 4.02cm 0cm 0.1cm, clip, width=0.42\textwidth]{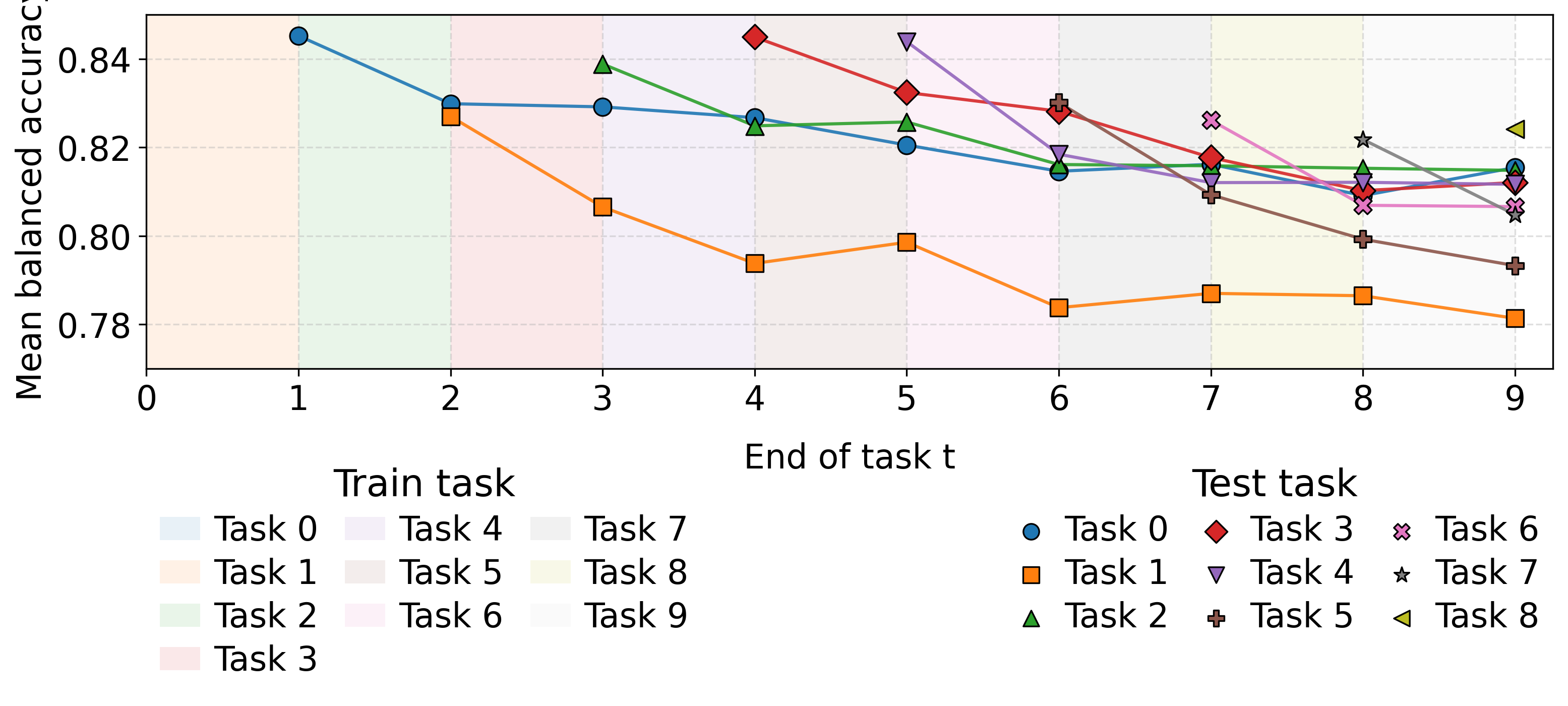}
        \label{fig:mean_bacc_scatter_sgd}}
    \hfill
    \subfloat[SGD - Mean F1 score]{
        \includegraphics[trim=0cm 4.02cm 0cm 0.1cm, clip, width=0.42\textwidth]{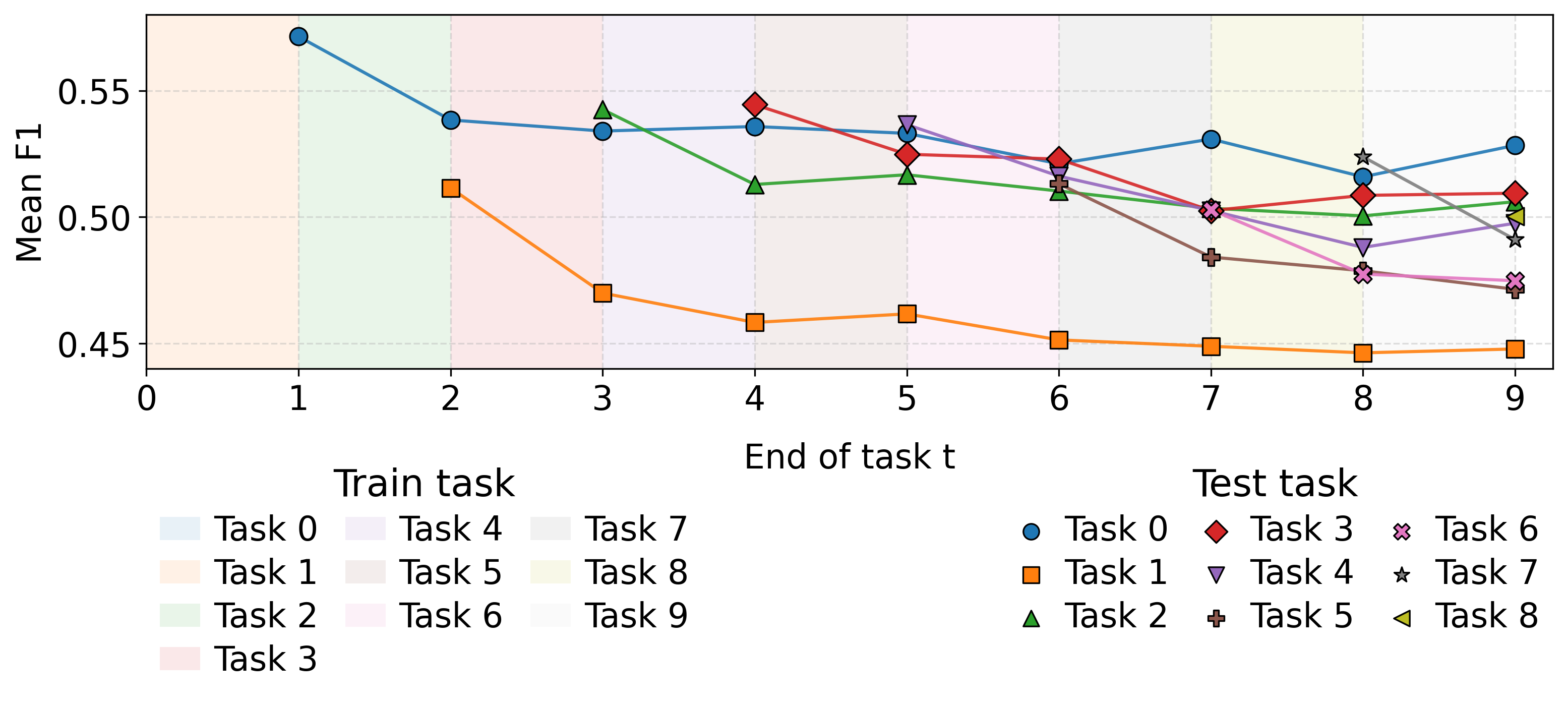}
        \label{fig:mean_f1_scatter_sgd}}
    \vspace{-3mm}

    \subfloat[LwF - Mean balanced accuracy]{
        \includegraphics[trim=0cm 4.02cm 0cm 0.1cm, clip, width=0.42\textwidth]{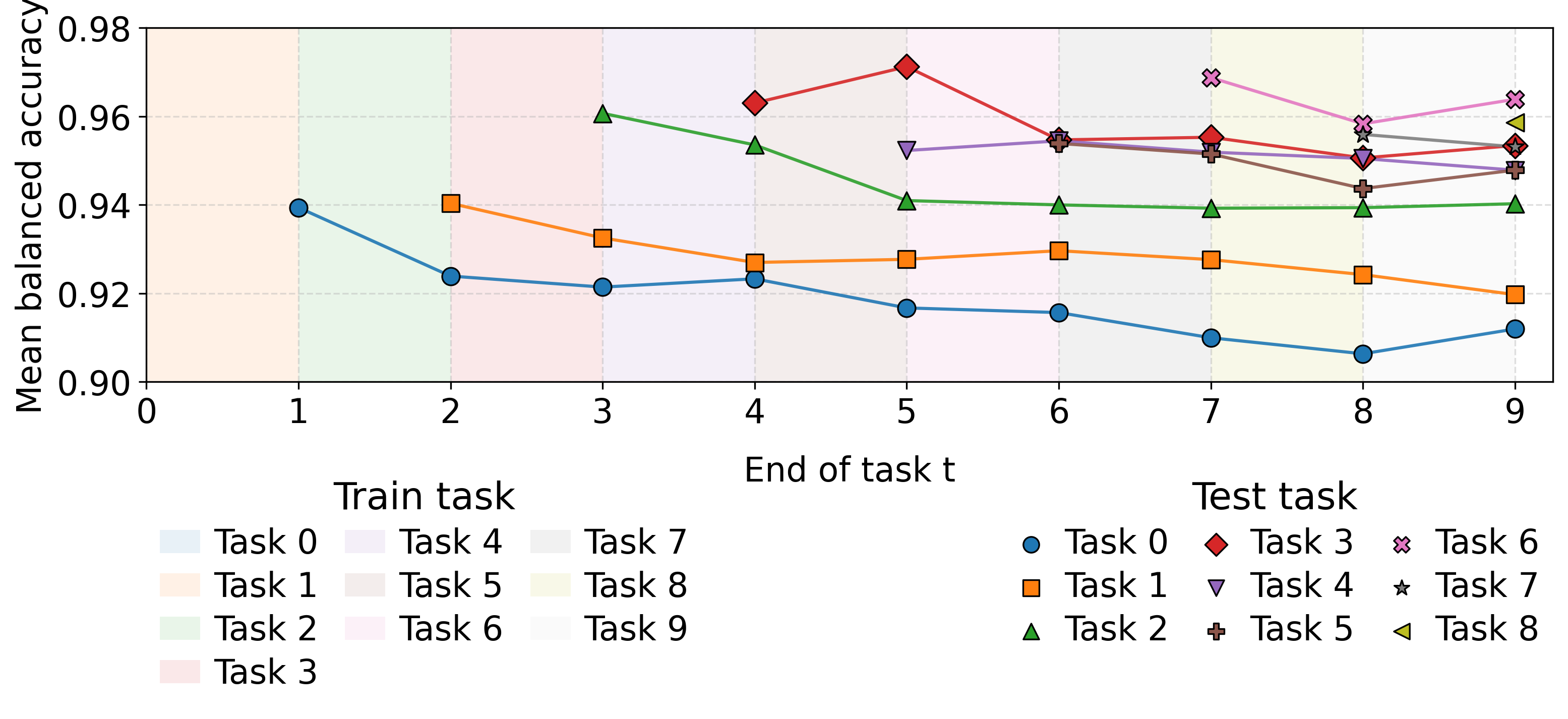}
        \label{fig:mean_bacc_scatter_lwf}}
    \hfill
    \subfloat[LwF - Mean F1 score]{
        \includegraphics[trim=0cm 4.02cm 0cm 0.1cm, clip, width=0.42\textwidth]{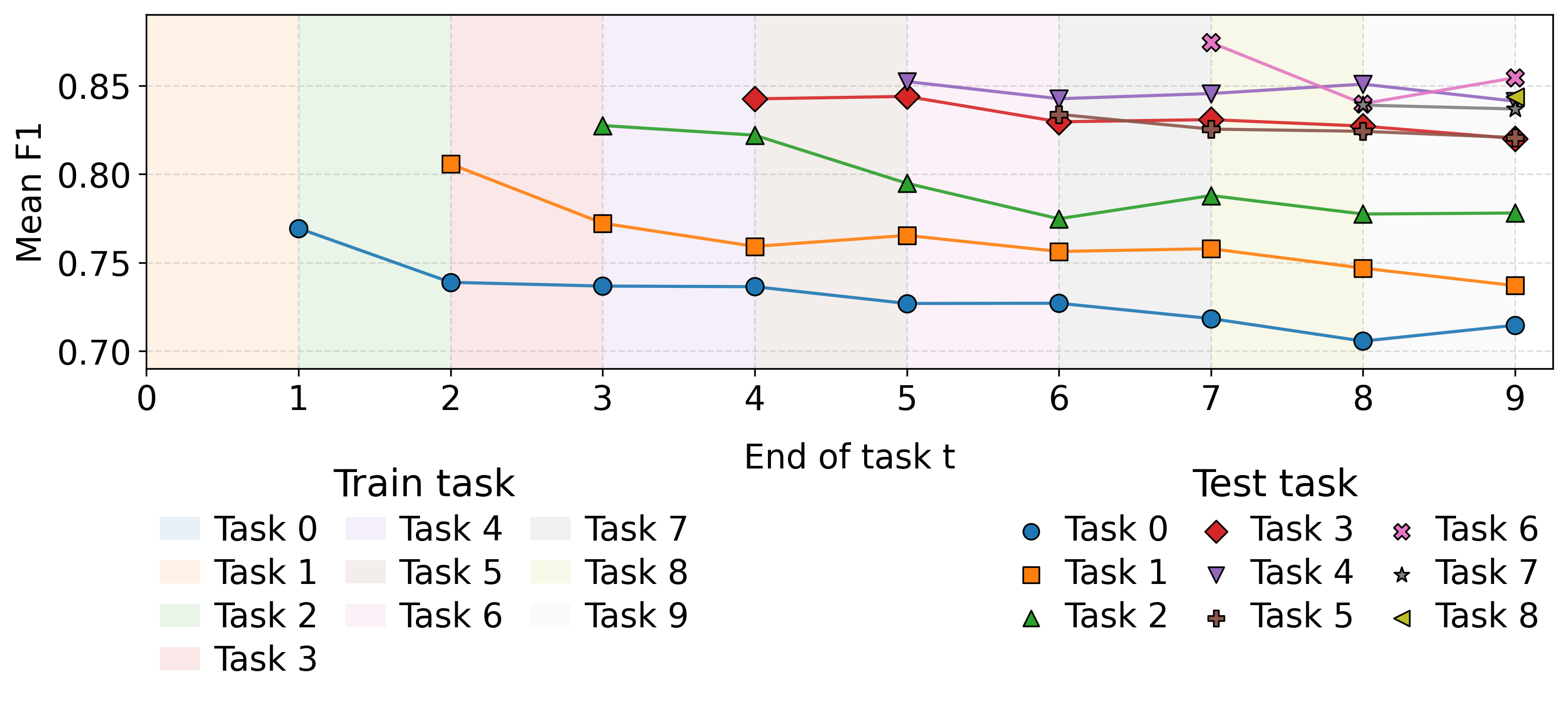}
        \label{fig:mean_f1_scatter_lwf}}
    \vspace{-3mm}

    \subfloat[EWC - Mean balanced accuracy]{
        \includegraphics[trim=0cm 4.02cm 0cm 0.1cm, clip, width=0.42\textwidth]{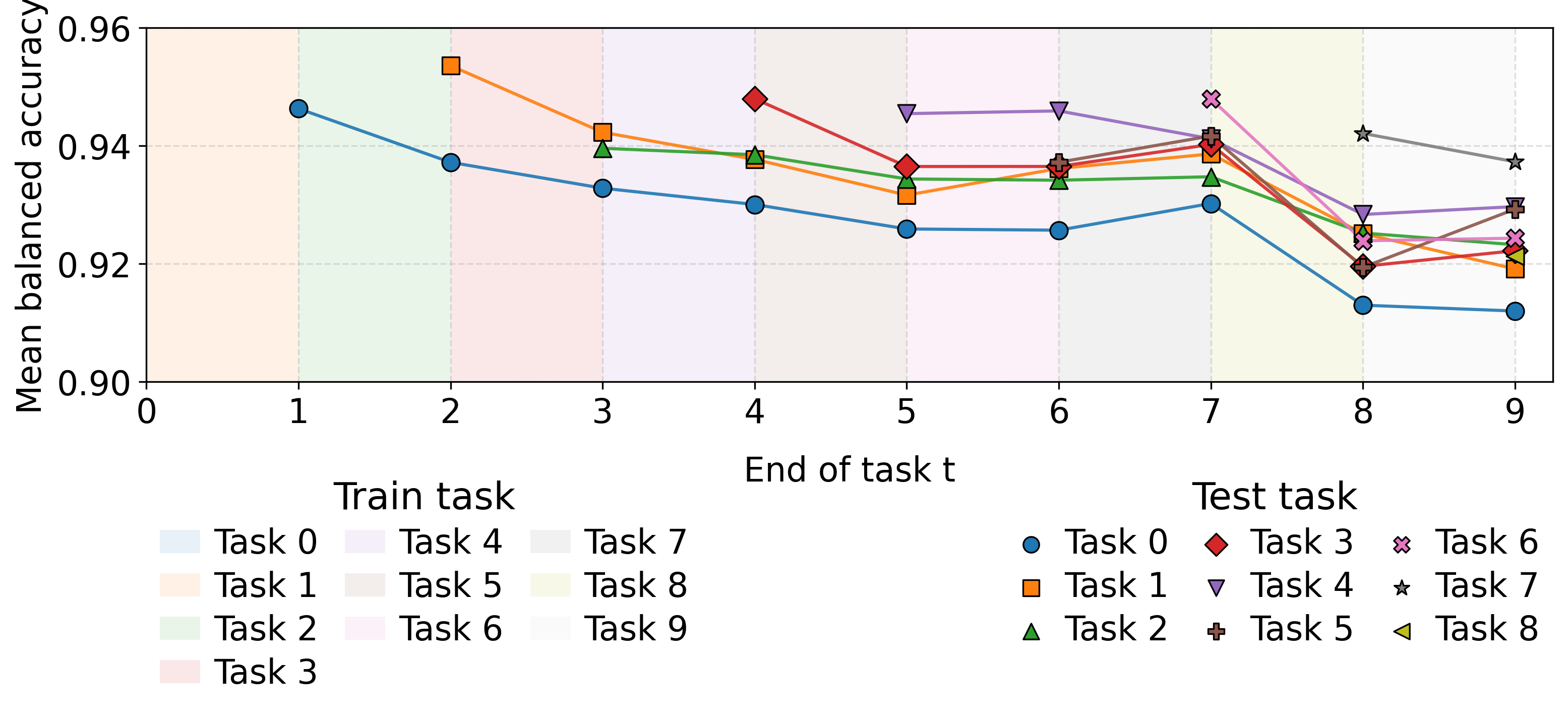}
        \label{fig:mean_bacc_scatter_ewc}}
    \hfill
    \subfloat[EWC - Mean F1 score]{
        \includegraphics[trim=0cm 4.02cm 0cm 0.1cm, clip, width=0.42\textwidth]{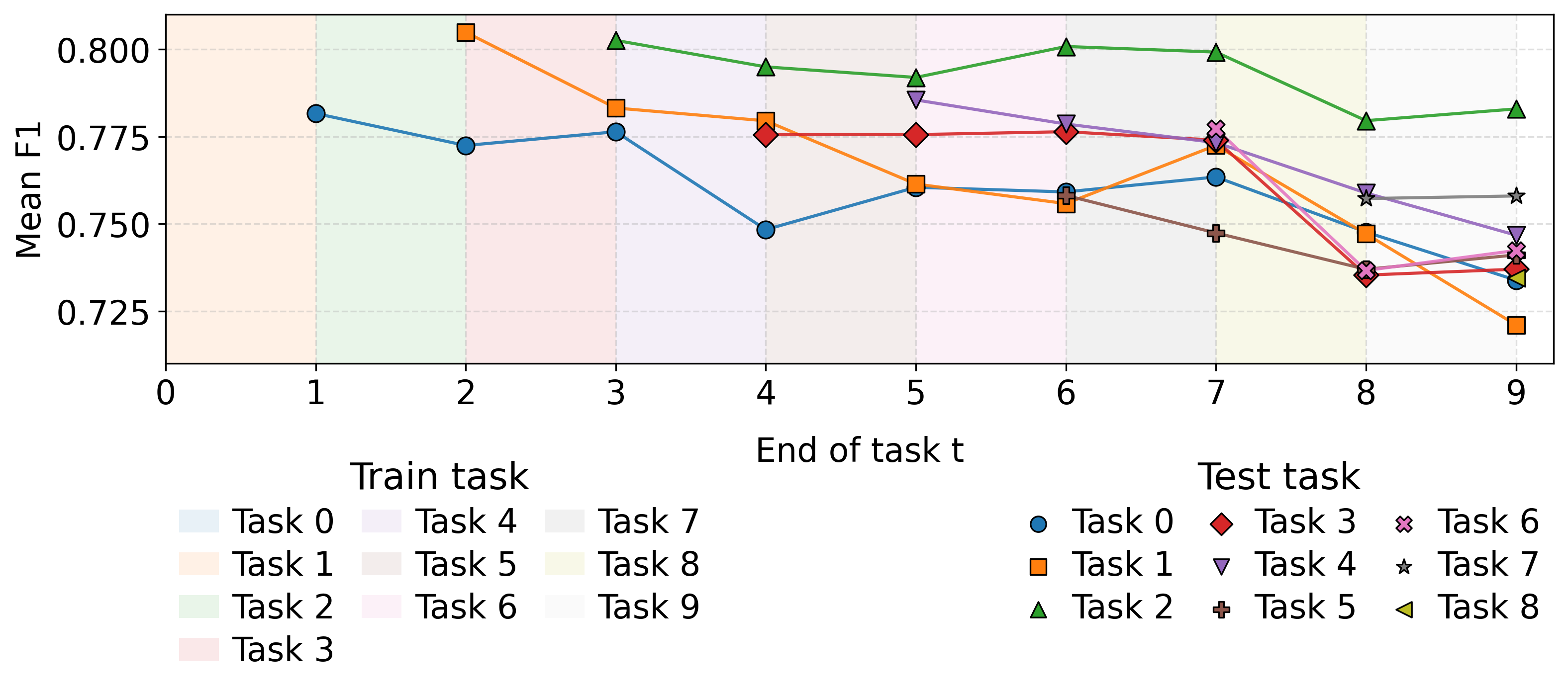}
        \label{fig:mean_f1_scatter_ewc}}
    \vspace{-3mm}

     \subfloat[DER++ - Mean balanced accuracy]{
        \includegraphics[trim=0cm 4.02cm 0cm 0.1cm, clip, width=0.42\textwidth]{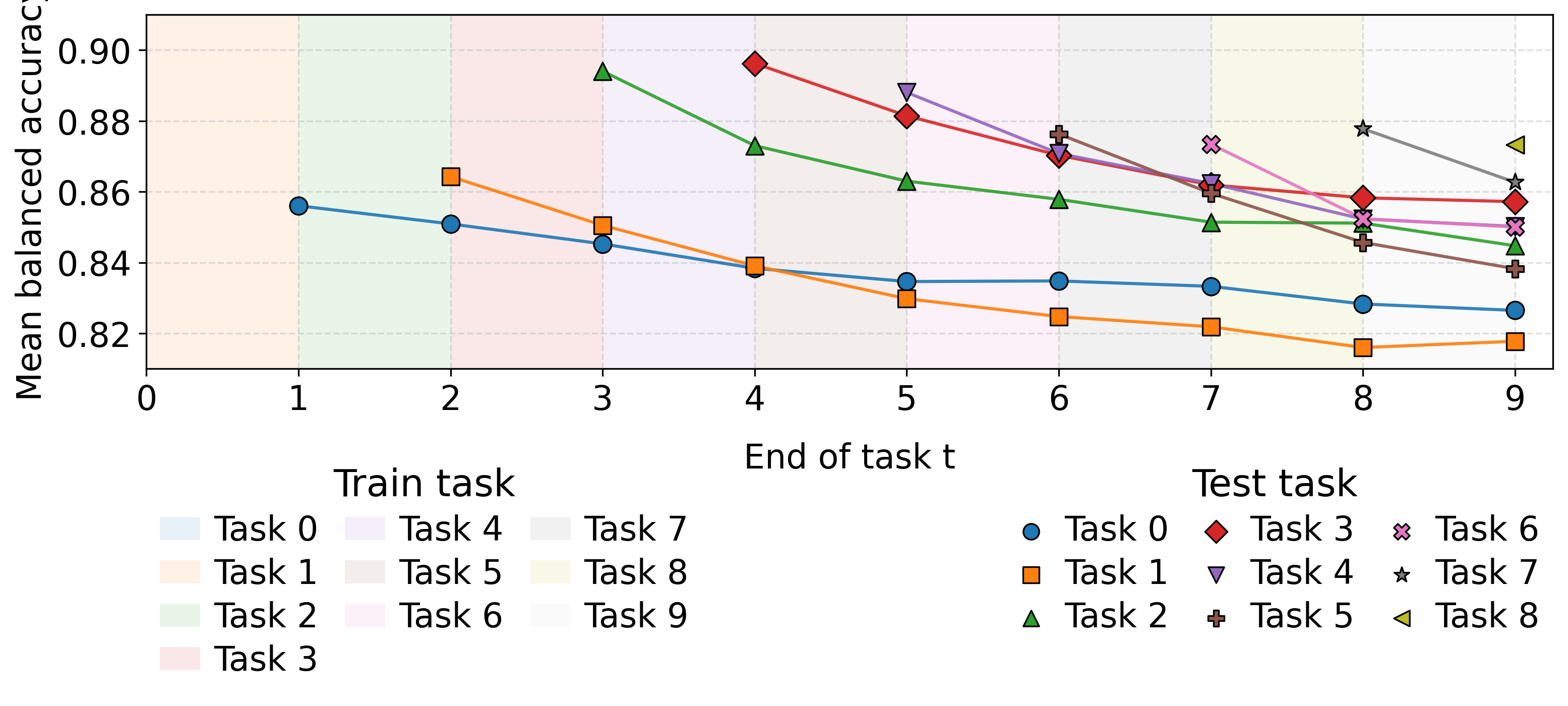}
        \label{fig:mean_bacc_scatter_derpp}}
    \hfill
    \subfloat[DER++ - Mean F1 score]{
        \includegraphics[trim=0cm 4.02cm 0cm 0.1cm, clip, width=0.42\textwidth]{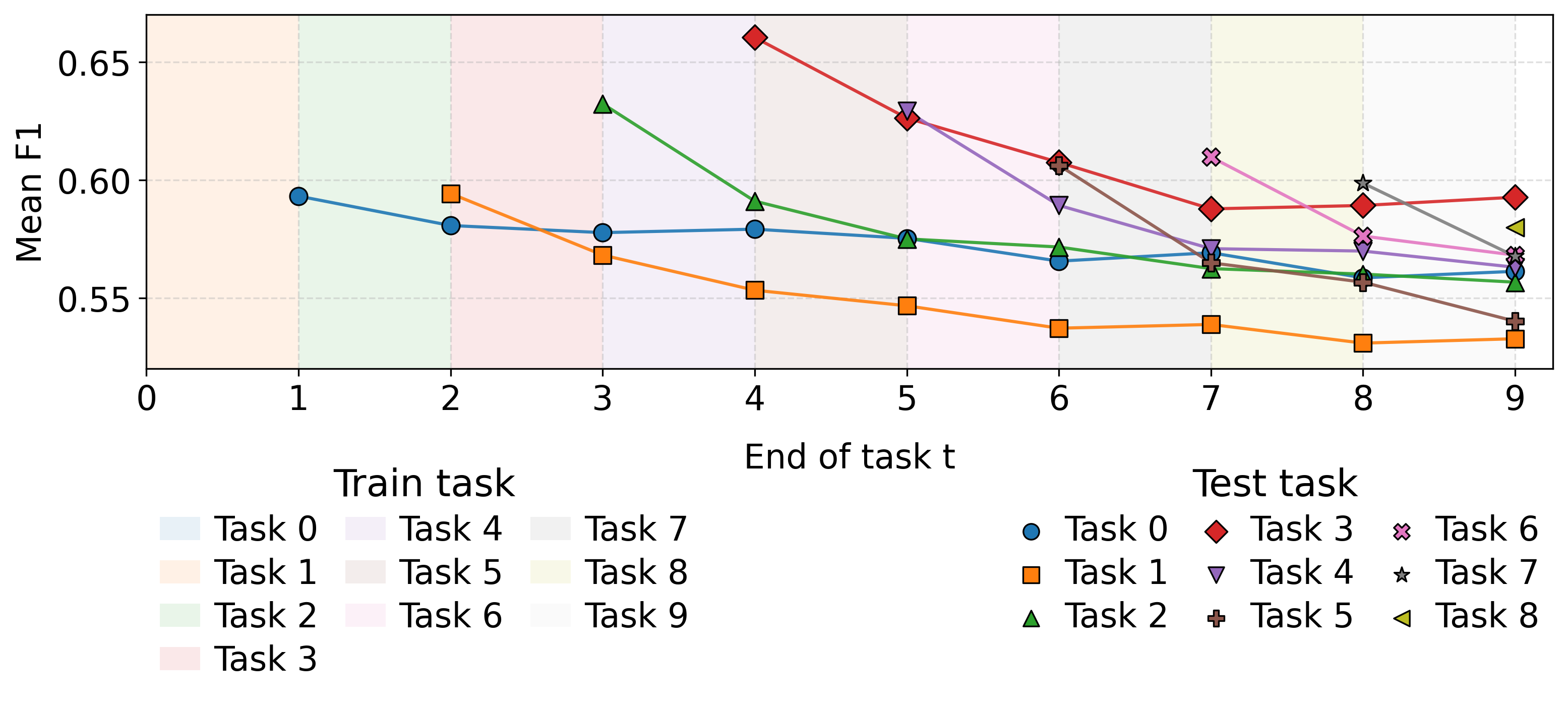}
        \label{fig:mean_f1_scatter_derpp}}
    \vspace{-3mm}

    \caption{\textbf{Mean balanced accuracy} and \textbf{mean F1 score} for the \textbf{concept prediction probe} on 10seq-tiny-ImageNet for all tasks throughout continual training. We use the same legend as in Fig. \ref{fig:deletion_scatter}.}\label{fig:mean_f1_bacc_scatter}
\end{figure}

Fig. \ref{fig:mean_f1_bacc_scatter} shows the evolution of mean balanced accuracy and F1 of concept predictions throughout all future continual tasks for all tasks. These plots also show significant differences: for SGD and DER++ both balanced accuracy and F1 drop with subsequent training tasks. For LwF it is almost constant (only some small drops appear) showing that this forgetting-aware strategy preserves task information also at the more fine-grained level of concepts. As the ground truth labels are constant (task t data at t determines them), this is true decodability  degradation caused by possible information loss or increasing readout issues at the level of individual concepts. Our analysis did not identify fully \textbf{lost} neurons at the strictest possible criterion, i.e., $\mathrm{F1}=0$, suggesting that concept-level information was at most partially lost rather than entirely erased in our setup. As a qualitative complement, in Appendix \ref{app:concept_viz} we compare the top activating images of selected \textbf{decodable} concepts (see Figs. \ref{fig:neuron_14_cifar_examples_decodable}, \ref{fig:neuron_677_cifar_examples_deleted_decodable}). Concepts with varying decodability (F1, balanced accuracy) still share motifs across $t$ and $t+1$, supporting that at least some part of concept-level information can remain decodable.

\subsection{Task prediction via linear probes}

\begin{figure}[th]
    \centering
    \subfloat[Representation-level]{
        \includegraphics[width=0.42\textwidth]{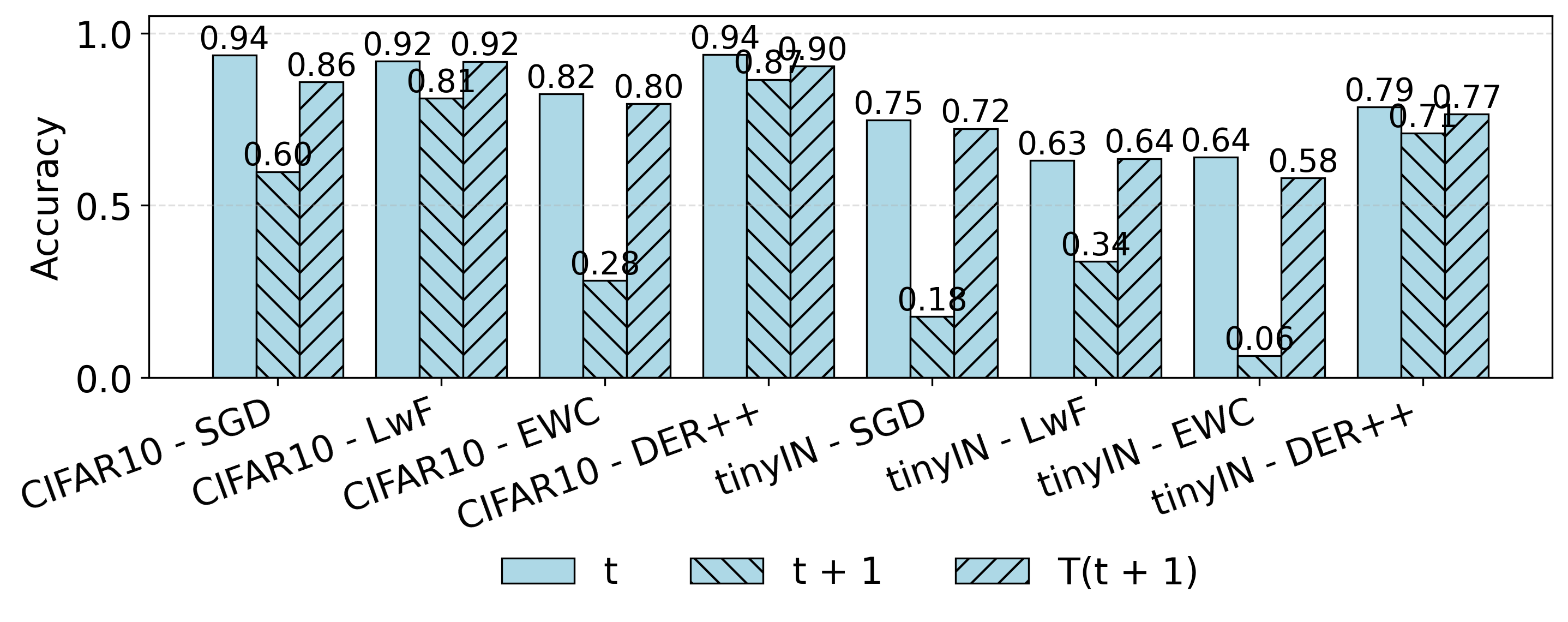}
        \label{fig:accuracy_raw_bar}}
    \hfill
    \subfloat[Concept-level]{
        \includegraphics[width=0.42\textwidth]{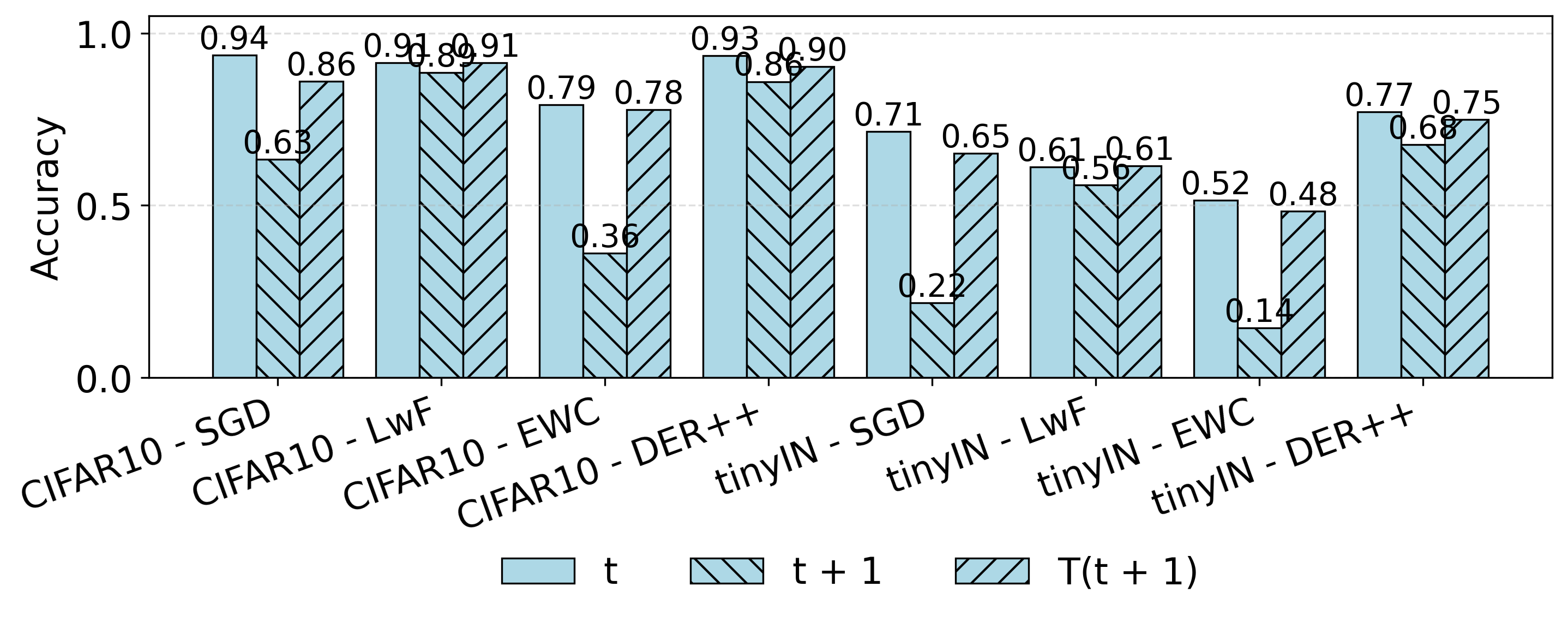}
        \label{fig:accuracy_sae_bar}}
    \caption{\textbf{Accuracy of probes} predicting 2seq-CIFAR10 and 2seq-tiny-ImageNet task 0 classes from the continual model features (representation-) and SAE latent activations (concept-level).}\label{fig:accuracy_probe_bar}
\end{figure}

Fig.~\ref{fig:accuracy_probe_bar} reports task 0 accuracy for 2seq-CIFAR10 and 2seq-tiny-ImageNet, using probes trained on continual model features and SAE latents. As expected, the largest drops at $t+1$ are observed for SGD and EWC. Linear translation restores most performance. However, incomplete recovery for all strategies except LwF indicates that degradation cannot be explained solely by linear misalignment, pointing to residual readability limitations and partial loss of task information. Importantly, the recovery of accuracy for latent-based probes suggests that task-relevant information remains largely compatible with the original task-anchored concept basis after alignment, and thus provides functional evidence that the translation restores access to fine-grained information at the level of the analyzed concept proxies. This further indicates that a large portion of fine-grained information is not fully deleted due to forgetting, but becomes misaligned and less directly accessible. The remaining performance gap after alignment implies, however, partial degradation of concept-level structure, consistent with the observed variability in F1 scores for concept prediction. 

To further understand the temporal dynamics of probe performance, we provide an extended analysis across all tasks~$t$ after ~$t+s$ of 10seq-tiny-ImageNet in Appendix~\ref{app:linear_probe_raw} (Fig. \ref{fig:scatter_accuracy_raw} -- continual model features-based probes, Fig.~\ref{fig:scatter_accuracy_sae} -- concept-based probes). Accuracy at $t+1$ without translation declines as~$s$ increases for all strategies, with the strongest effect on SGD and EWC. Operating on SAE latents seems to stabilize the probe accuracies on non-translated data for LwF and DER++. 
Translation recovers much of the performance at all $s$, thus information recoverability persists over time. However, it slightly decreases as~$s$ grows for some strategies (SGD, DER++), further suggesting that stronger forgetting may involve more advanced drifts or partial information erasure.


\section{Conclusion}\label{sec:conclusions}

In this paper, we proposed an SAE-based framework that defines a fixed, more disentangled concept space to analyze forgetting in continual learning through task-specific knowledge preservation and recoverability under a linear readout assumption. Our results show that forgetting mitigation strategies affect not only the magnitude but also the nature of forgetting. Methods such as LwF keep representational drift largely linear, making class- and concept-level information more decodable and reversible. In contrast, other strategies induce less recoverable distortions. Overall, in the studied settings, much apparent forgetting reflects approximately linear representational changes, with seemingly deleted information often recoverable through linear remapping. Importantly, our results suggest that knowledge preservation extends beyond coarse-grained class information to a more fine-grained level. At the same time, our results indicate that some part of the concept-level information becomes no longer fully accessible in a linear setting. Under the common assumption of a linear readout as a classifier, this leads to a practical loss of decodability, which can be functionally equivalent to forgetting. Thanks to the observation of the distributions of concept-prediction scores we find that concept-level forgetting is not uniform. At the same time, our results suggest that continual learning strategies preserve past knowledge at different levels: some retain their functionality more at the level of outputs (DER++), whereas others better preserve it at the level of the internal organization of concept proxies (LwF, EWC). Overall, our findings suggest that, in supervised continual learning, past fine-grained structure of model knowledge and computations are often not lost but hidden: they may remain linearly recoverable, although their preservation depends on the continual learning strategy. This emphasizes the importance of preserving such structures for interpretable continual learning. As improving transparency and trustworthiness of continual models is a positive societal impact, future work will extend our framework to broader settings and problems, as well as design strategies that better preserve fine-grained task-relevant information.

\paragraph{Limitations.} One may view the use of SAE latent features in a concept-level analysis as a limitation, since they are concept proxies rather than ground-truth semantic entities. However, they provide an automatic, label-free, and computationally feasible approximation of more disentangled, fine-grained structure in the continual model’s representation space, supported by strong task-level concept-based probe performance, quantitative quality assessment in App.~\ref{app:ms_analysis}, and qualitative examples in App. \ref{app:concept_viz} showing clear shared motifs. A limitation of our taxonomy is that the boundary for lost concepts is not absolute: very low $\mathrm{F1}$ may indicate only marginal recoverability. For clarity, we use $\mathrm{F1}=0$ to define \emph{Lost}, which represents a clear non-parametric case of zero decodability. However, we emphasize in the paper that residual concept degradation can still be functionally equivalent to partial concept loss under a linear readout.
Finally, a potential negative societal impact is that concept-level analyses may increase confidence in deploying continual models in high-stakes settings, even though the identified concepts are proxies and may miss sensitive or safety-critical information.

\bibliography{bibliography}
\bibliographystyle{abbrv}

\newpage

\appendix

\newpage

\setcounter{figure}{0}
\counterwithin{figure}{section} 
\setcounter{table}{0}
\counterwithin{table}{section} 

\section{Concept definition}\label{app:concept}

\begin{figure*}[h]
    \centering
                \includegraphics[width=0.7\textwidth]{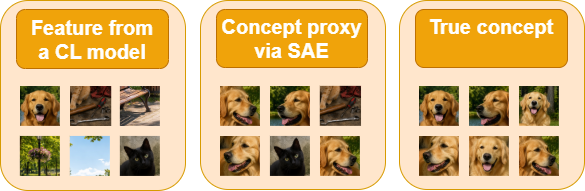}
    \caption{SAE latents provide cleaner, more concept-aligned representations than raw model features, and are therefore used here as proxies for concepts. i) Feature from a CL model corresponds to a single neuron in the representation layer in CL model; ii) SAE allows us to build a concept proxy, disentangling features; iii) ground truth semantic concept in real-world}
    \label{fig:conceps_proxy}
\end{figure*}

In this work, we understood a \textbf{concept} as a computational representation of a shared, recurring pattern that helps a model structure information. Methods from mechanistic interpretability such as Sparse Autoencoders (SAEs) can extract such representations via self-supervised learning, often providing more coherent patterns than standard continual models. Analogous to human perception, in which recognizing an object (e.g. a dog) relies on features like ears, a nose, or paws, model concepts capture recurring patterns in data. While they do not fully align with human semantic concepts and remain approximate (see Fig. \ref{fig:conceps_proxy}), potentially less coherent representations, they provide a practical and well-established way to model abstract, fine-grained information, serving as useful \textbf{concept proxies}, to which we refer to as \textbf{concepts}. Studying whether these concepts persist in continual learning is equally as important as accuracy, as it reveals whether the model truly retains knowledge and preserves a consistent way of organizing information and computations, rather than merely sustaining performance through shifting internal representations. These all can be also connected to whether the model maintains stable and trustworthy interpretations, which is crucial for explainability and trustworthiness in artificial intelligence.

\section{Accuracy of the original continual models' classification heads}\label{app:accuracy_continual}

Table \ref{tab:cifar_taskil} presents the accuracy results for both  tasks for the 2seq-CIFAR10 case. After learning the second task, performance on the first task drops significantly for SGD, and notably less for DER++ and LwF, while accuracy on the new task remains similarly high for SGD, LwF and DER++ (96.4 for SGD, 96.5 for DER++, 95.3 for LwF), and slightly lower for EWC (89). This suggests that on CIFAR10, all methods learn the new task well, but forgetting mitigation mainly affects retention of the old one, with LwF preserving past-task accuracy best, DER++ offering an intermediate preservation, and SGD, EWC forgetting the most.

\begin{table}[ht]
\centering
\caption{Task-IL accuracy (\%) on 2seq-CIFAR10. For task 0, we report the accuracy after learning task 0. For task 1, we report the accuracies on task 0 and task 1 after learning task 1.}
\label{tab:cifar_taskil}
\begin{tabular}{lccc}
\toprule
\textbf{Method} & \textbf{Task 0 after $t{=}0$} & \textbf{Task 0 after $t{=}1$} & \textbf{Task 1 after $t{=}1$} \\
\midrule
SGD    & 93.30 & 66.94 & 96.36 \\
DER++  & 93.42 & 85.58 & 96.48 \\
LwF    & 91.36 & 90.82 & 95.32 \\
EWC    & 72.34 & 46.08 & 88.96 \\
\bottomrule
\end{tabular}
\end{table}

\begin{figure}[h]
    \centering
    \subfloat[SGD]{
        \includegraphics[width=0.42\textwidth]{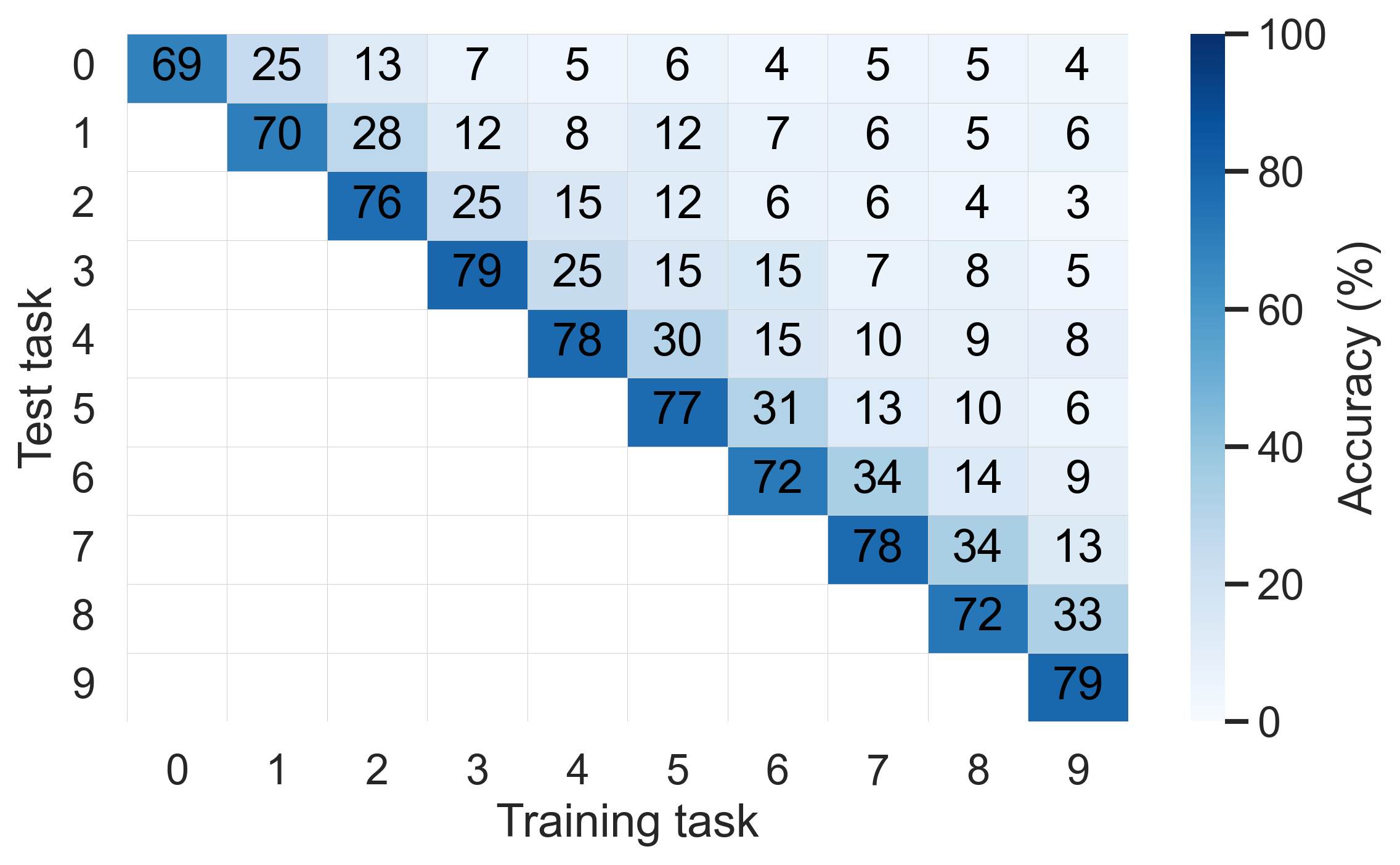}
        \label{fig:acc_tiny_sgd}}
    \hfill
    \subfloat[DER++]{
        \includegraphics[width=0.42\textwidth]{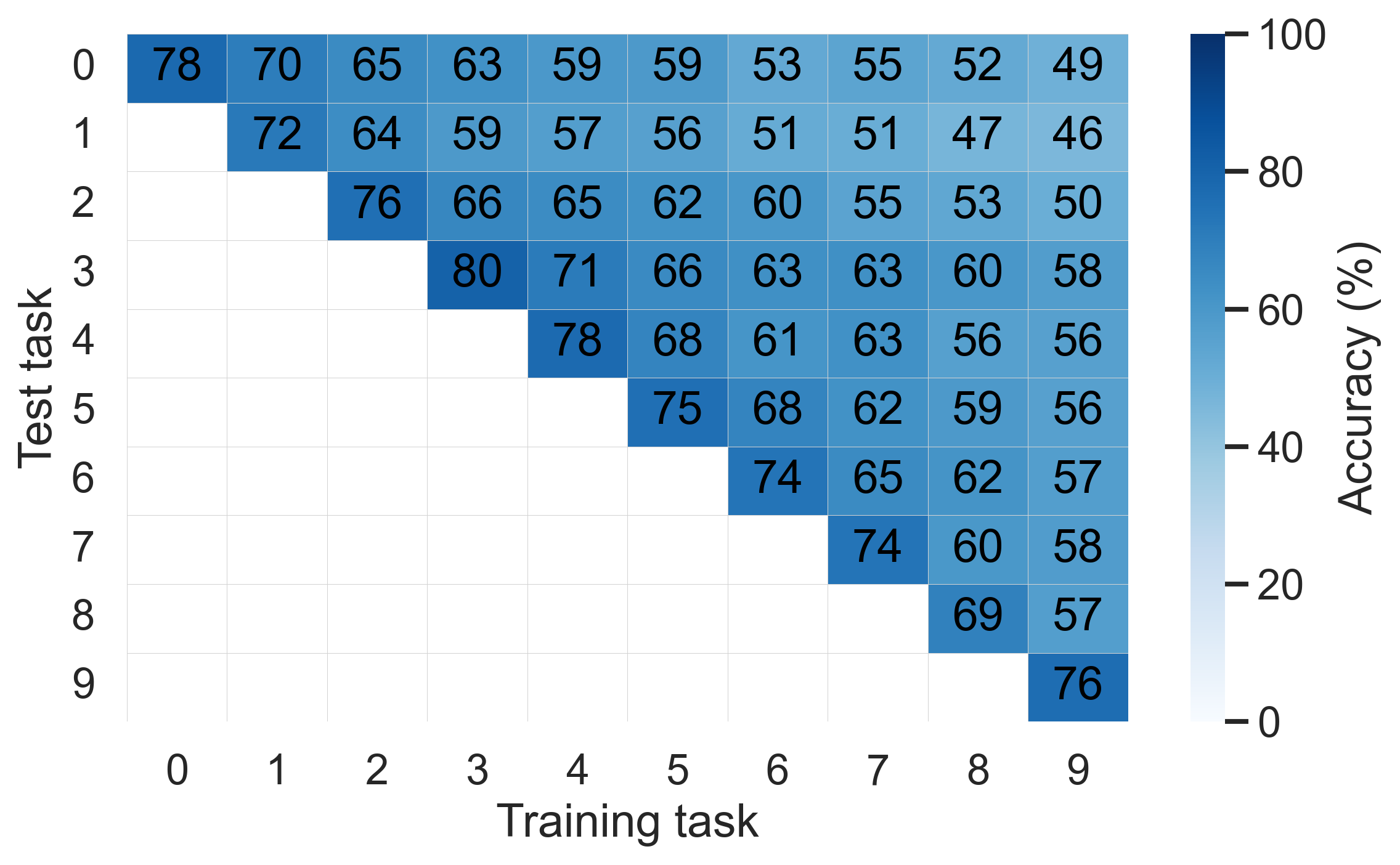}
        \label{fig:acc_tiny_derpp}}
    
    \subfloat[LwF]{
        \includegraphics[width=0.42\textwidth]{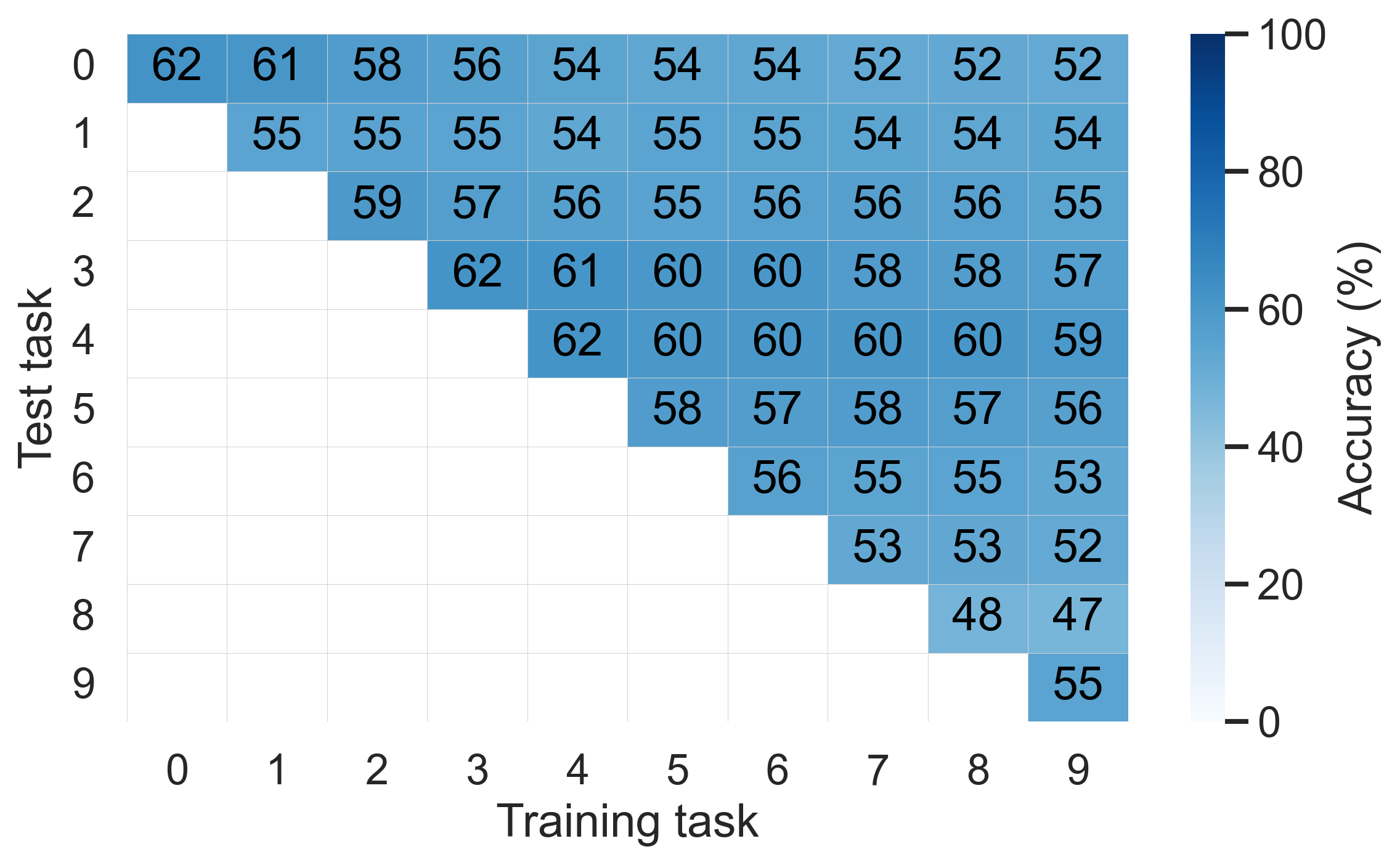}
        \label{fig:acc_tiny_lwf}}
    \hfill
    \subfloat[EWC]{
        \includegraphics[width=0.42\textwidth]{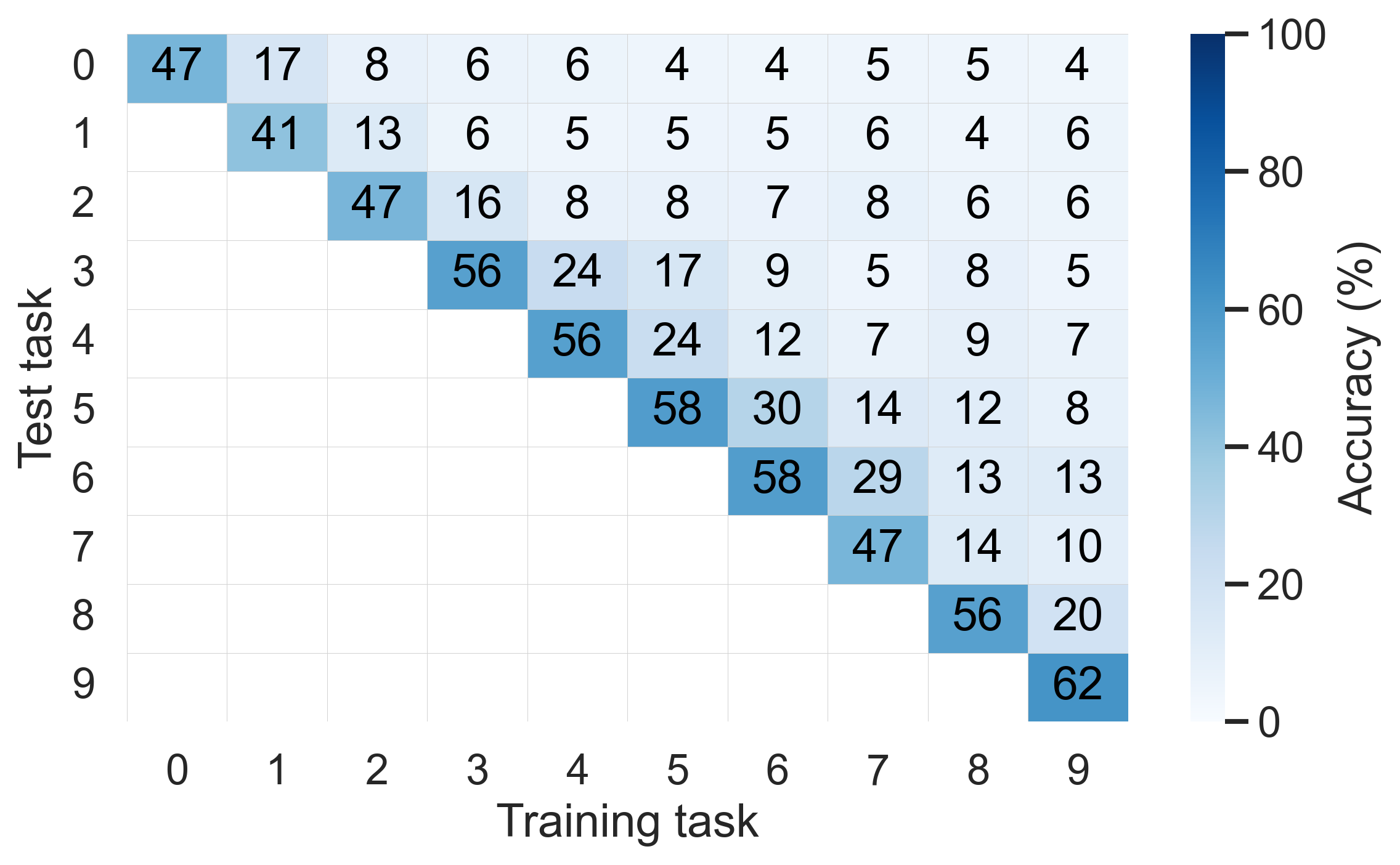}
        \label{fig:acc_tiny_ewc}}
    \caption{Accuracy of the original continual models' classification heads in the task incremental learning setup on 10seq-tiny-ImageNet.}\label{fig:accuracy_matrix}
\end{figure}

We present the accuracy matrices for all tasks under the Task-IL setting for 10seq-tiny-ImageNet in Fig. \ref{fig:accuracy_matrix}. The four examined strategies resulted in expected behaviors. SGD achieves reasonable accuracy on the currently learned task, but older-task performance drops sharply as training proceeds -- as an indication of strong forgetting. DER++ provides the best overall  Task-IL accuracy, with noticeably better retention of past tasks than SGD while still maintaining strong performance on newer tasks. In contrast, LwF is more uniform across tasks, but at a lower accuracy level, suggesting better stability at the cost of plasticity and overall per-task performance.

\FloatBarrier

\section{Evolution  of task-level linear probes performance when trained on raw features and latent features}\label{app:linear_probe_raw}

\begin{figure}[h]
    \centering
    \subfloat[SGD - Raw features]{
        \includegraphics[trim=0cm 4.02cm 0cm 0.1cm, clip, width=0.42\textwidth]{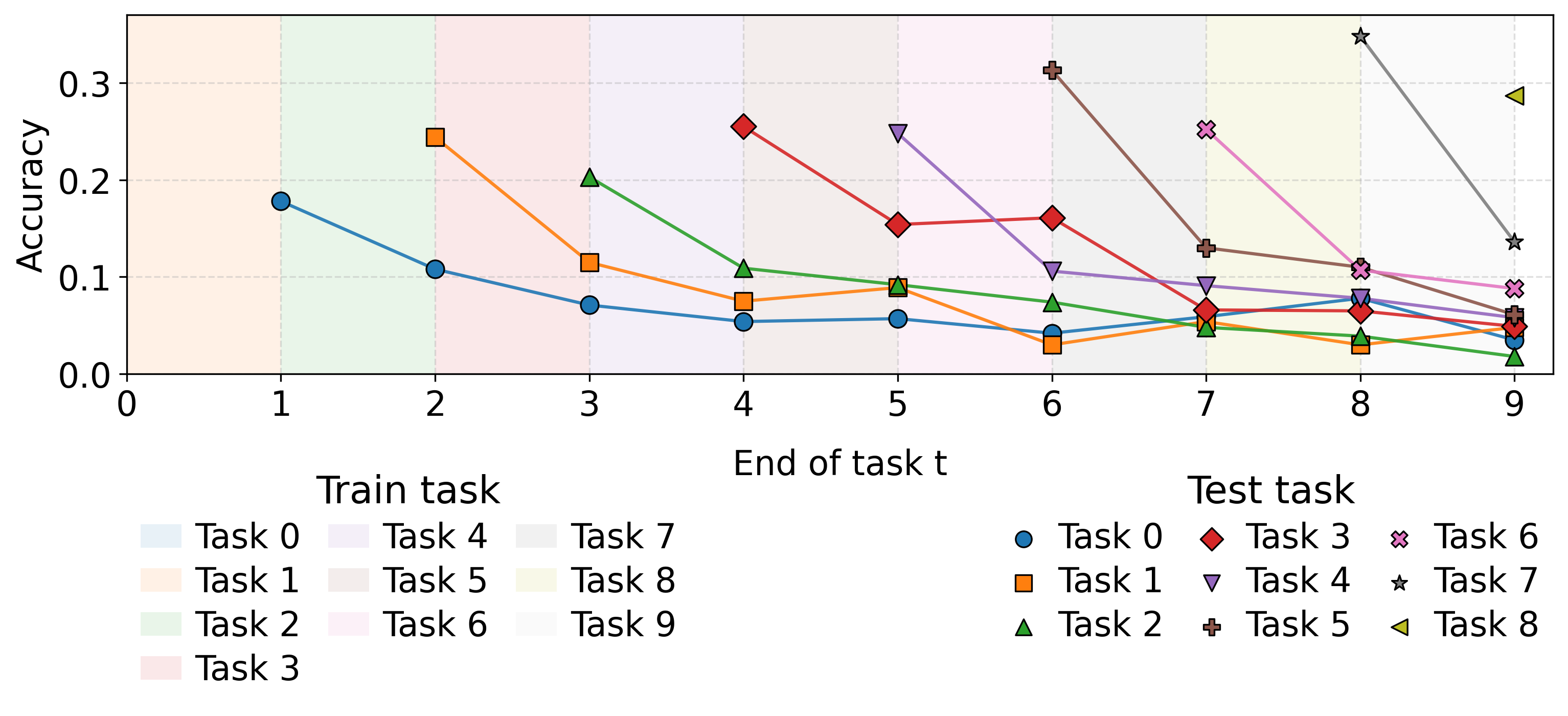}
        \label{fig:accuracy_raw_scatter_sgd}}
    \hfill
    \subfloat[SGD - Linear translation]{
        \includegraphics[trim=0cm 4.02cm 0cm 0.1cm, clip, width=0.42\textwidth]{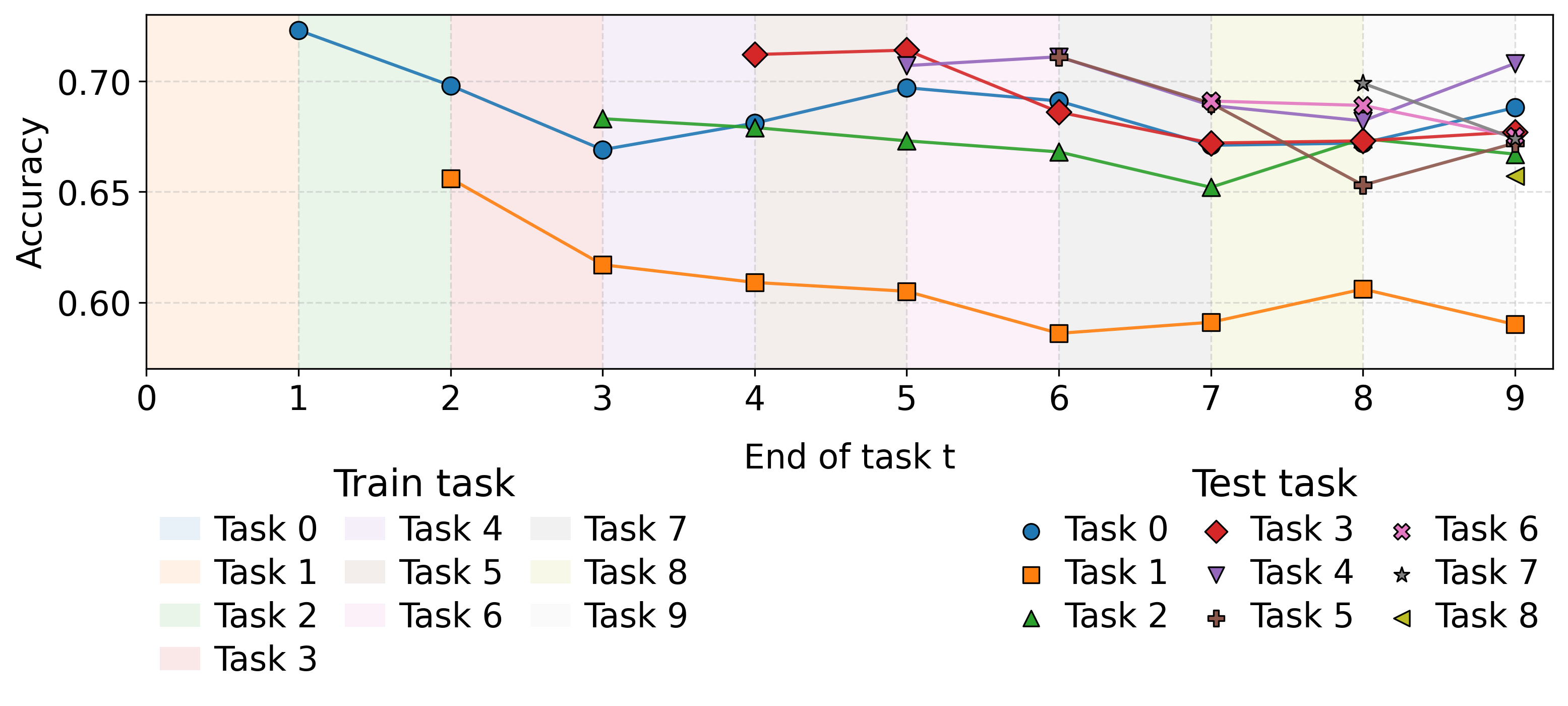}
        \label{fig:accuracy_scatter_linear_sgd}}

        \subfloat[LwF - Raw features]{
        \includegraphics[trim=0cm 4.02cm 0cm 0.1cm, clip, width=0.42\textwidth]{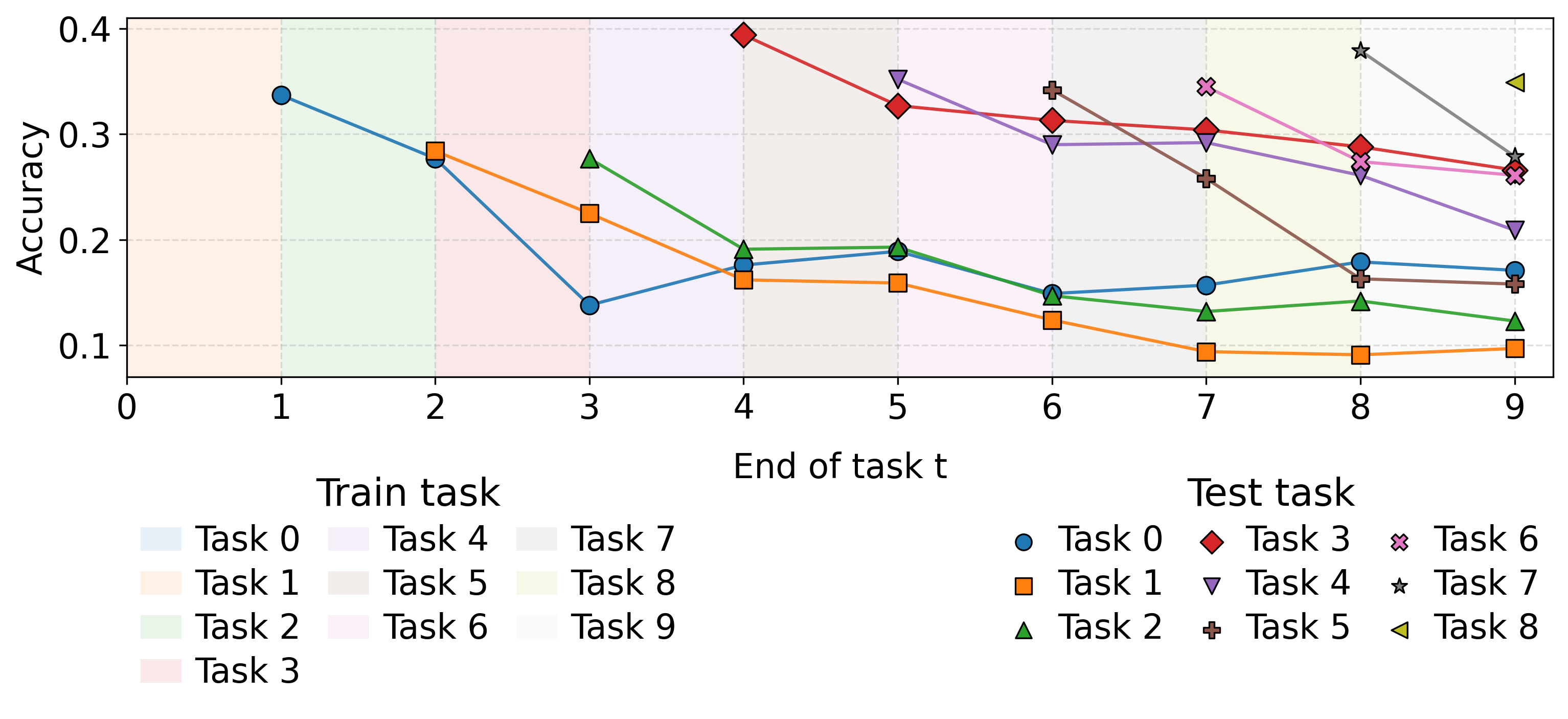}
        \label{fig:accuracy_raw_scatter_lfw}}
    \hfill
    \subfloat[LwF - Linear translation]{
        \includegraphics[trim=0cm 4.02cm 0cm 0.1cm, clip, width=0.42\textwidth]{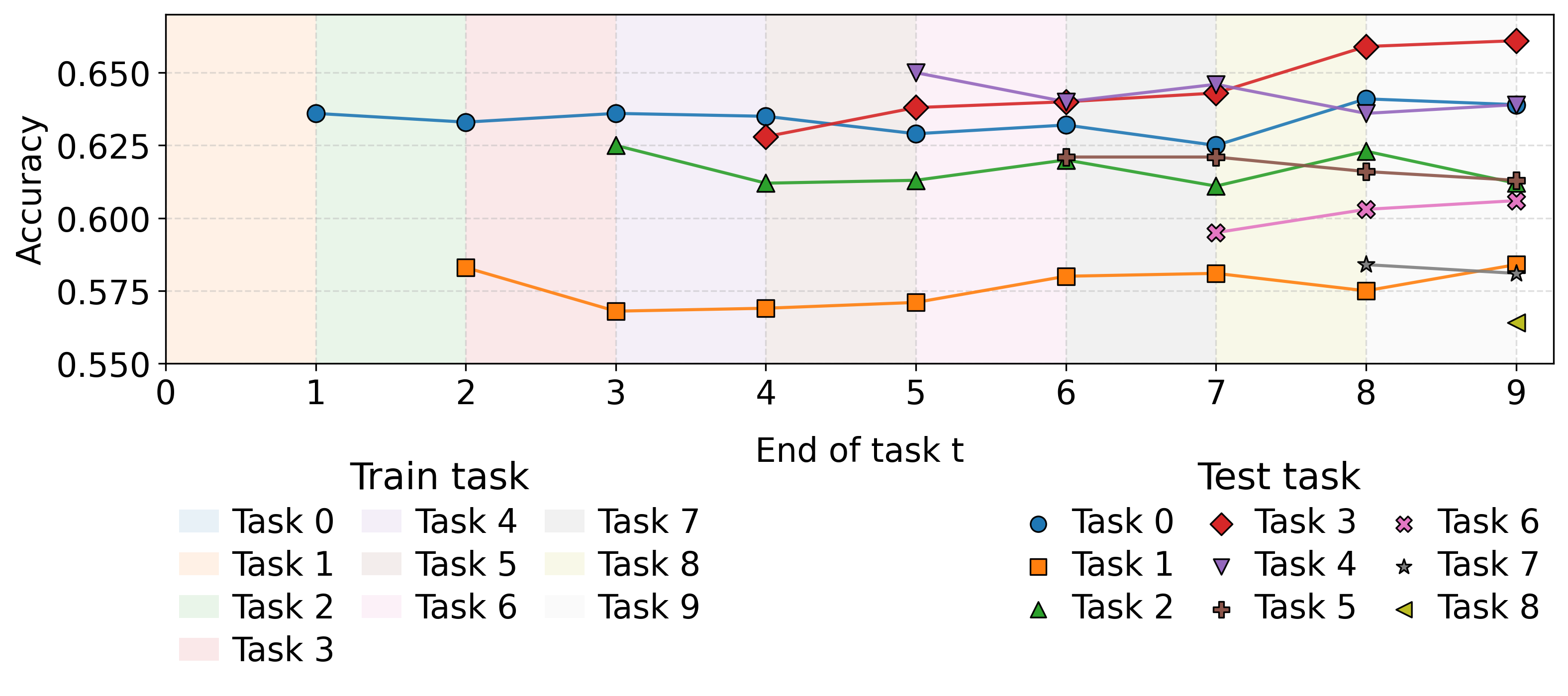}
        \label{fig:accuracy_scatter_linear_lfw}}

    \subfloat[EWC - Raw features]{
        \includegraphics[trim=0cm 4.02cm 0cm 0.1cm, clip, width=0.42\textwidth]{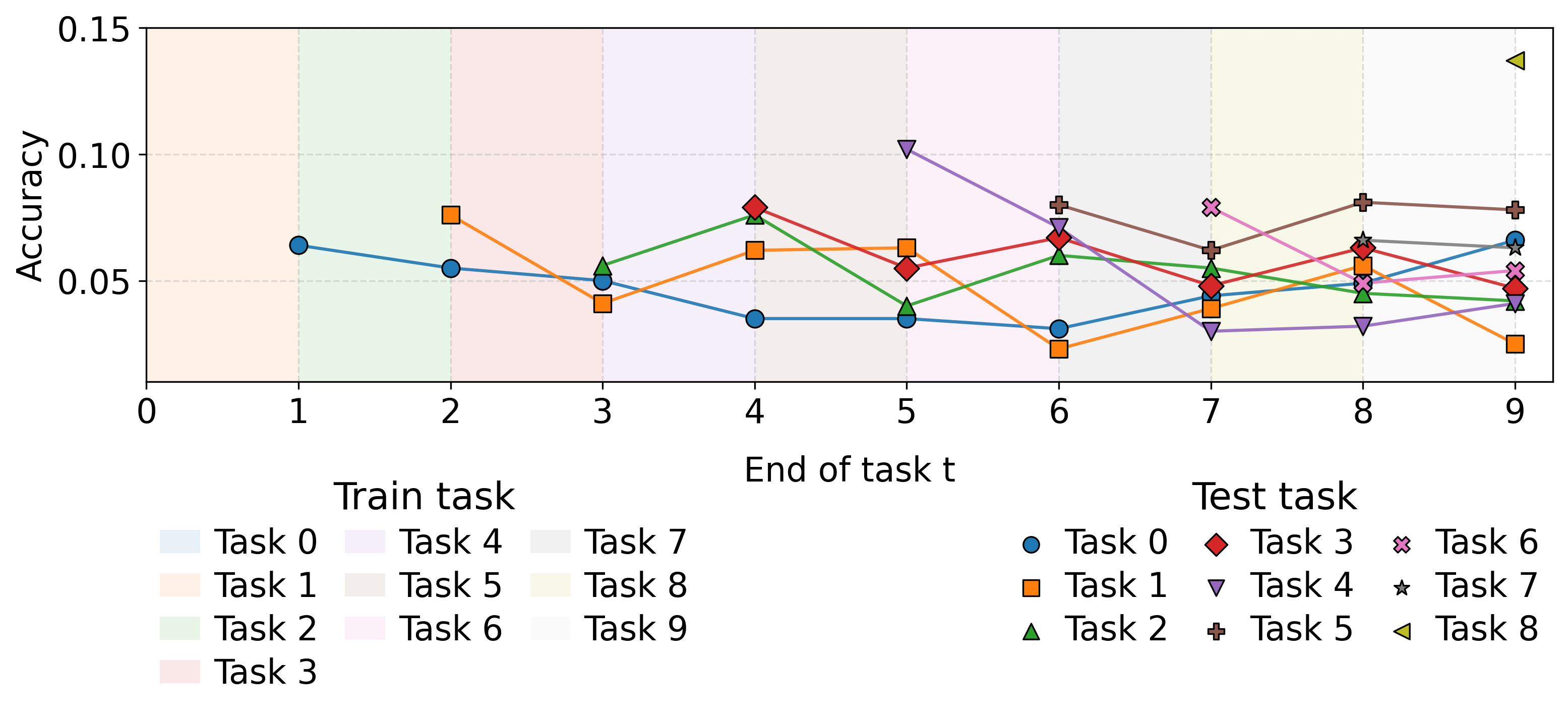}
        \label{fig:accuracy_raw_scatter_ewc}}
    \hfill
    \subfloat[EWC - Linear translation]{
        \includegraphics[trim=0cm 4.02cm 0cm 0.1cm, clip, width=0.42\textwidth]{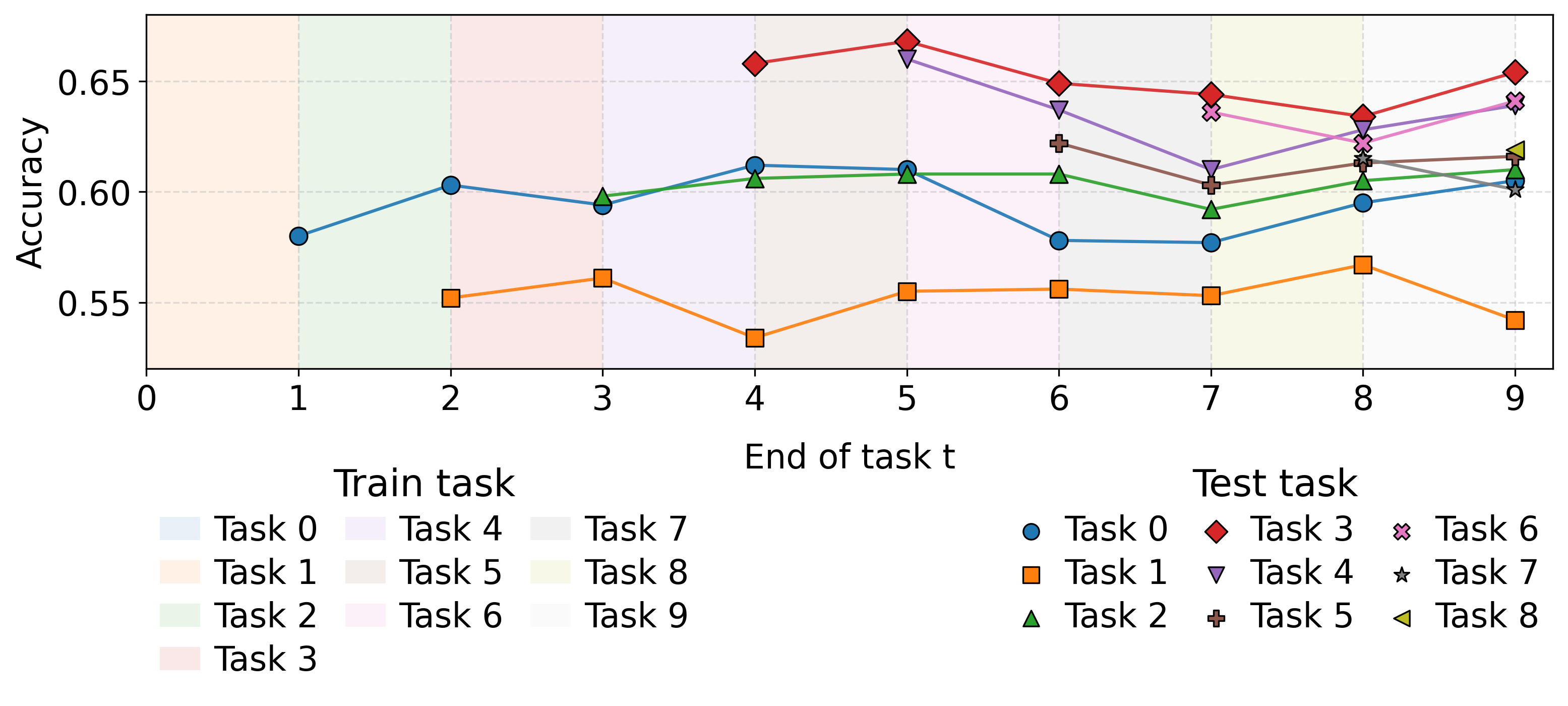}
        \label{fig:accuracy_scatter_linear_ewc}}

    \subfloat[DER++ - Raw features]{
        \includegraphics[trim=0cm 4.02cm 0cm 0.1cm, clip, width=0.42\textwidth]{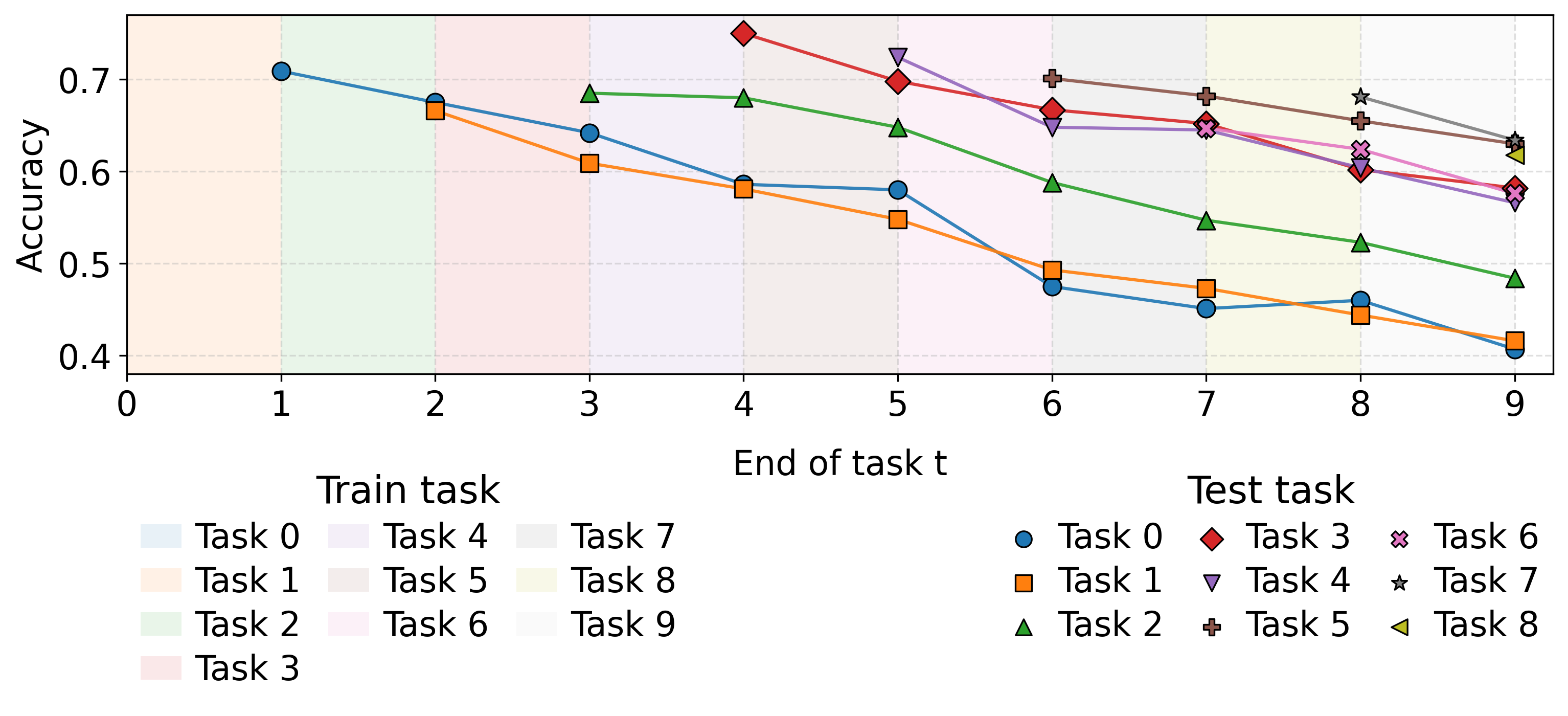}
        \label{fig:accuracy_raw_scatter_derpp}}
    \hfill
    \subfloat[DER++ - Linear translation]{
        \includegraphics[trim=0cm 4.02cm 0cm 0.1cm, clip, width=0.42\textwidth]{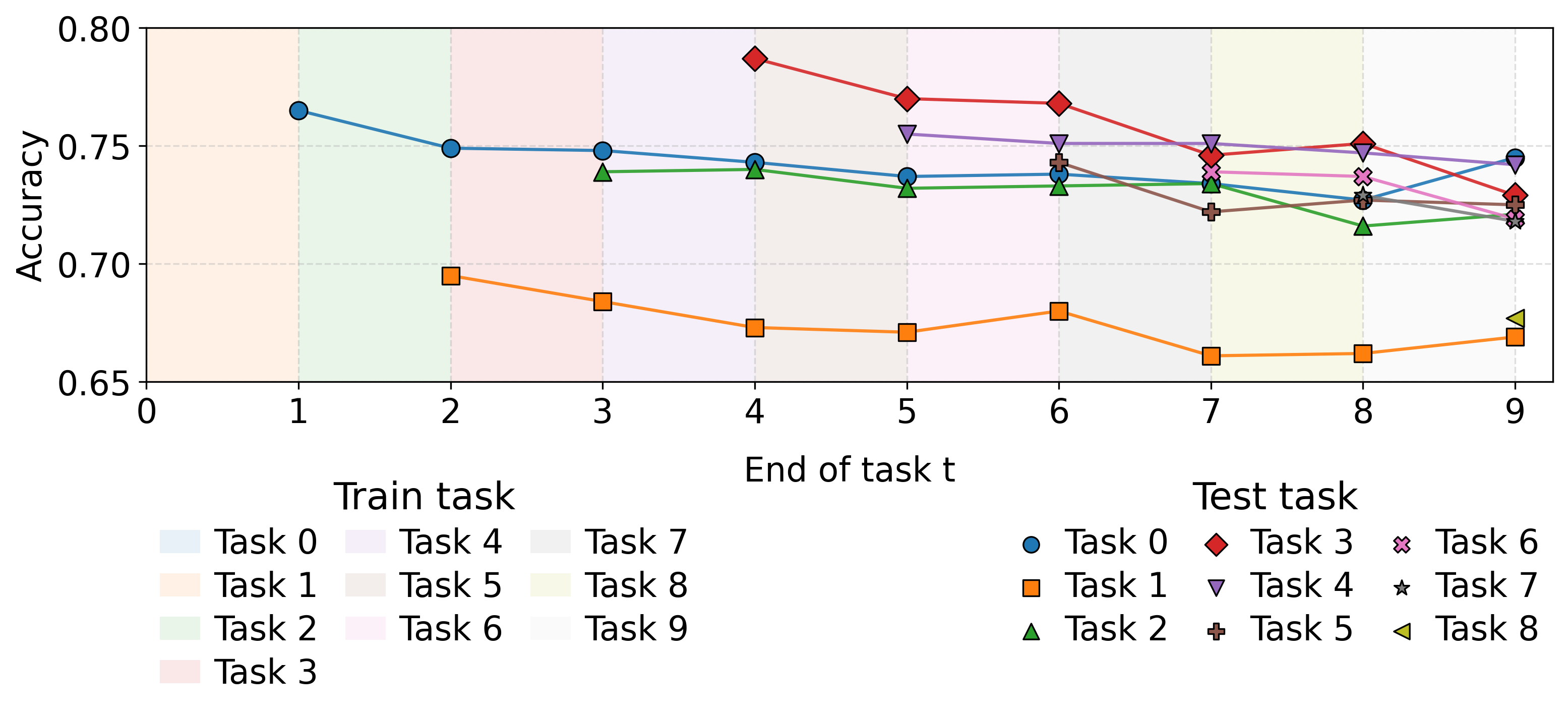}
        \label{fig:accuracy_scatter_linear_derpp}}

    \caption{\textbf{Accuracy of probes} for 10seq-tiny-ImageNet classes prediction throughout continual training based on the \textbf{raw representations} from the continual model (representation-level). Accuracy is higher in all cases after applying the linear translation to the raw features. We use the same legend as in Fig. \ref{fig:deletion_scatter}.}\label{fig:scatter_accuracy_raw}
\end{figure}

\begin{figure}[h]
    \centering
    \subfloat[SGD - Raw features]{
        \includegraphics[trim=0cm 4.02cm 0cm 0.1cm, clip, width=0.42\textwidth]{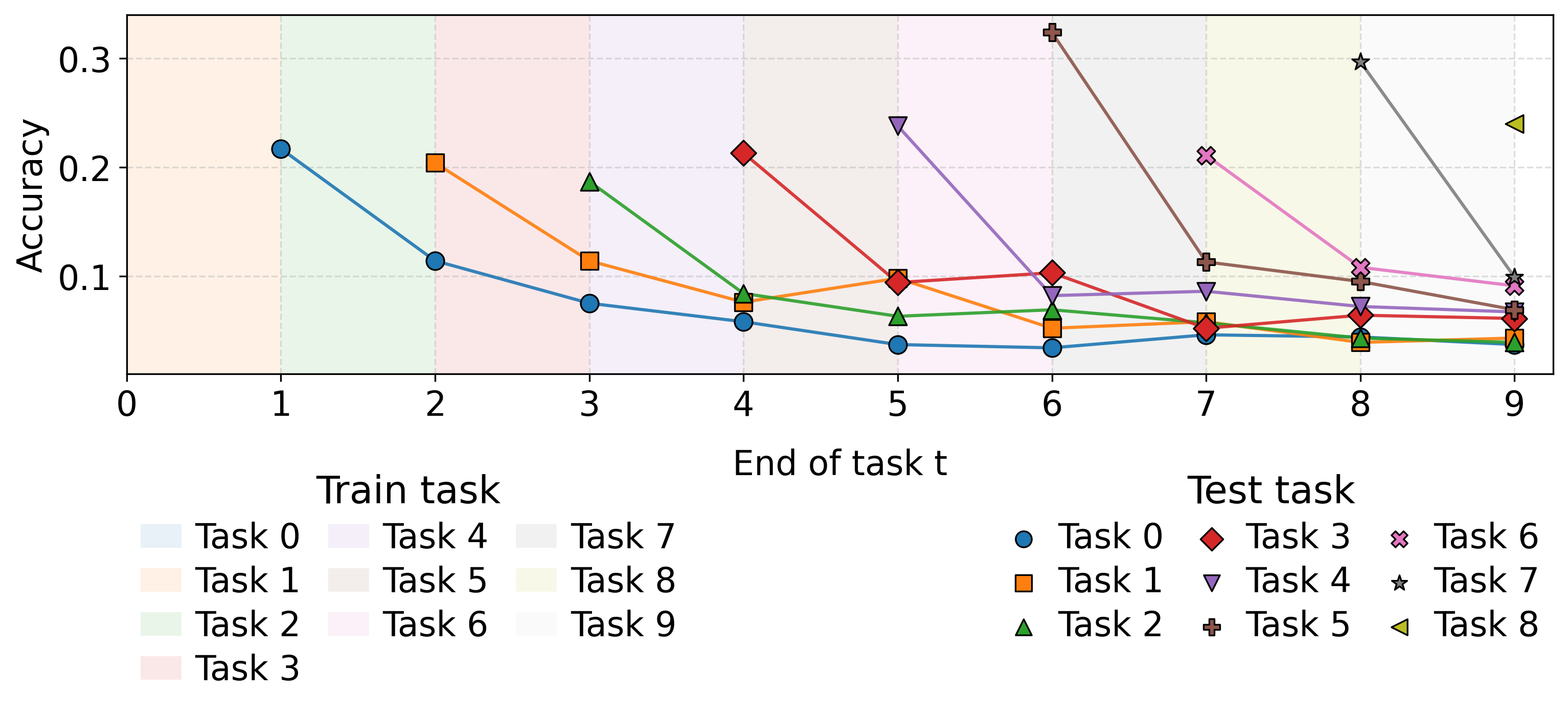}
        \label{fig:accuracy_sae_raw_sgd}}
    \hfill
    \subfloat[SGD - Linear translation]{
        \includegraphics[trim=0cm 4.02cm 0cm 0.1cm, clip, width=0.42\textwidth]{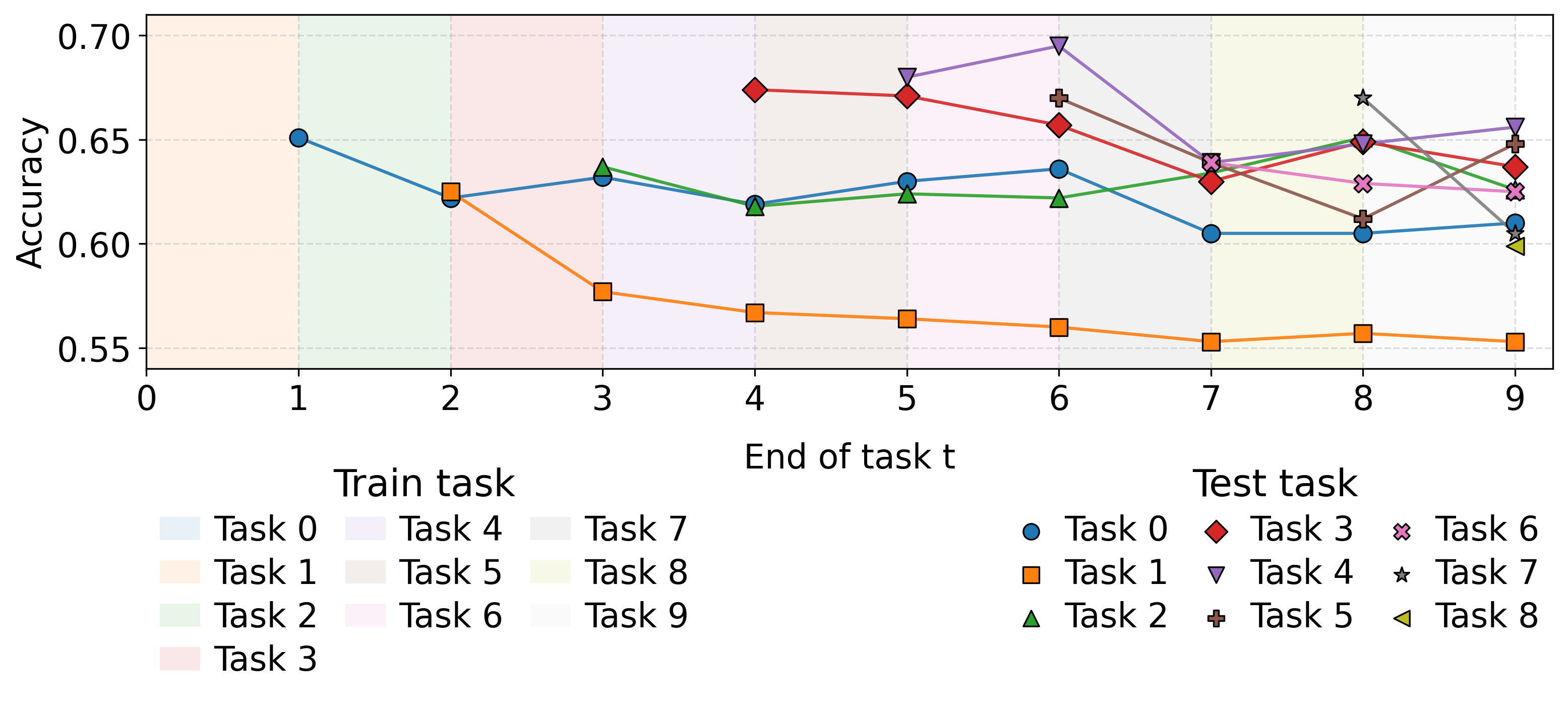}
        \label{fig:accuracy_sae_linear_sgd}}

    \subfloat[LwF - Raw features]{
        \includegraphics[trim=0cm 4.02cm 0cm 0.1cm, clip, width=0.42\textwidth]{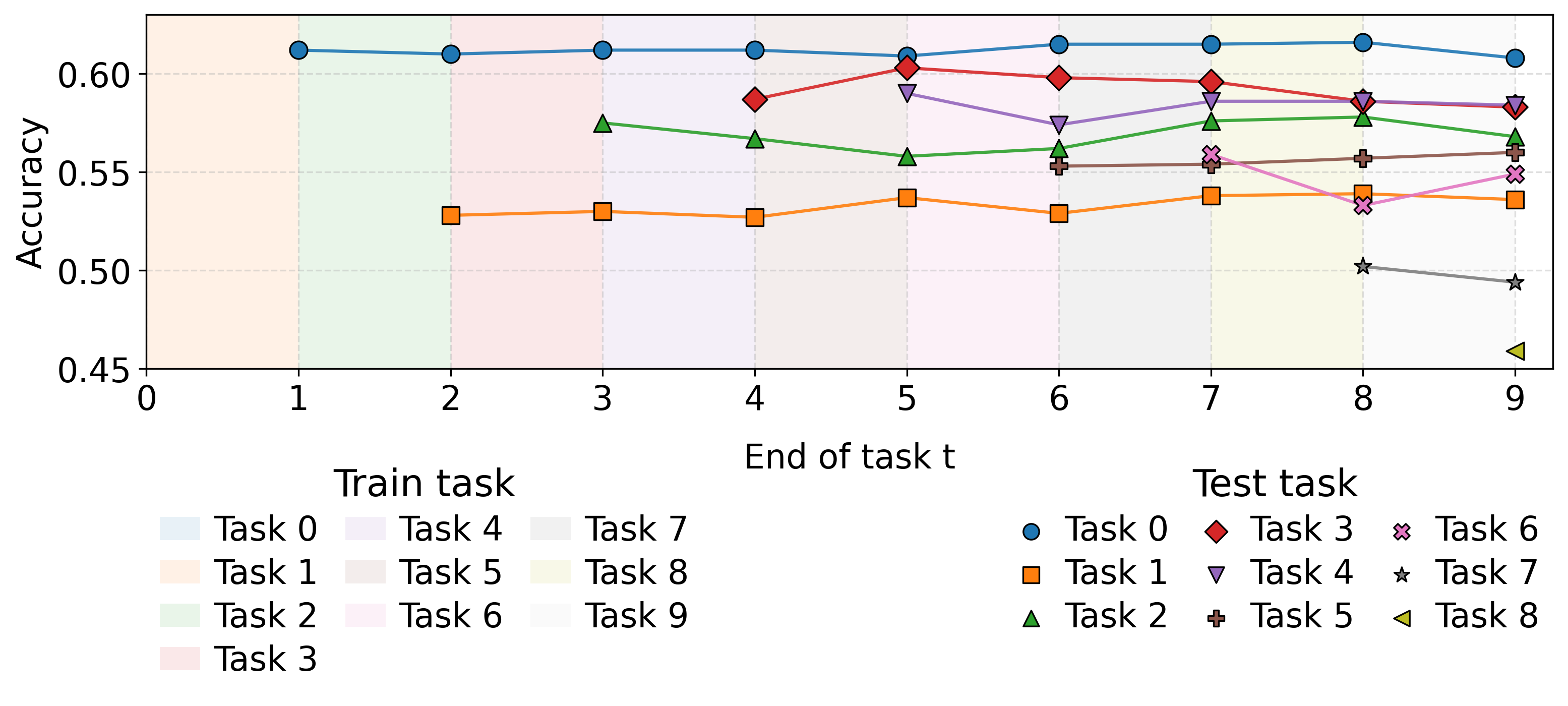}
        \label{fig:accuracy_sae_raw_lwf}}
    \hfill
    \subfloat[LwF - Linear translation]{
        \includegraphics[trim=0cm 4.02cm 0cm 0.1cm, clip, width=0.42\textwidth]{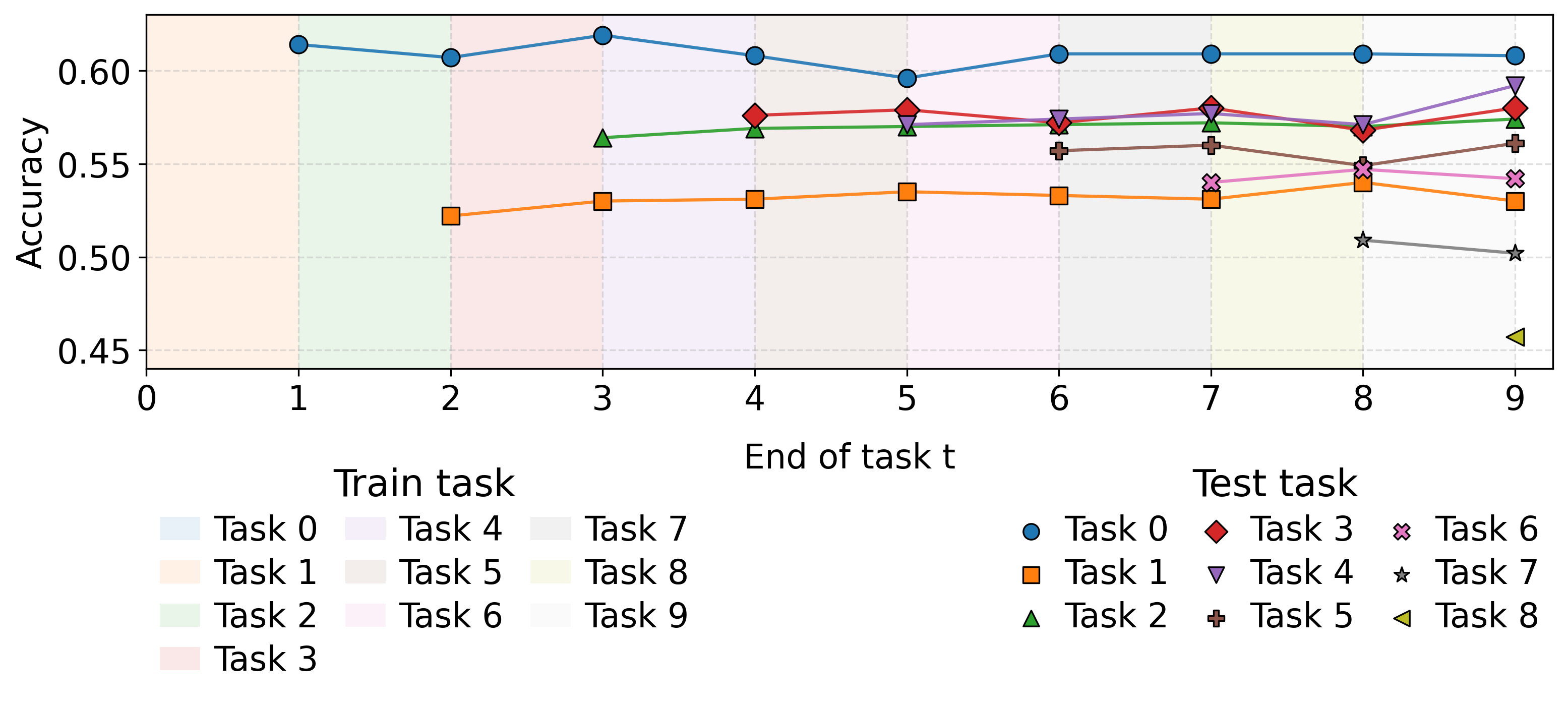}
        \label{fig:accuracy_sae_linear_lwf}}

    \subfloat[EWC - Raw features]{
        \includegraphics[trim=0cm 4.02cm 0cm 0.1cm, clip, width=0.42\textwidth]{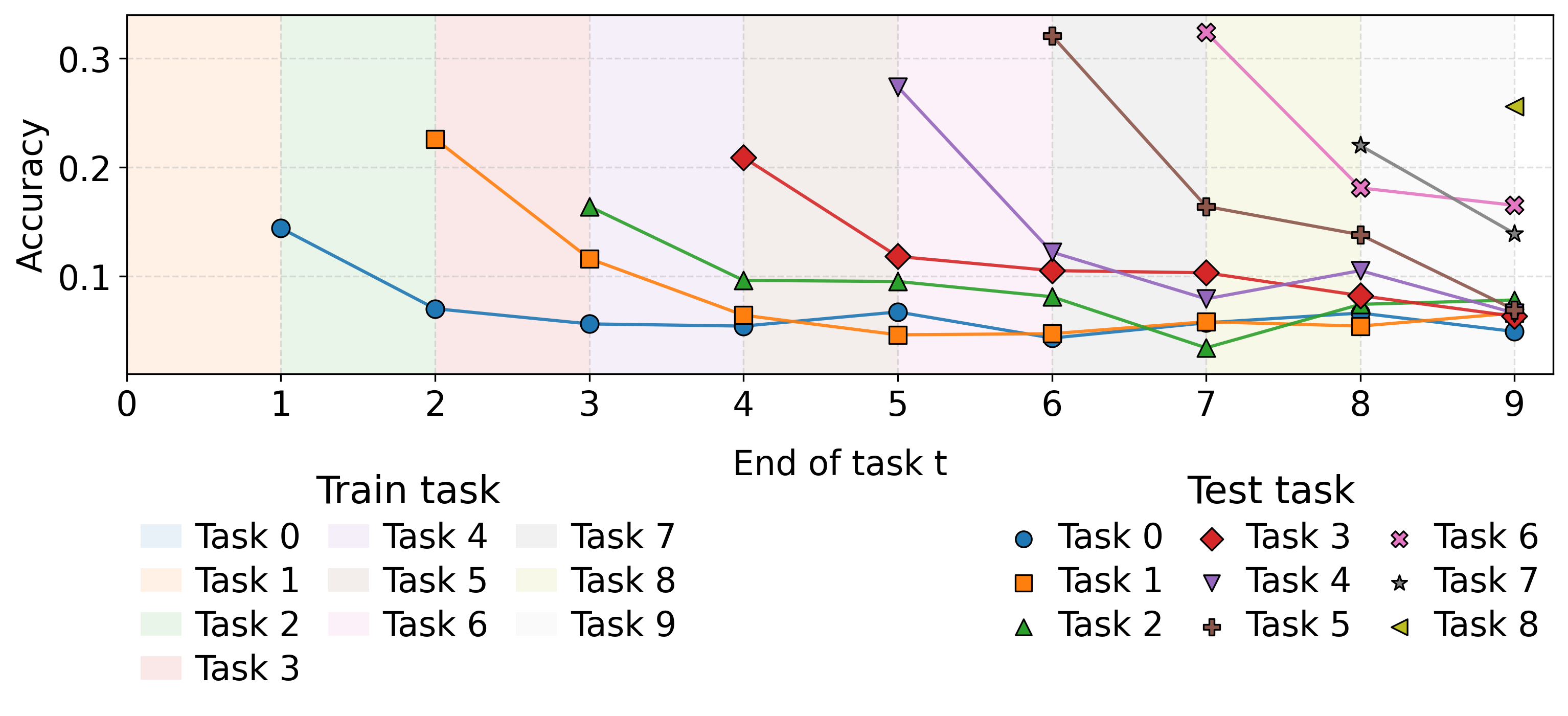}
        \label{fig:accuracy_sae_raw_ewc}}
    \hfill
    \subfloat[EWC - Linear translation]{
        \includegraphics[trim=0cm 4.02cm 0cm 0.1cm, clip, width=0.42\textwidth]{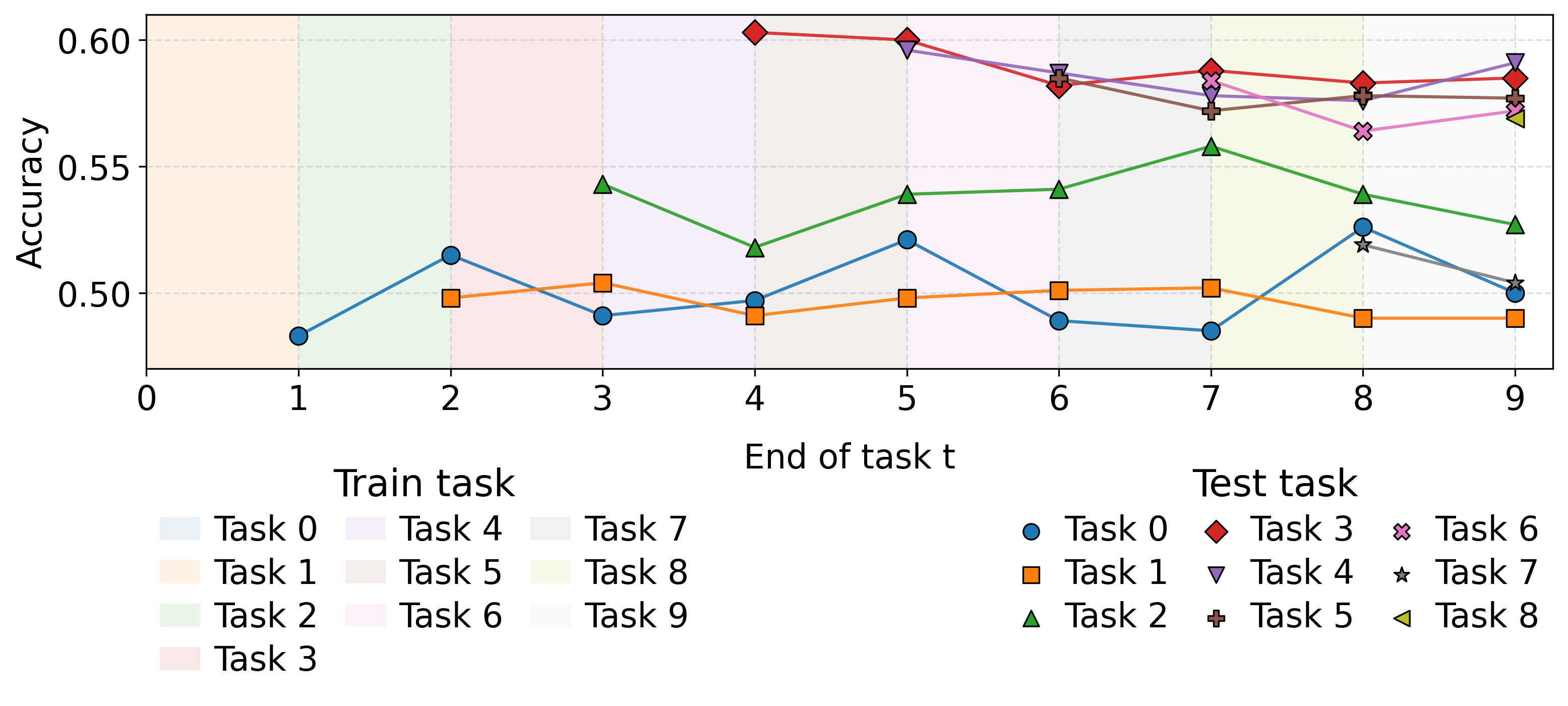}
        \label{fig:accuracy_sae_linear_ewc}}

    \subfloat[DER++ - Raw features]{
        \includegraphics[trim=0cm 4.02cm 0cm 0.1cm, clip, width=0.42\textwidth]{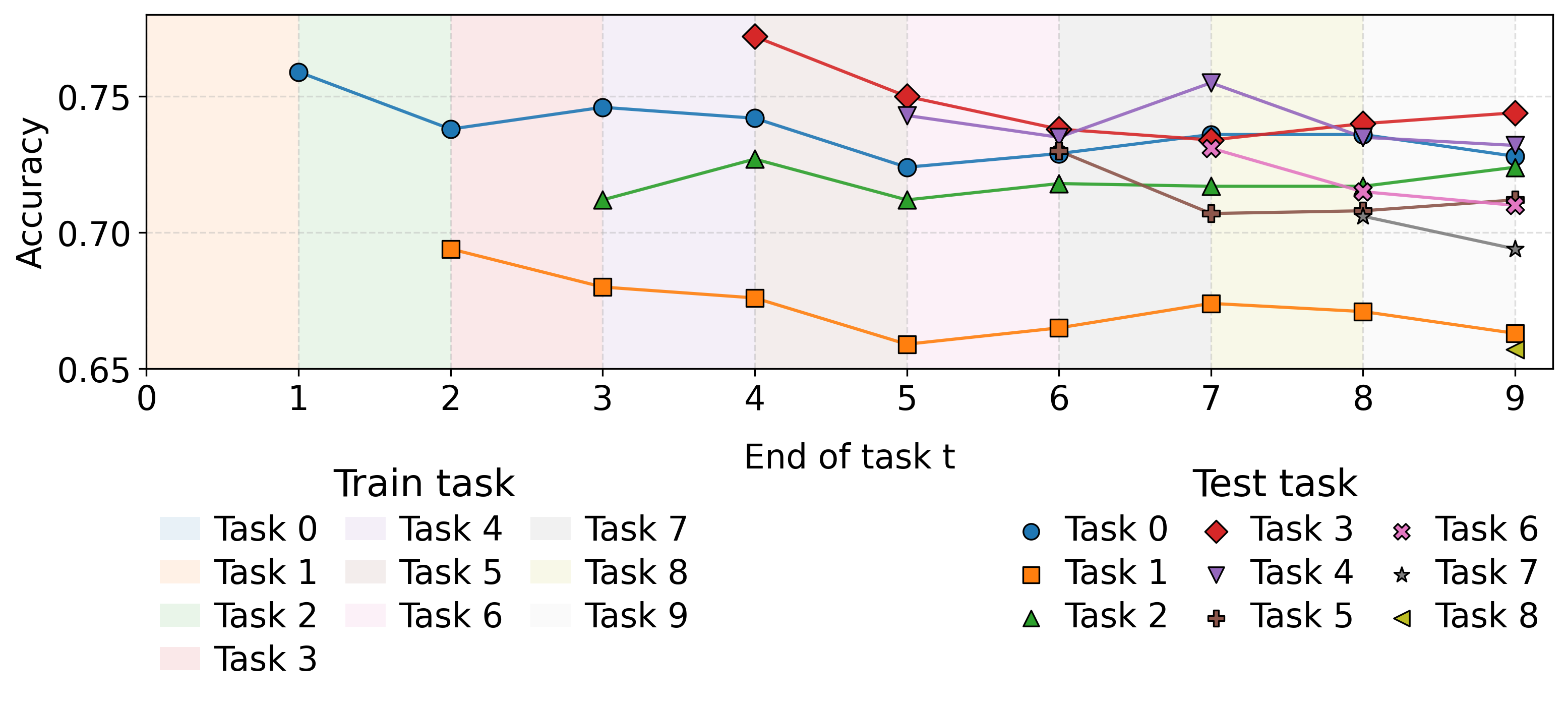}
        \label{fig:accuracy_sae_raw_derpp}}
    \hfill
    \subfloat[DER++ - Linear translation]{
        \includegraphics[trim=0cm 4.02cm 0cm 0.1cm, clip, width=0.42\textwidth]{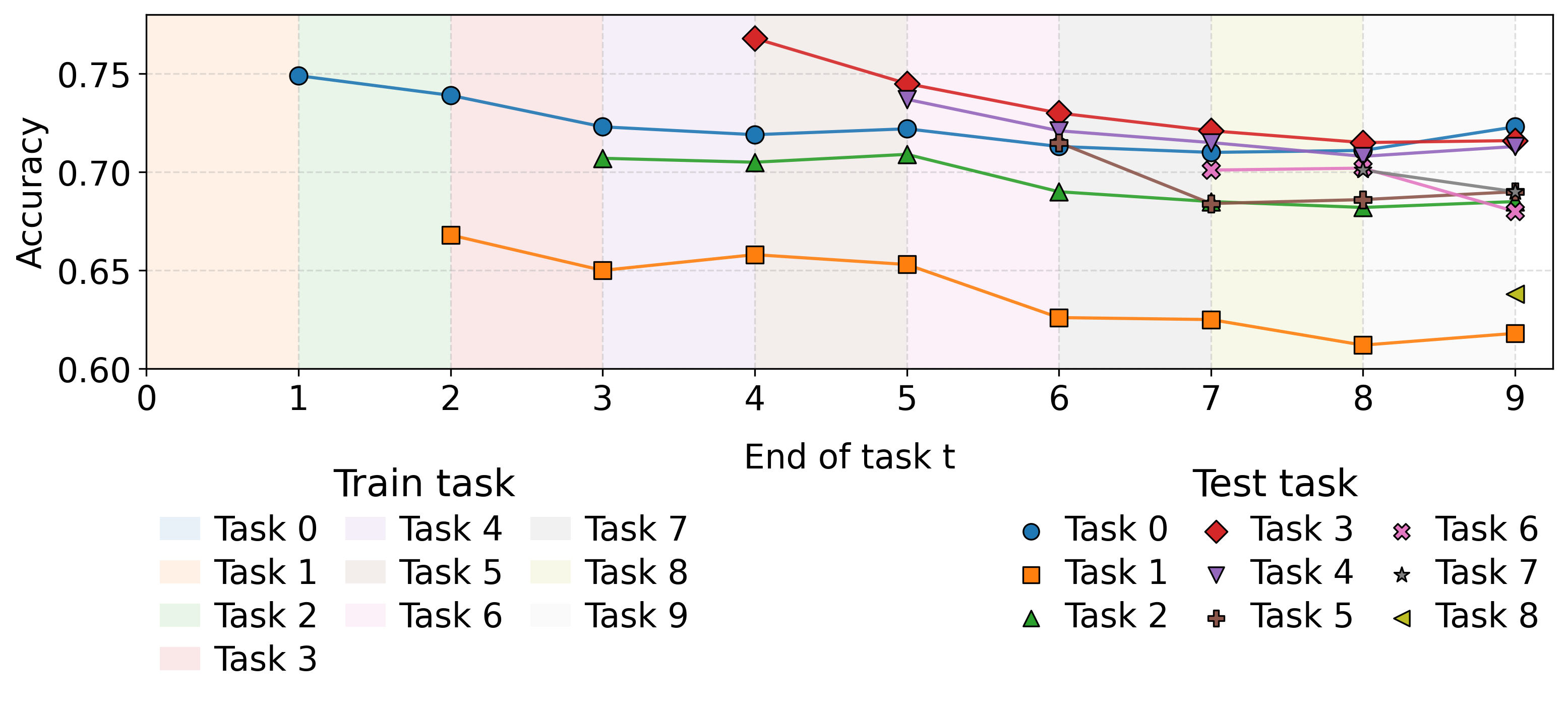}
        \label{fig:accuracy_sae_linear_derpp}}

    \caption{\textbf{Accuracy of probes} for 10seq-tiny-ImageNet classes prediction throughout continual training based on the \textbf{SAE latent neuron activations} (concept-level). Accuracy is higher in all cases after applying the linear translation to the raw features. We use the same legend as in Fig. \ref{fig:deletion_scatter}.}\label{fig:scatter_accuracy_sae}
\end{figure}

Figures~\ref{fig:scatter_accuracy_raw} and~\ref{fig:scatter_accuracy_sae} show the temporal dynamics of feature- and concept-based probe accuracy for task~$t$, measured on task-$t$ data after subsequent tasks~$t+s$. In all cases, accuracy on raw task-$t$ representations declines as~$s$ increases, but with different dynamics: under SGD and EWC it drops quickly and stabilizes only slightly above~0, whereas under DER++ and LwF it remains much higher. DER++ preserves task performance in non-translated data better than LwF, likely due to combining replay and distillation. For LwF, accuracy after linear translation stays nearly constant over time, indicating that the apparent forgetting is largely reversible by a simple linear remapping. For SGD, EWC and DER++, translation  recovers much of the lost performance, but less so as~$s$ grows, further suggesting that stronger forgetting may involve more advanced drifts or partial information erasure. Similar trends both at the concept level and the level of continual model entangled features, show that these effects persists even in sparse, more disentangled representations. Additionally, operating at the concept level appears to slow down the decline for DER++ and LwF relative to the raw-feature probe. Importantly, the recovery of accuracy for probes operating on SAE latents suggests that task-relevant information remains largely compatible with the original task-anchored concept basis after alignment. This indicates that a significant portion of fine-grained, concept-level information is not fully deleted during continual learning, but instead becomes misaligned and less directly accessible. Moreover, this recovery provides functional evidence that the linear translation of the continual model features restores access to fine-grained concept-based information.

\section{Stability of concept-based forgetting analysis across runs}\label{app:stability_across_runs}

\begin{figure}[h]
    \centering
    \subfloat[Active neurons]{
        \includegraphics[width=0.31\textwidth]{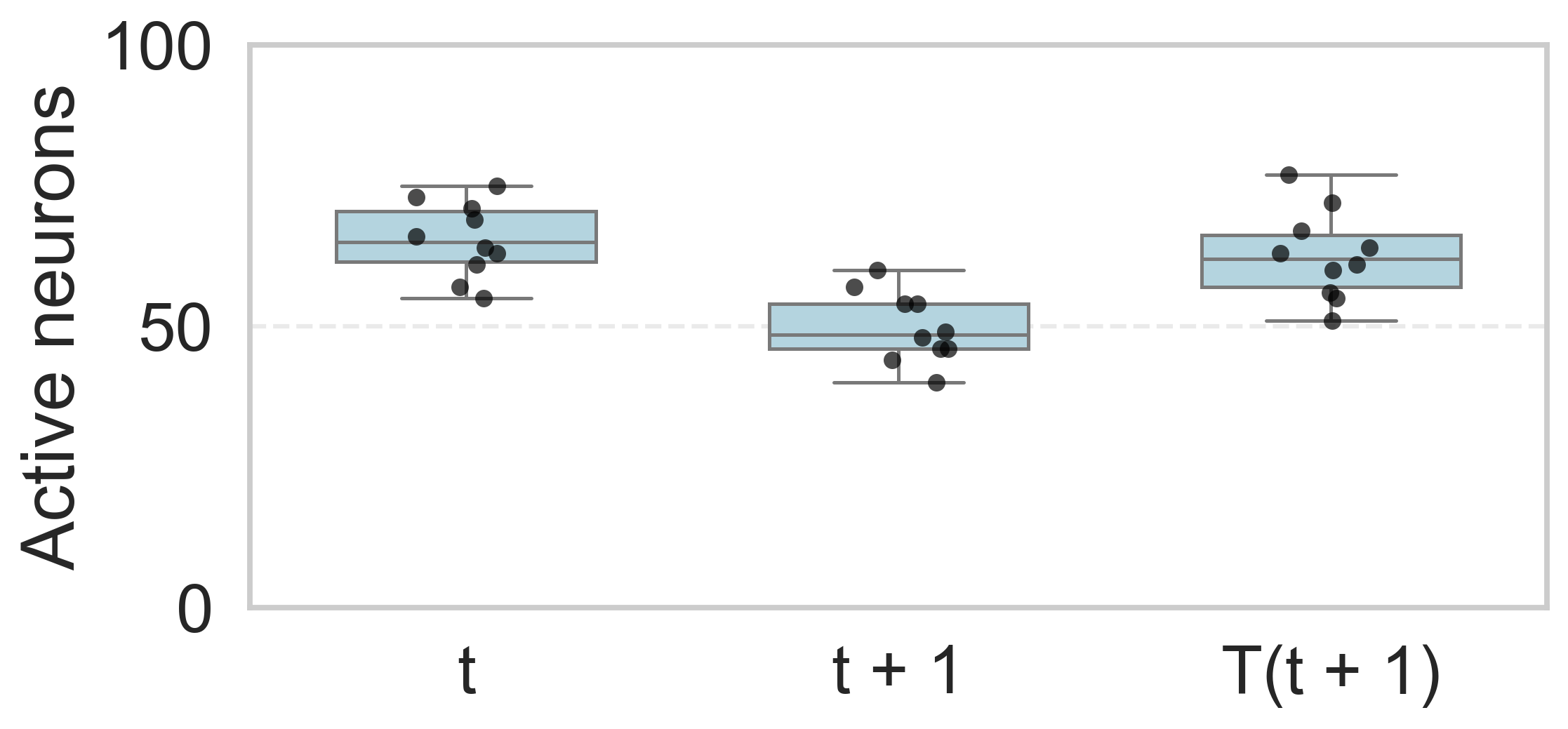}
        \label{fig:run_stab_analysis_active}}
    \hfill
    \subfloat[Deletion ratio]{
        \includegraphics[width=0.31\textwidth]{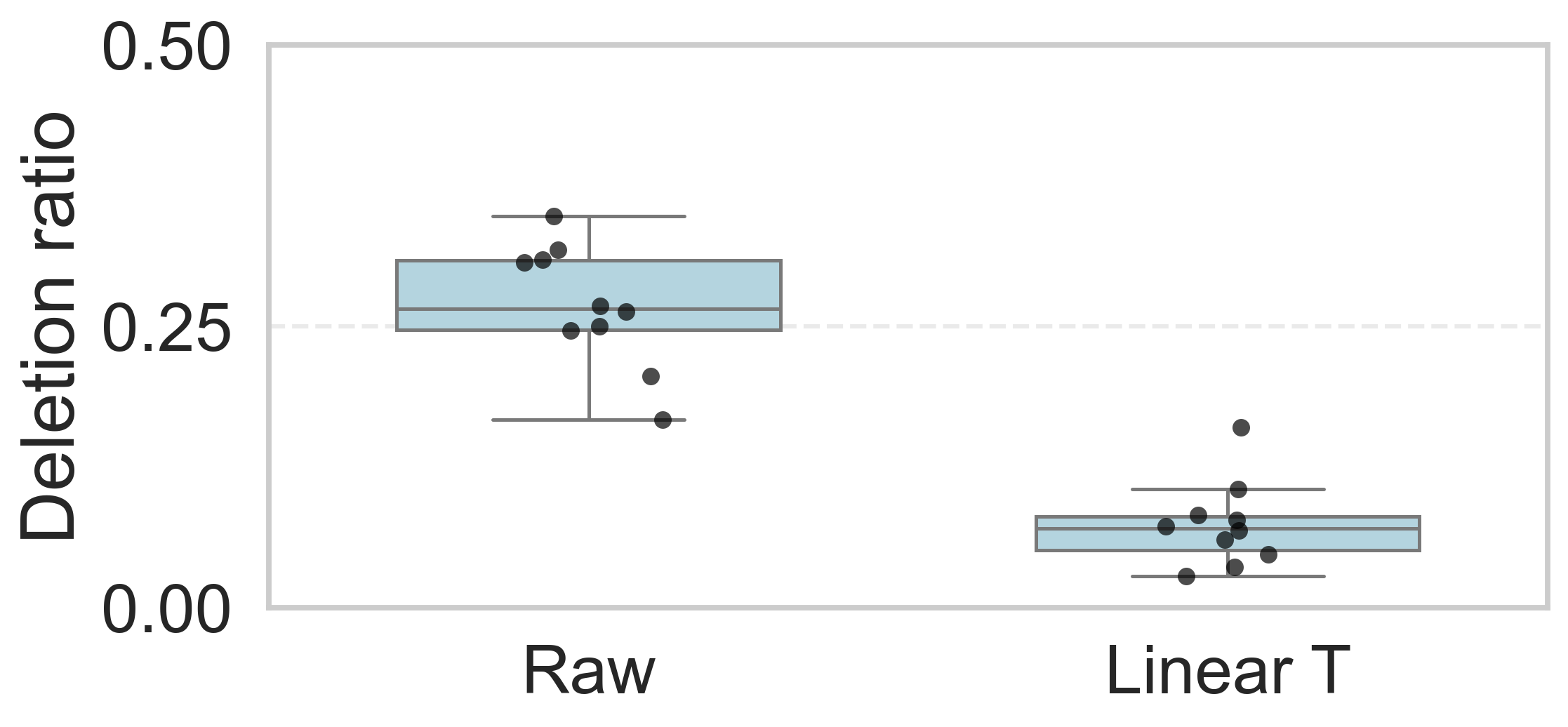}
        \label{fig:run_stab_analysis_deleted}}
        \hfill
    \subfloat[Regained count and mass ratio]{
        \includegraphics[width=0.31\textwidth]{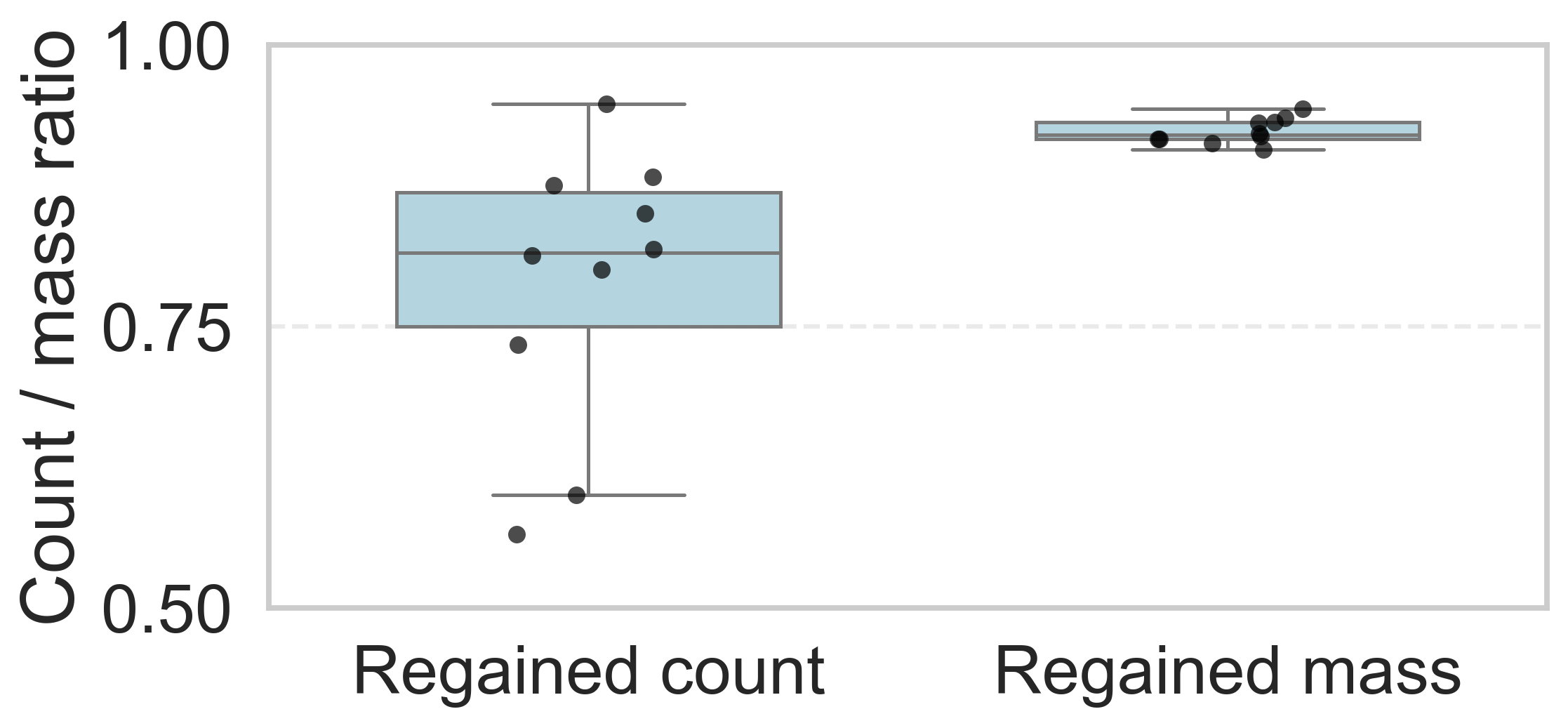}
        \label{fig:run_stab_analysis_regained}}

    \subfloat[Representation-based probe accuracy]{
        \includegraphics[width=0.31\textwidth]{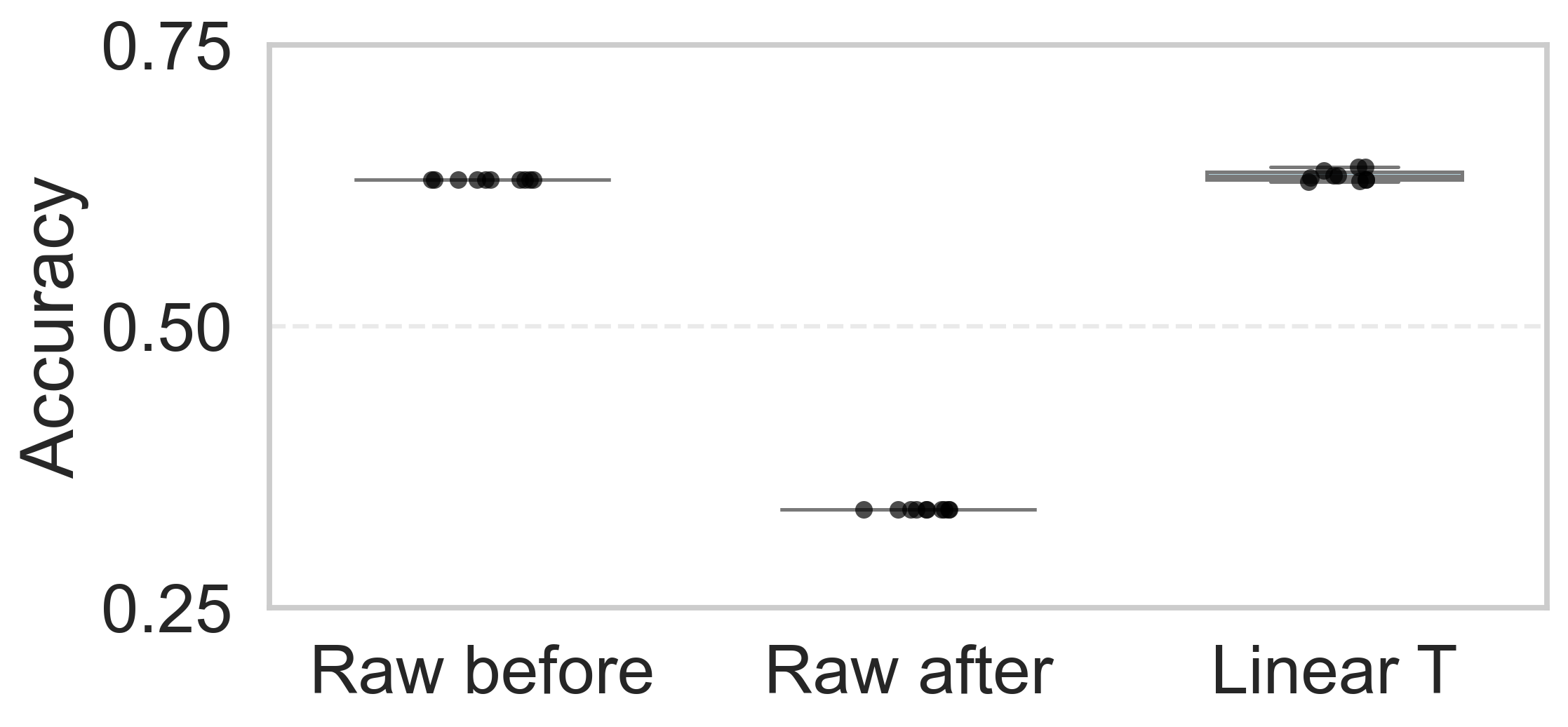}
        \label{fig:run_stab_analysis_acc}}
    \hfill
    \subfloat[Concept-based probe accuracy]{
        \includegraphics[width=0.31\textwidth]{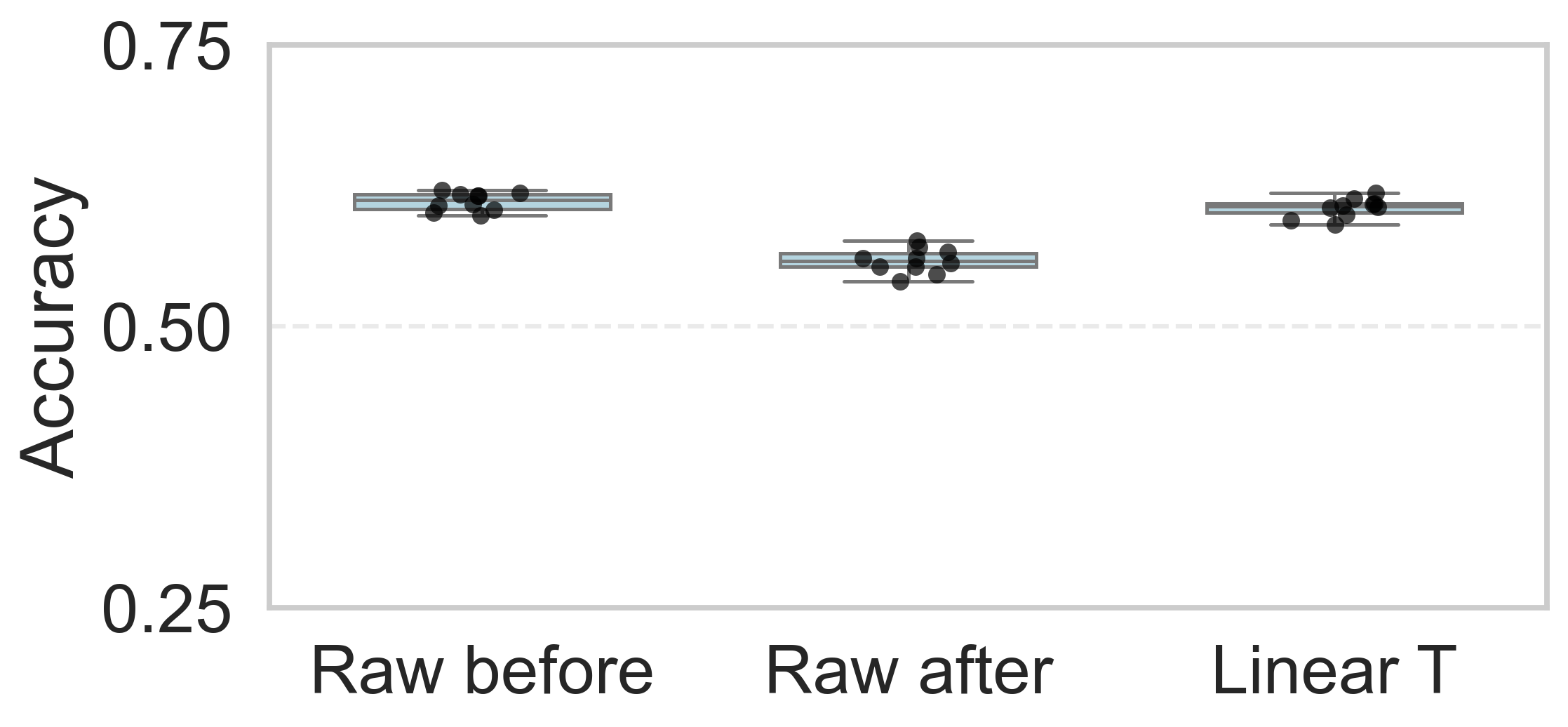}
        \label{fig:run_stab_analysis_sae_acc}}
        \hfill
    \subfloat[Mean balanced accuracy and F1]{
        \includegraphics[width=0.31\textwidth]{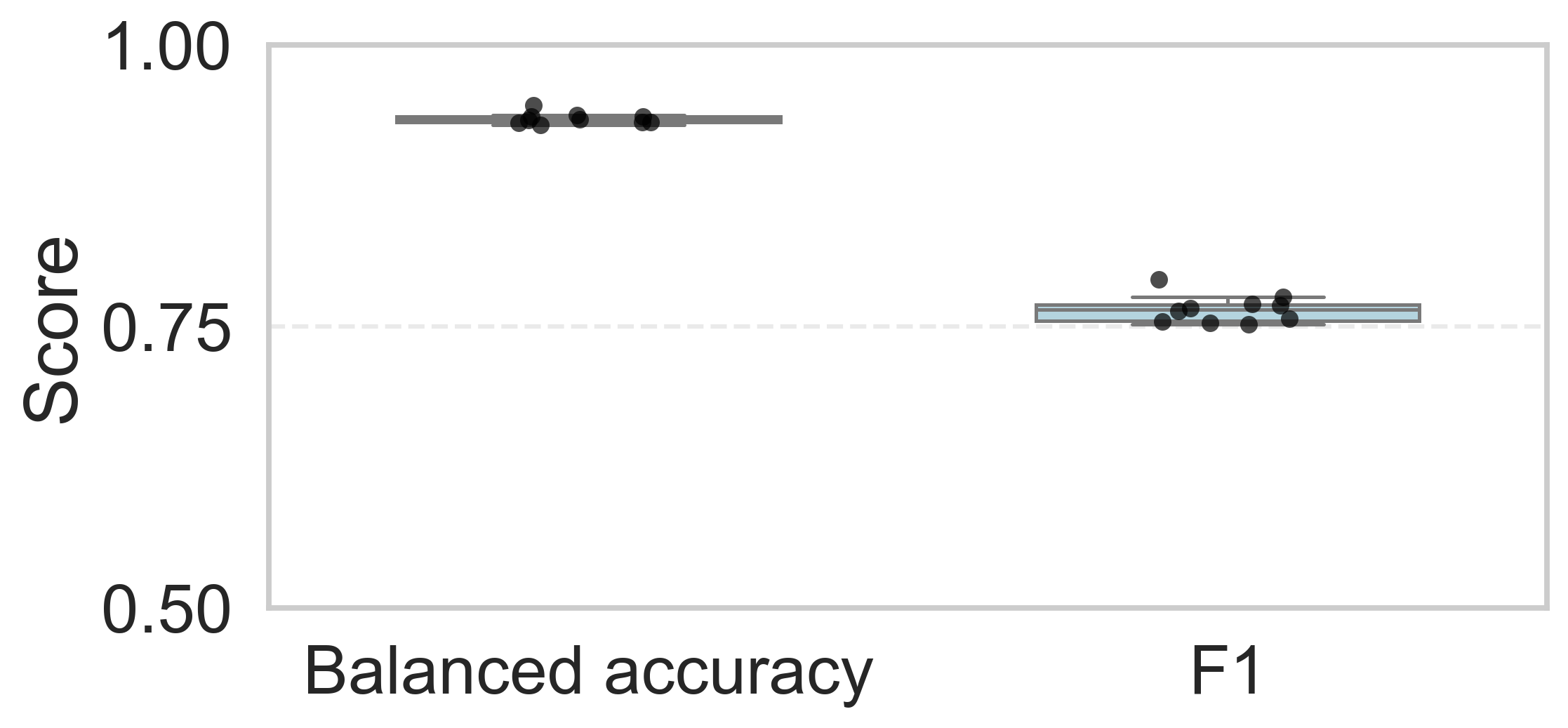}
        \label{fig:run_stab_analysis_mbacc_f1}}
    \caption{\textbf{Stability analysis}: box plots of the distribution of values of our key measures for the 2seq-tiny-ImageNet on the example LwF continual learning strategy case for 10 individual runs. For all measures bounded to the $[0,1]$ range, the y-axes were rescaled to dataset-specific intervals of width $0.5$ in order to facilitate visual comparison of the spread across plots }\label{fig:run_stab_analysis}
\end{figure}

Box plots of key metric values across 10 runs for an example configuration (2seq-tiny-ImageNet, LwF) presented in Fig. \ref{fig:run_stab_analysis} show that the results are largely stable across runs. The probe-based measures, especially the representation- and concept-level probe accuracies and the mean balanced accuracy and F1 for concept prediction, show the most compact distributions, indicating high consistency of the main conclusions. Larger variability is observed for the regained count ratio, although its main mass remains concentrated in a relatively narrow range. This suggests that the exact number of regained concepts may vary between runs, but the overall conclusion that linear alignment restores a substantial fraction of the concept information remains robust.

\section{Impact of frequency-based binarization threshold on concept-based forgetting analysis}\label{app:stability_binarization}

\begin{figure}[h]
    \centering
    \subfloat[Active neurons]{
        \includegraphics[width=0.42\textwidth]{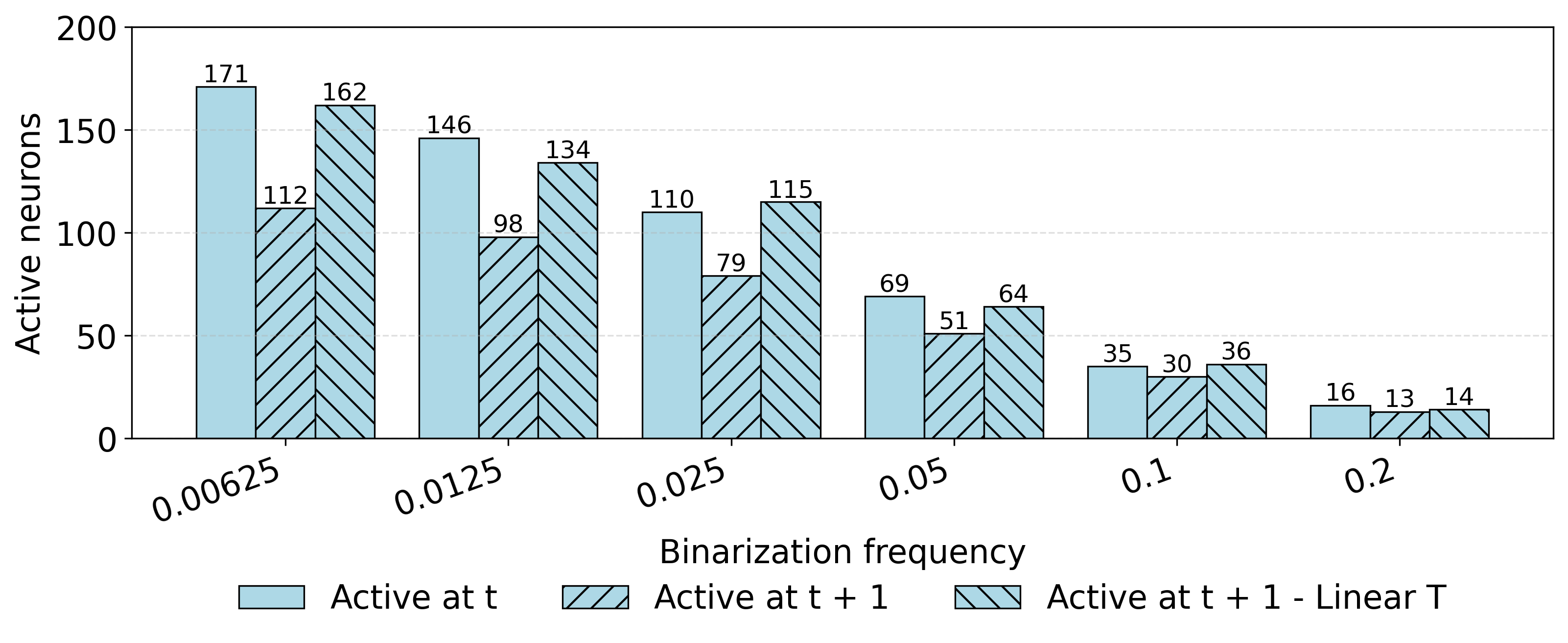}
        \label{fig:frec_act}}
    \hfill
    \subfloat[Deletion ratio]{
        \includegraphics[width=0.42\textwidth]{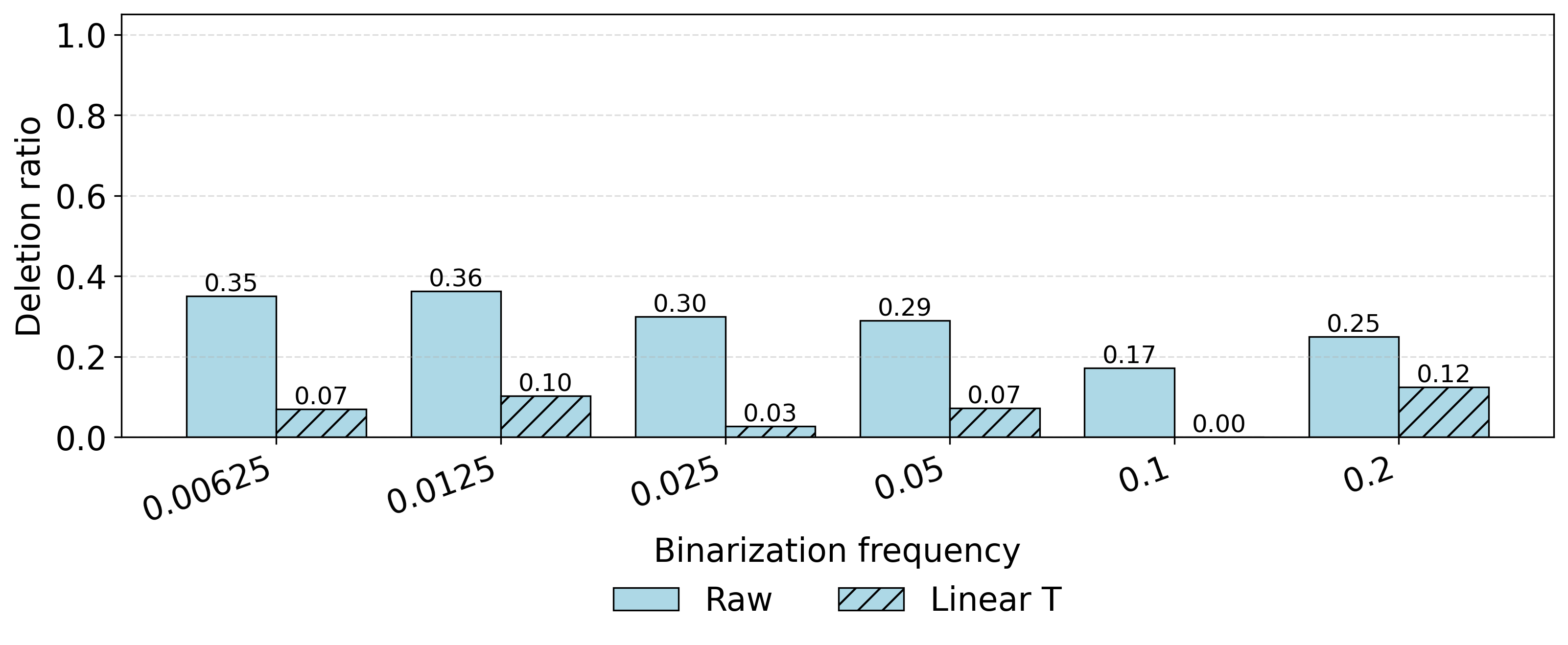}
        \label{fig:frec_del}}

    \subfloat[Regained count and mass ratio]{
        \includegraphics[width=0.42\textwidth]{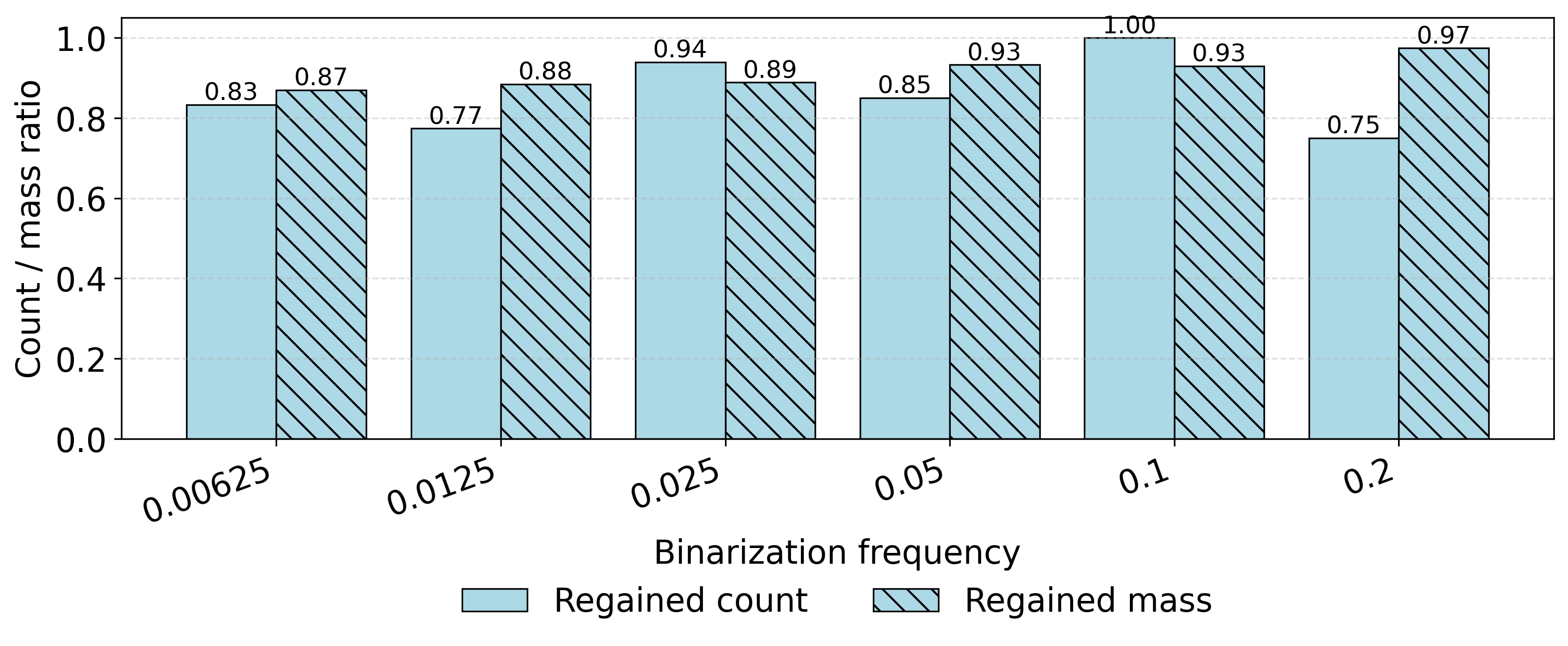}
        \label{fig:frec_reg}}
        \hfill
    \subfloat[Representation-based probe accuracy]{
        \includegraphics[width=0.42\textwidth]{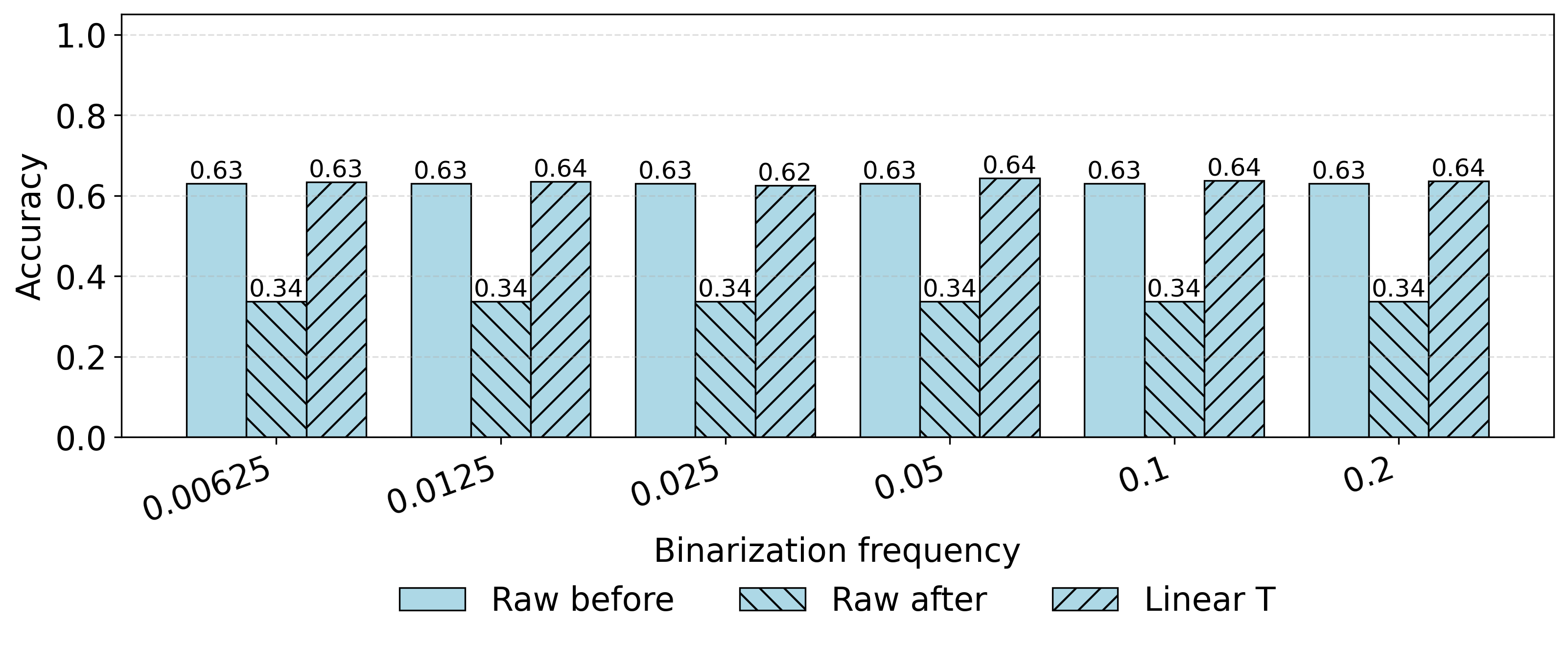}
        \label{fig:frec_acc}}
    
    \subfloat[Concept-based probe accuracy]{
        \includegraphics[width=0.42\textwidth]{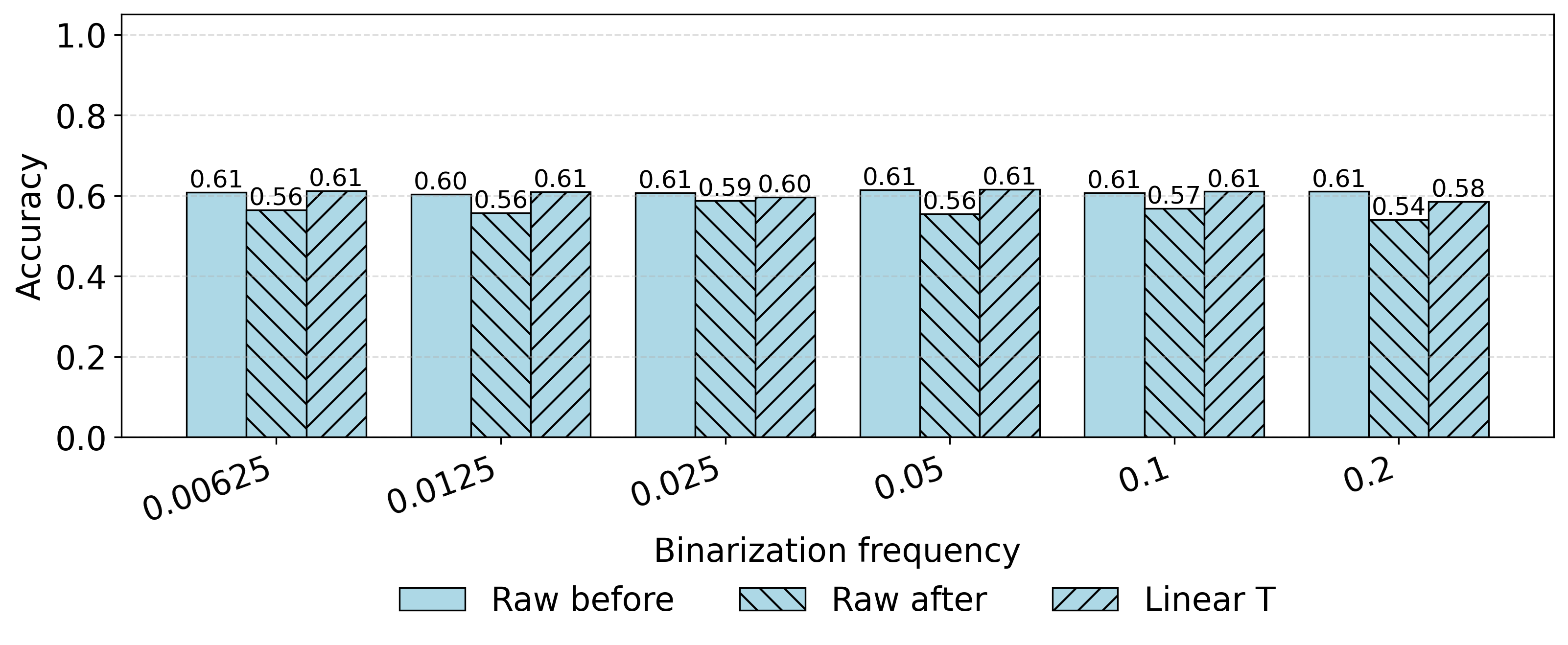}
        \label{fig:frec_acc_sae}}
        \hfill
    \subfloat[Mean balanced accuracy and F1]{
        \includegraphics[width=0.42\textwidth]{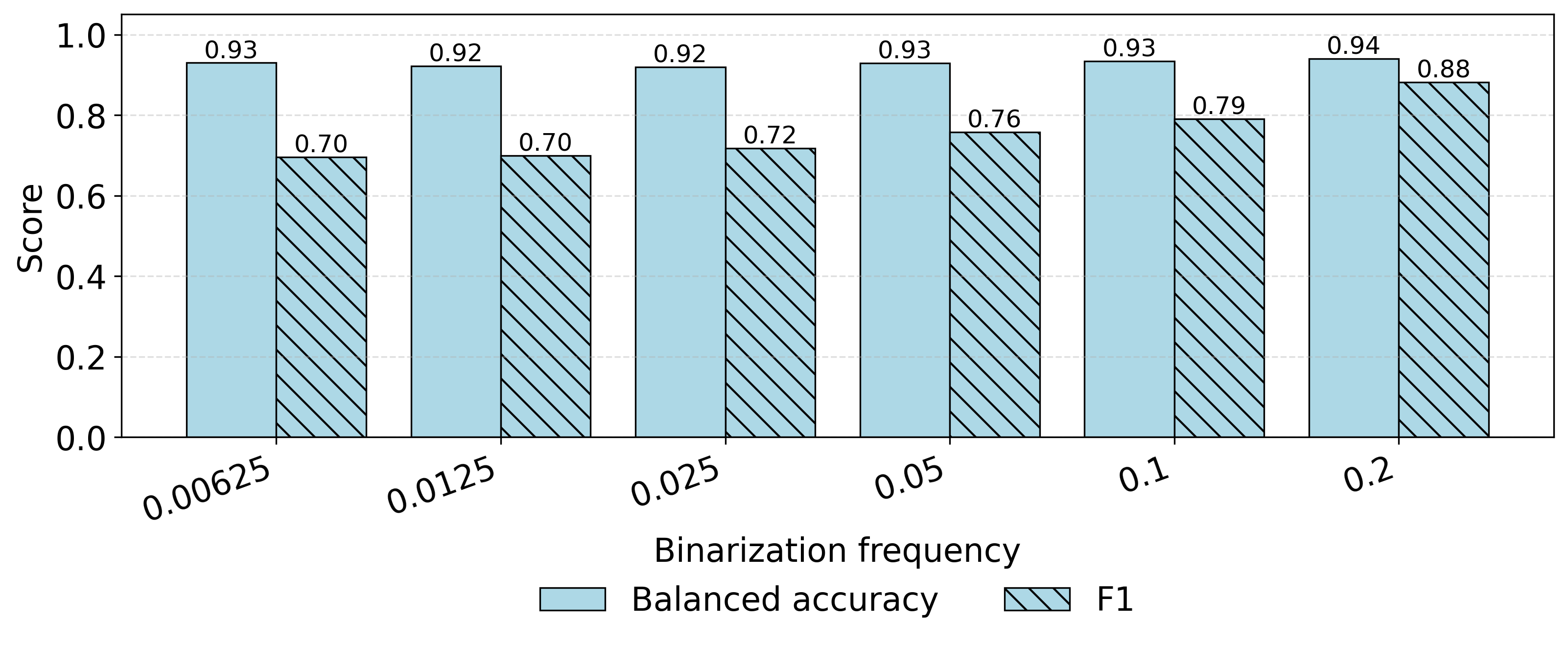}
        \label{fig:frec_f1}}
    \caption{The \textbf{impact of frequency-based binarization threshold} on metrics for 2seq-tiny-ImageNet task 0 under LwF. }\label{fig:freq_bars}
\end{figure}

With the frequency-based binarization criterion, a latent is considered active if it exceeds its task-$t$ train-set mean on at least a specified fraction of task-$t$ samples. As shown in Fig.~\ref{fig:freq_bars} for the example LwF 2seq-tiny-ImageNet case, increasing this threshold reduces the number of detected active neurons, as could be predicted, but the overall forgetting picture remains largely unchanged. Across all thresholds, raw $t+1$ representations of $t$ exhibit fewer active concepts and non-zero deletion, while linear alignment restores most of the lost activity and keeps both feature-level and concept-level probe accuracy close to the task-$t$ reference. This indicates that our concept-forgetting observations are not an artifact of a particular binarization threshold, but are qualitatively robust across a range of frequency-based thresholds.

\begin{figure}[h]
    \centering
    \subfloat[Balanced accuracy]{
        \includegraphics[width=0.42\textwidth]{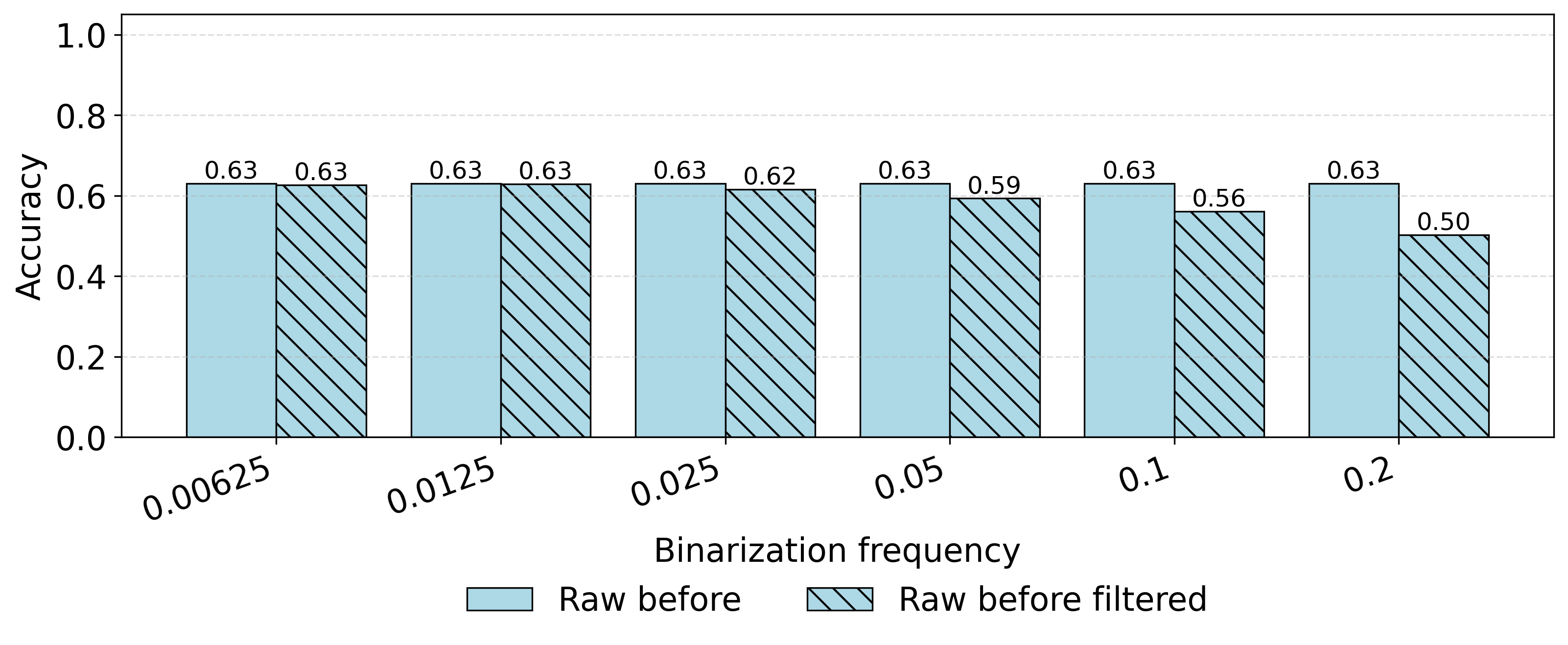}
        \label{fig:accuracy_binarization_raw_multiple}}
    \hfill
    \subfloat[F1 score]{
        \includegraphics[width=0.42\textwidth]{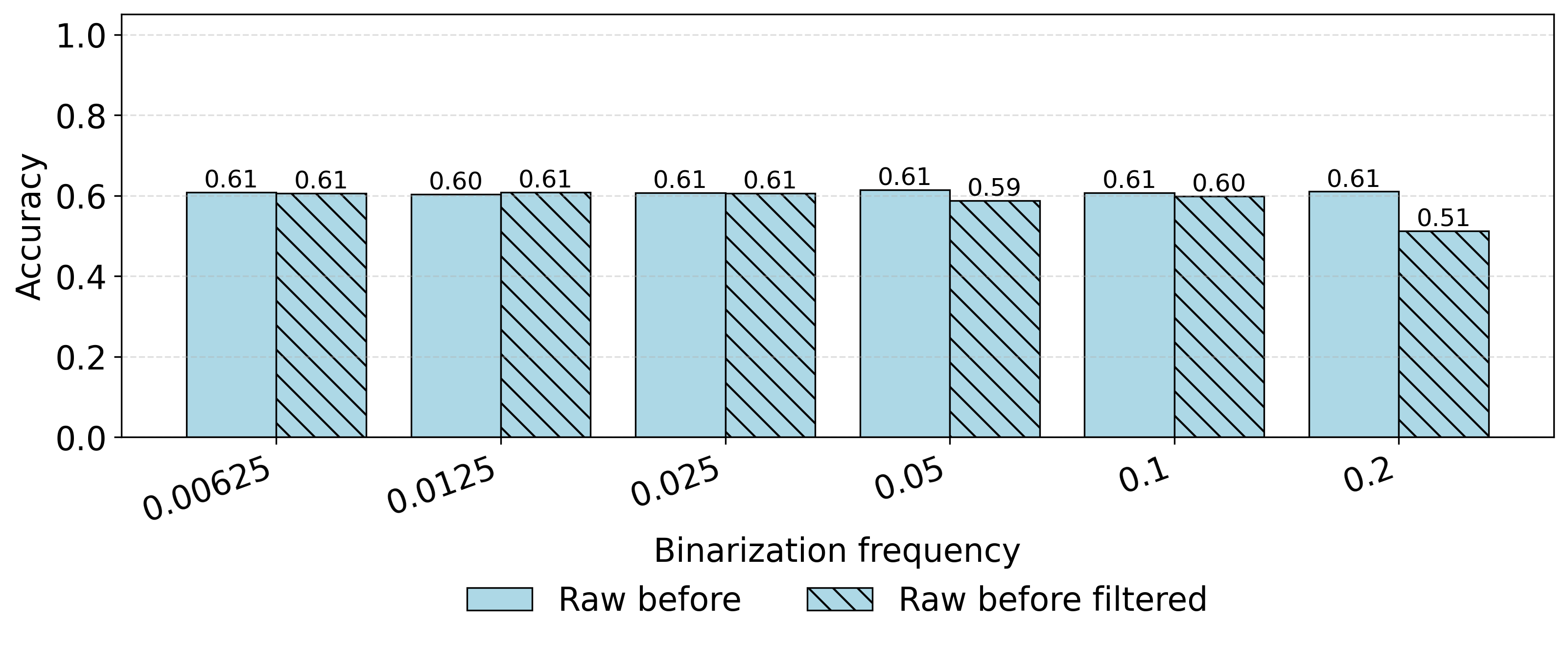}
        \label{fig:accuracy_binarization_sae_multiple}}
    \caption{The \textbf{impact of frequency-based binarization threshold} on accuracy of probes built for prediction of 2seq-tiny-ImageNet task 0 classes under LwF based on the raw representations from the continual model (representation-level) and based on the SAE latent neuron activations (concept-level). All accuracy values are computed for the task 0 data after training on task 0, but for the "Raw before filtered" case, we use our SAE decoder to zero out the neurons that are not active binary under our binarization rule. We examine frequency-based threshold in $[0.00625, 0.0125, 0.025, 0.05, 0.1, 0.2]$.}\label{fig:accuracy_binarization_multiple_bin_thr}
\end{figure}

To verify that the set of binary active neurons is functionally relevant, we additionally measure task-0 classification accuracy after zeroing out, through the SAE decoder, all latent neurons that are not marked as active under the given frequency-based rule. Figure~\ref{fig:accuracy_binarization_sae_multiple} shows that this filtering leads only to a limited accuracy drop across the examined thresholds, confirming that the selected active neurons retain most of the task-level information. For 2seq-tiny-ImageNet under LwF, the performance remains relatively stable across thresholds, although slightly larger drops appear for thresholds below 0.05. This suggests that very permissive thresholds include less functionally relevant activations, while stricter thresholds can entail some loss of the task-relevant information. Therefore, thresholds around 0.05 provide a reasonable trade-off between retaining task information and enforcing a meaningful notion of concept activity. We therefore use 0.05 in all setups, including CIFAR10, to keep the analysis unified across datasets.

\section{Impact of $0.05$ frequency-based binarization on probe-based task accuracy}\label{app:binarization_at_05}

As an additional check, to examine whether the active set of neurons under the frequency-based binarization rule used in our main experiments retains the functionally relevant task information for both datasets, we perform an experiment, in which we zero-out the neurons that are overall not active under the binarization rule (i.e. due to values lower than mean train activations or due to the frequency). We use the SAE's decoder for this intervention by first zeroing out the activations of the selected neurons, decoding the modified latent vector, and comparing it with the decoder output obtained from the raw latent activations. We then use this difference to update the original input representation, rather than replacing it directly with the decoded vector, in order to avoid propagating the SAE reconstruction error.

\begin{figure}[h]
    \centering
    \subfloat[Representation-level]{
        \includegraphics[width=0.42\textwidth]{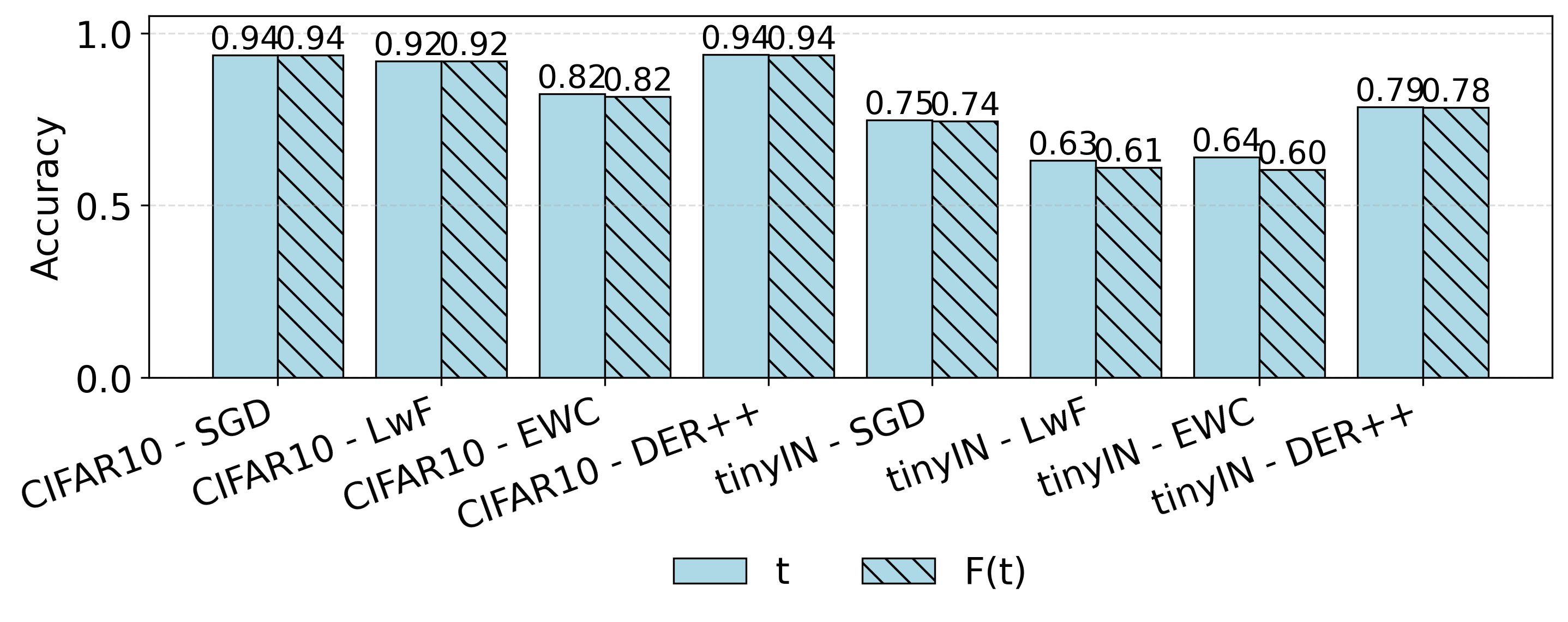}
        \label{fig:probe_accuracy_sanity_raw}}
    \hfill
    \subfloat[Concept-level]{
        \includegraphics[width=0.42\textwidth]{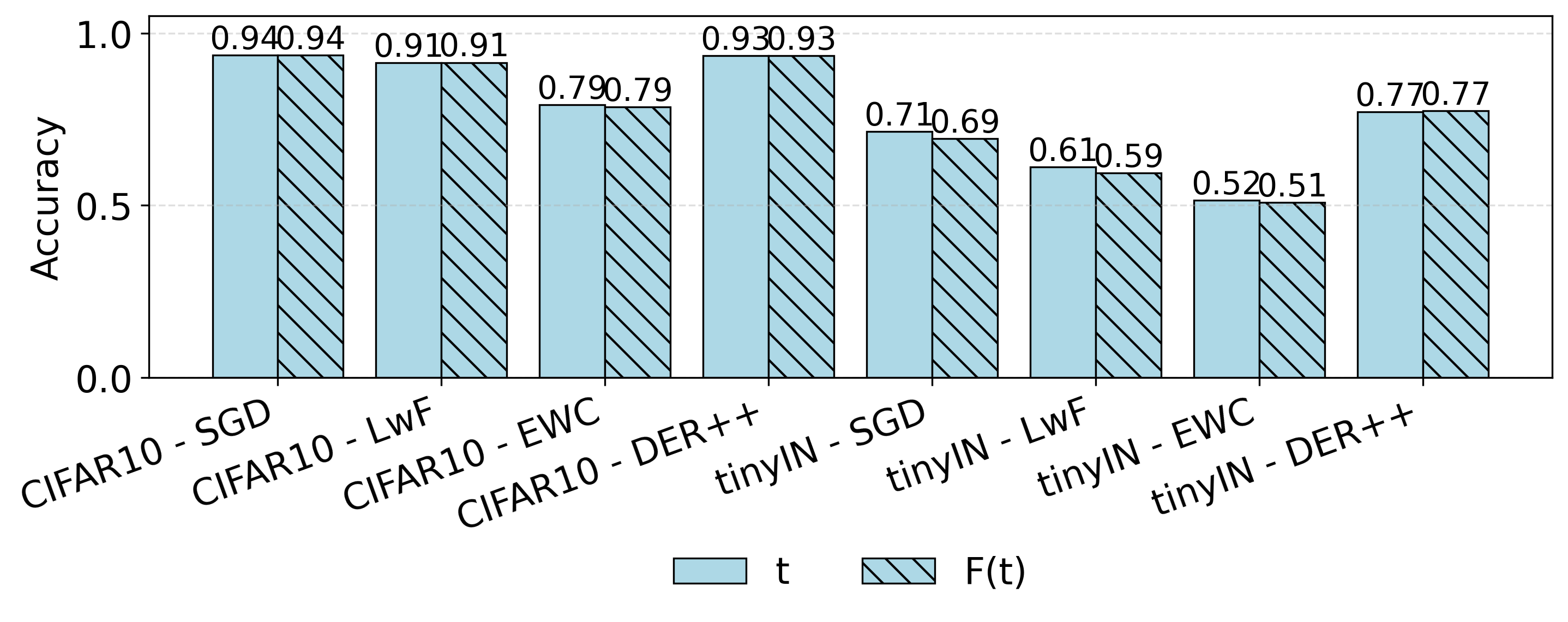}
        \label{fig:probe_accuracy_sanity_sae}}
    \caption{Accuracy of probes built for prediction of 2seq-CIFAR10 and 2seq-tiny-ImageNet task 0 classes  based on the raw representations from the continual model (representation-level) and based on the SAE latent neuron activations (concept-level). Both accuracy values are computed for the task 0 data after training on task 0, but for the $F(t+1)$ case, we use our SAE decoder to zero out the neurons that are not active binary under our binarization rule (e.g. the frequency of their values that are higher than the activation threshold is lower than the frequency threshold).}\label{fig:accuracy_binarization_zeroing_neurons}
\end{figure}

Fig. \ref{fig:accuracy_binarization_zeroing_neurons} shows that using the frequency-based threshold of 0.05 preserves most of the probe accuracy across all configurations, both at the representation and concept level. This indicates that the set of neurons selected as active retains the majority of task-relevant information despite being only a subset of the full latent space. Overall, these results support the reliability of the adopted binarization rule as the basis for our concept-level analysis across all examined settings.

\section{Impact of SAE's $K$ on concept-based forgetting analysis}\label{app:stability_sae_K}

\begin{figure}[h]
    \centering
    \subfloat[Active neurons]{
        \includegraphics[width=0.42\textwidth]{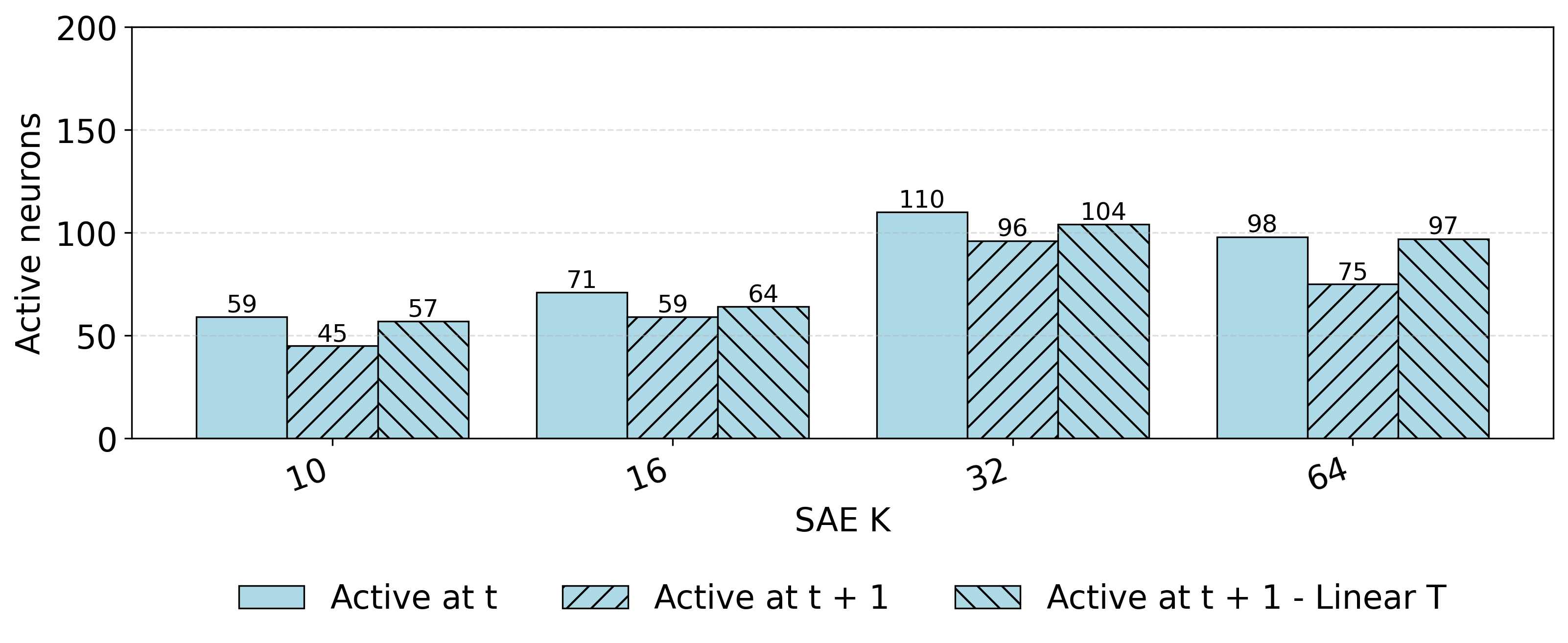}
        \label{fig:stabk_act}}
    \hfill
    \subfloat[Deletion ratio]{
        \includegraphics[width=0.42\textwidth]{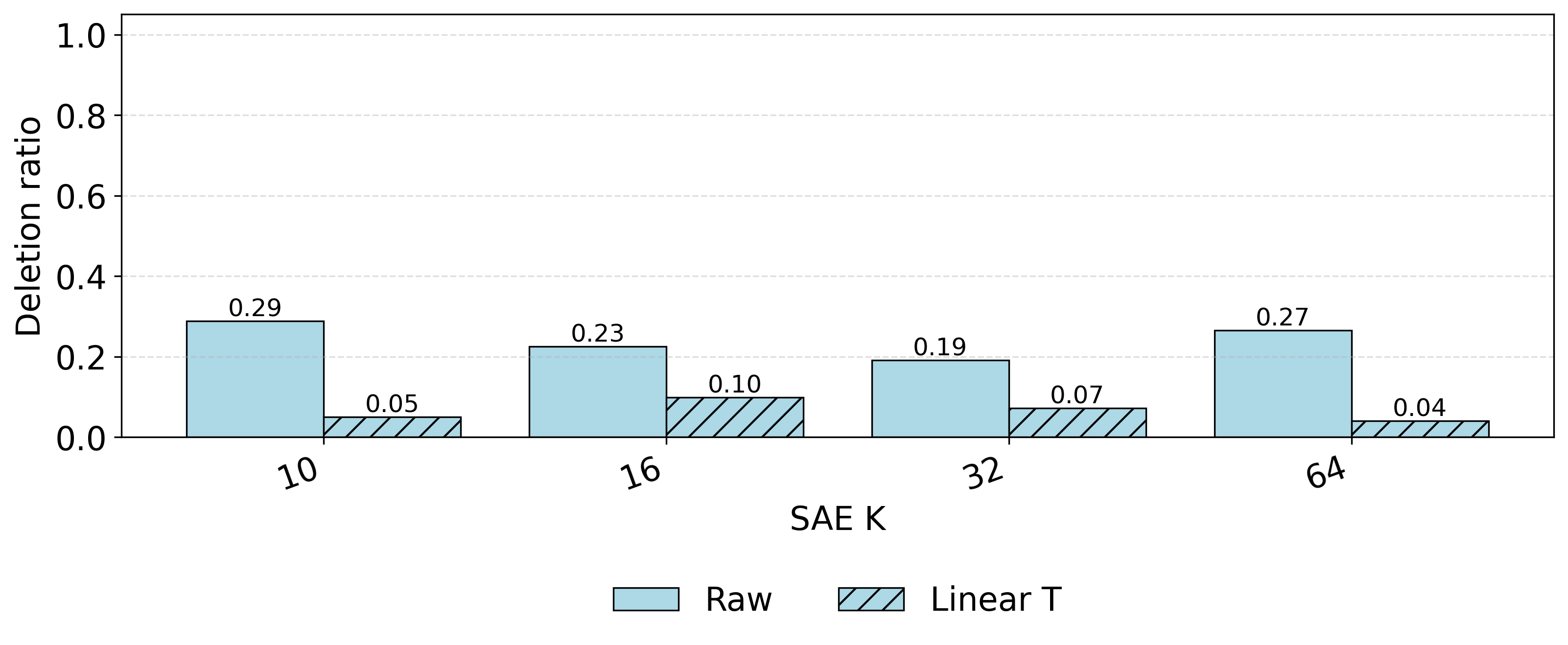}
        \label{fig:stabk_del}}

    \subfloat[Regained count and mass ratio]{
        \includegraphics[width=0.42\textwidth]{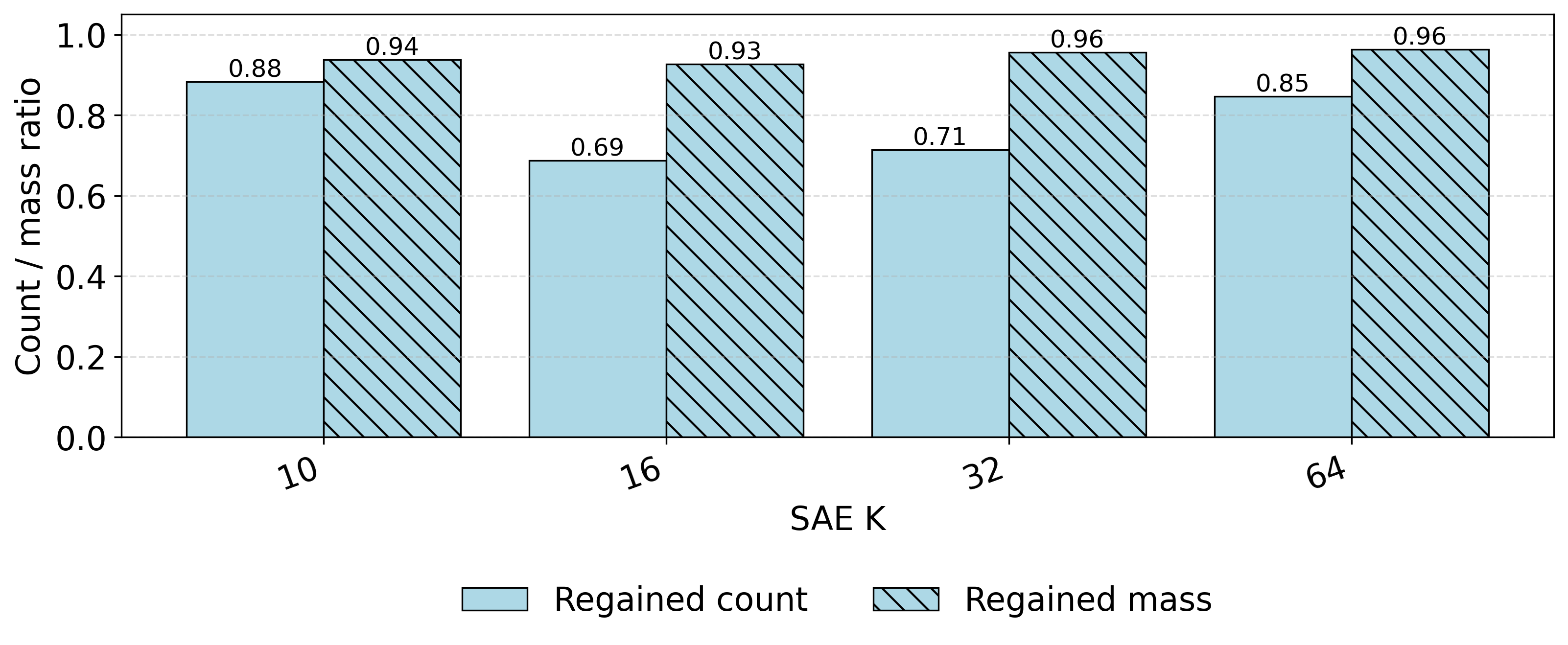}
        \label{fig:stabk_reg}}
        \hfill
    \subfloat[Representation-based probe accuracy]{
        \includegraphics[width=0.42\textwidth]{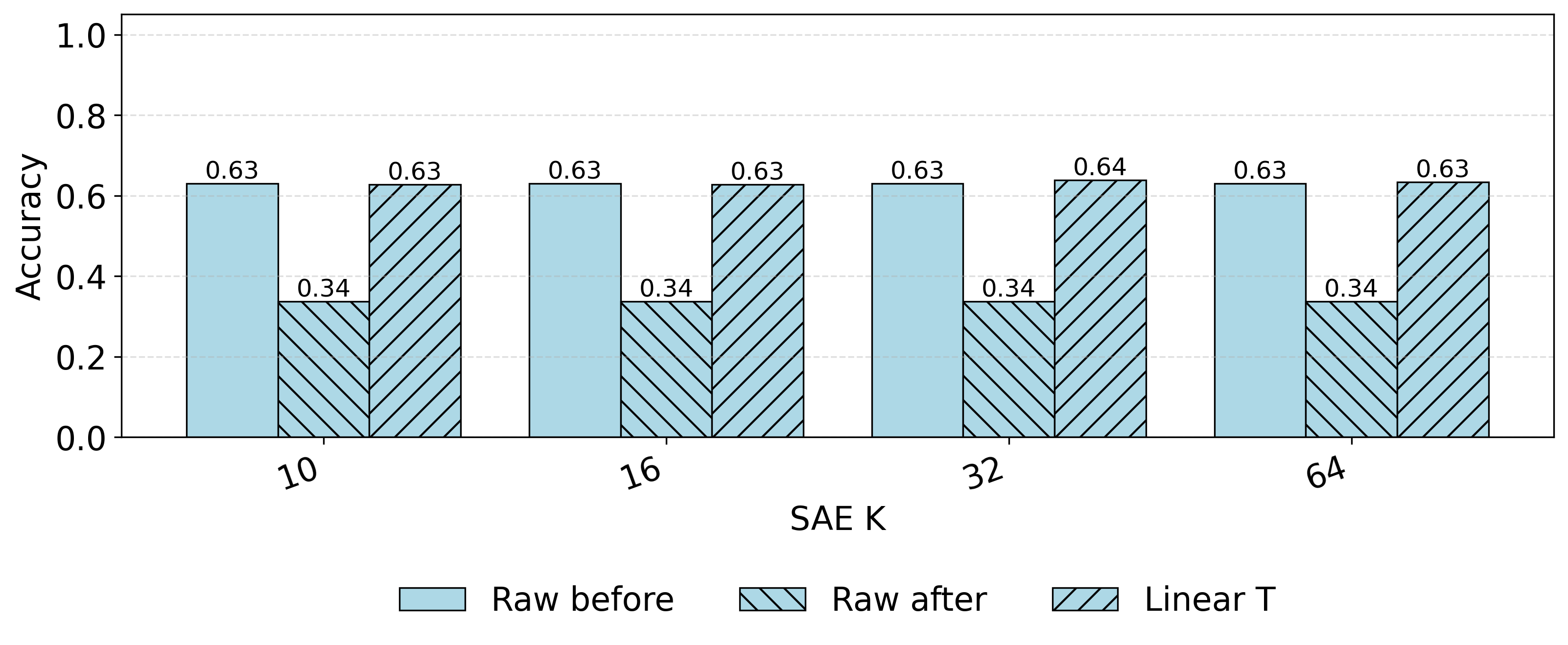}
        \label{fig:acc_stabk}}
    
    \subfloat[Concept-based probe accuracy]{
        \includegraphics[width=0.42\textwidth]{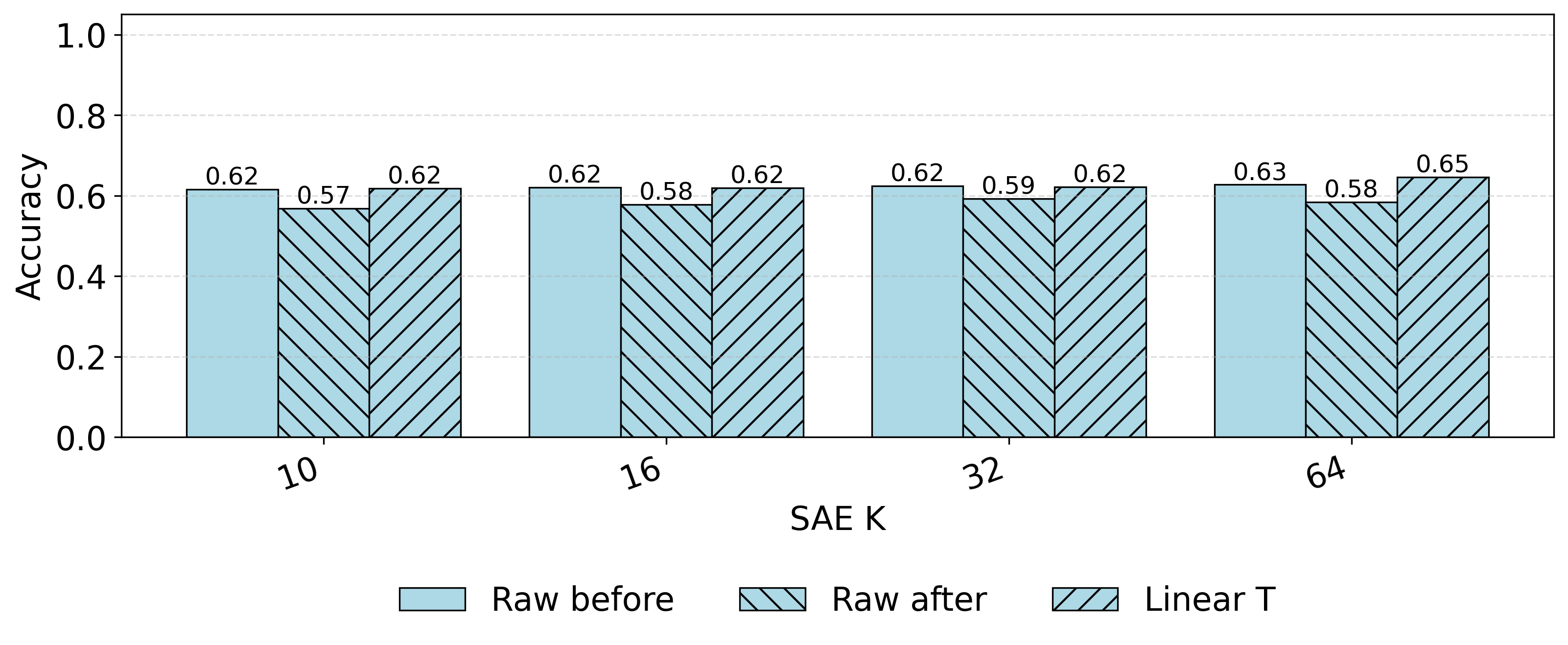}
        \label{fig:acc_sae_stabk}}
        \hfill
    \subfloat[Mean balanced accuracy and F1]{
        \includegraphics[width=0.42\textwidth]{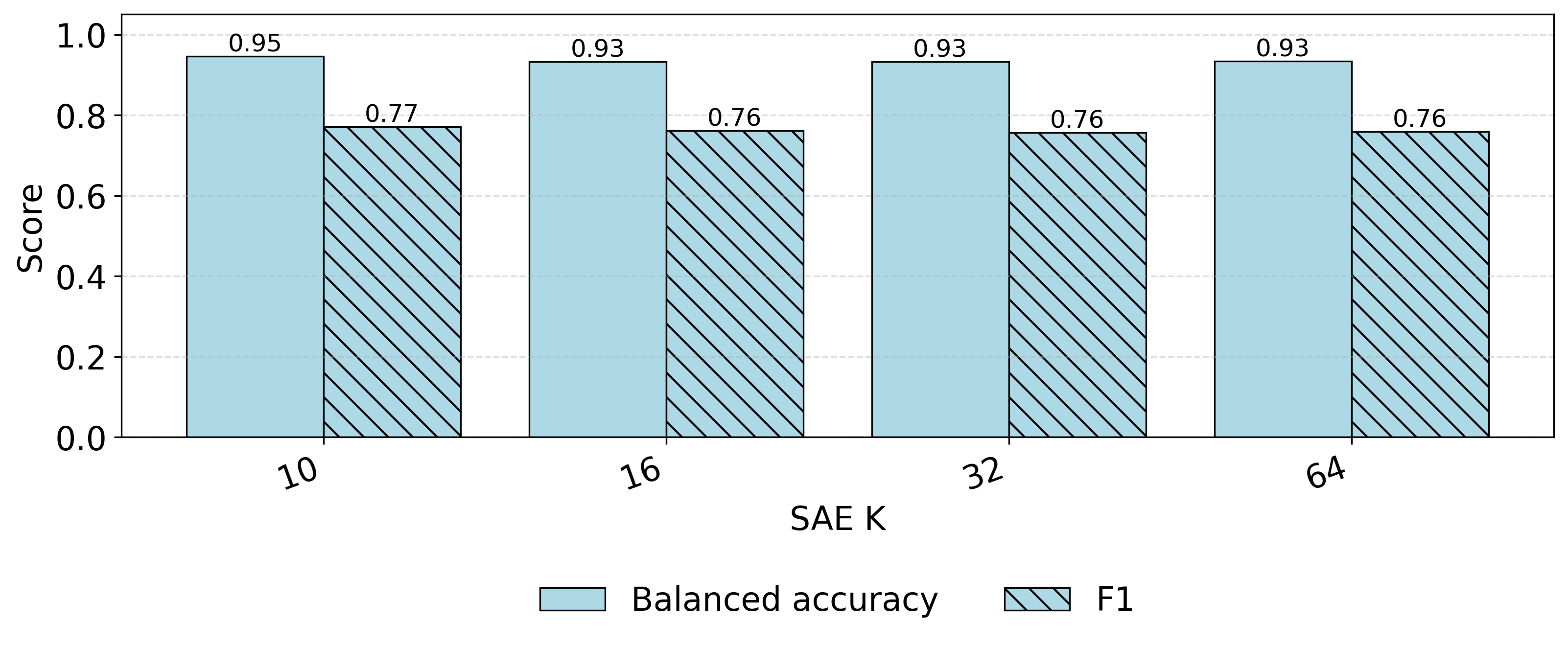}
        \label{fig:f1_stabk}}
    \caption{The \textbf{impact of SAE's K} on metrics for 2seq-tiny-ImageNet task 0 under LwF. }\label{fig:metrics_sae_stabk}
\end{figure}

Fig. \ref{fig:metrics_sae_stabk} shows the impact of the sparsity-control parameter $K$ used for the BatchTopK SAE on the key metrics for the example 2seq-tiny-ImageNet + LwF setup. Overall, the key results are qualitatively stable across the tested $K$ values, indicating that our conclusions are not sensitive to this hyperparameter. The main difference lies in the number of active neurons, which is expected since parameter $K$ directly controls the sparsity of the activations.  Although some other small quantitative differences in other metrics can be observed, the overall patterns remain unchanged: raw task-$t+1$ representations show concept deletion, linear alignment recovers a large fraction of the lost information, and both task- and concept-level probe performance remain close to the reference values. Importantly, most values fall within the spread previously observed across repeated runs of the same method, suggesting that the effect of batch size is comparable to the run-to-run variability.

\begin{figure}[h]
    \centering
    \subfloat[Balanced accuracy]{
        \includegraphics[width=0.42\textwidth]{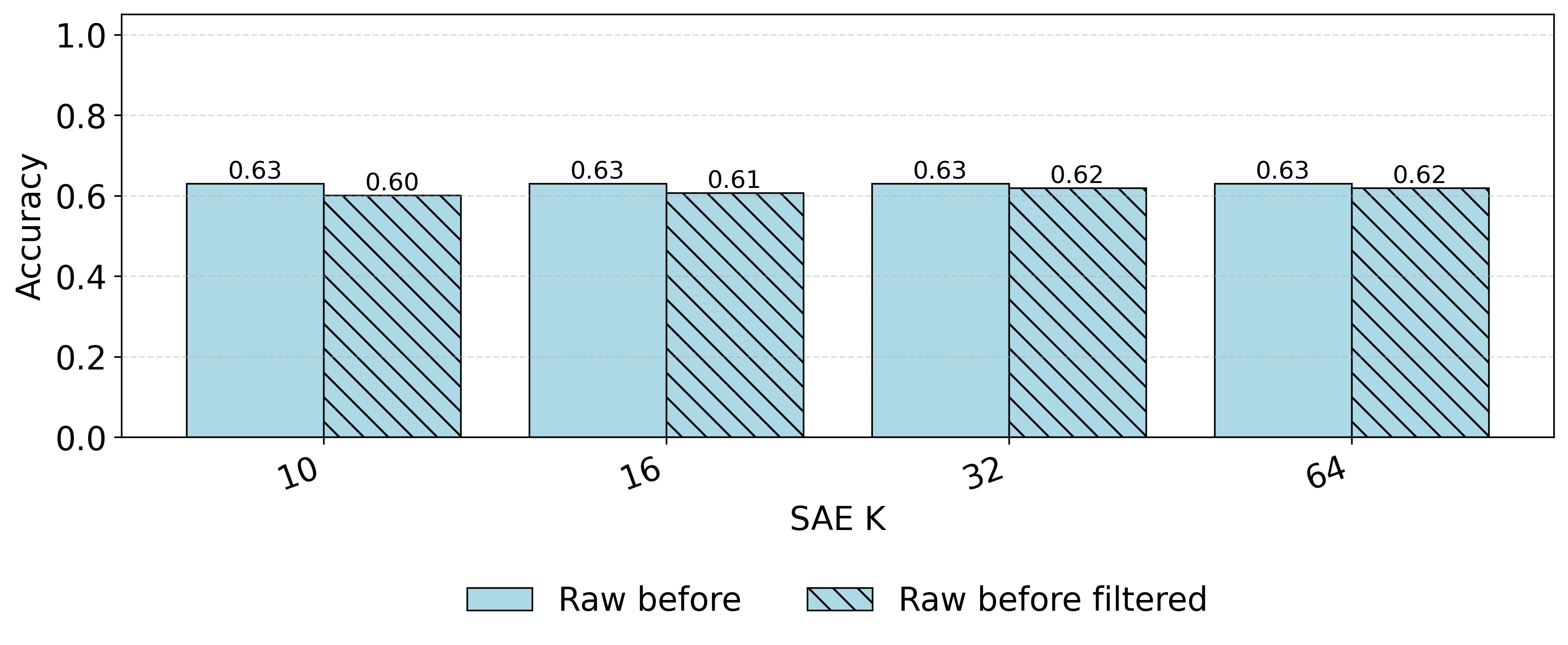}
        \label{fig:accuracy_sae_stabk_raw_multiple}}
    \hfill
    \subfloat[F1 score]{
        \includegraphics[width=0.42\textwidth]{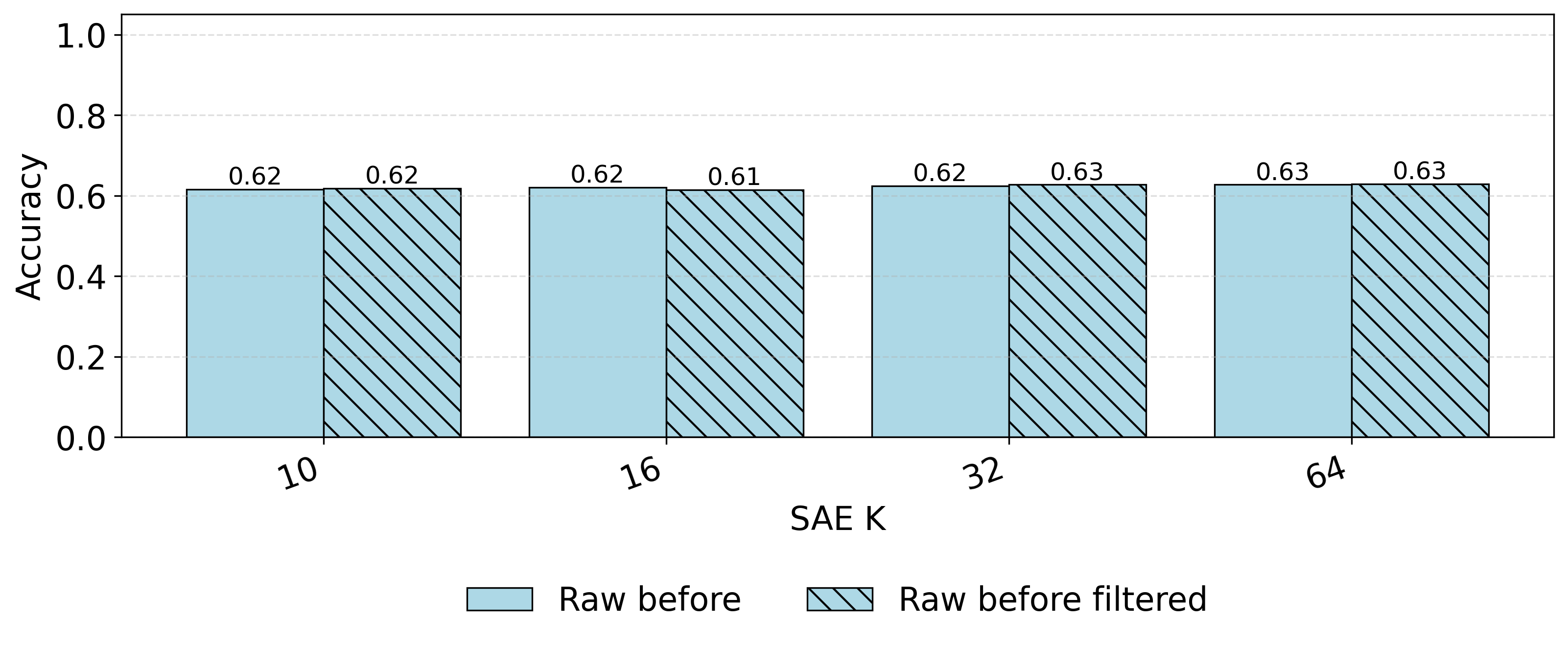}
        \label{fig:accuracy_sae_stabk_sae_multiple}}
    \caption{The \textbf{impact of SAE's K} on accuracy of probes built for prediction of 2seq-tiny-ImageNet task 0 classes under LwF based on the raw representations from the continual model (representation-level) and based on the SAE latent neuron activations (concept-level). All accuracy values are computed for the task 0 data after training on task 0, but for the "Raw before filtered" case, we use our SAE decoder to zero out the neurons that are not active binary under our binarization rule. }\label{fig:accuracy_sae_stabk}
\end{figure}

To verify that the set of binary active neurons is functionally relevant under different $K$ values, we additionally measure task-0 classification accuracy after zeroing out, through the SAE decoder, all latent neurons that are not marked as active under the given frequency-based rule. Figure~\ref{fig:accuracy_sae_stabk} shows that the approach is robust to the chosen $K$ for BatchTopK SAE regime.

\section{Impact of SAE batch size on concept-based forgetting analysis}\label{app:stability_sae_batch}

\begin{figure}[h]
    \centering
    \subfloat[Active neurons]{
        \includegraphics[width=0.42\textwidth]{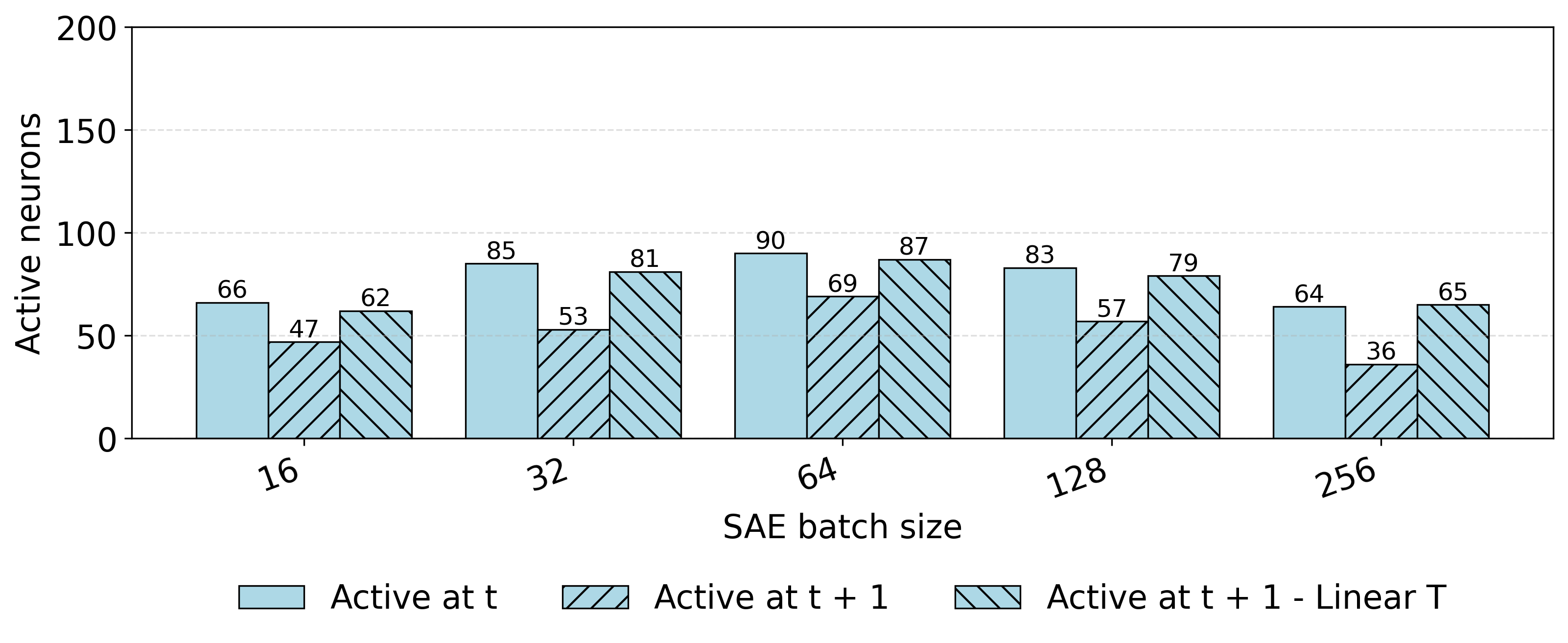}
        \label{fig:frec_act_batch}}
    \hfill
    \subfloat[Deletion ratio]{
        \includegraphics[width=0.42\textwidth]{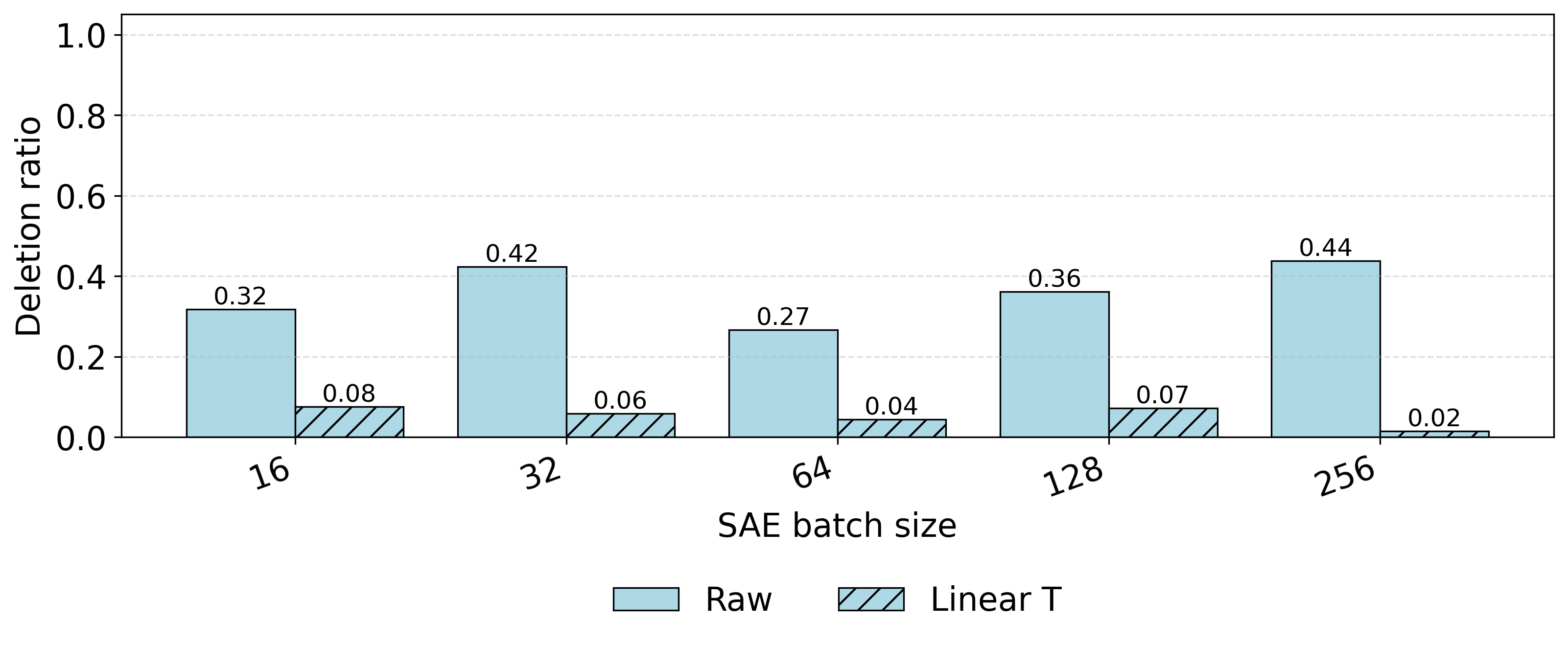}
        \label{fig:frec_del_batch}}

    \subfloat[Regained count and mass ratio]{
        \includegraphics[width=0.42\textwidth]{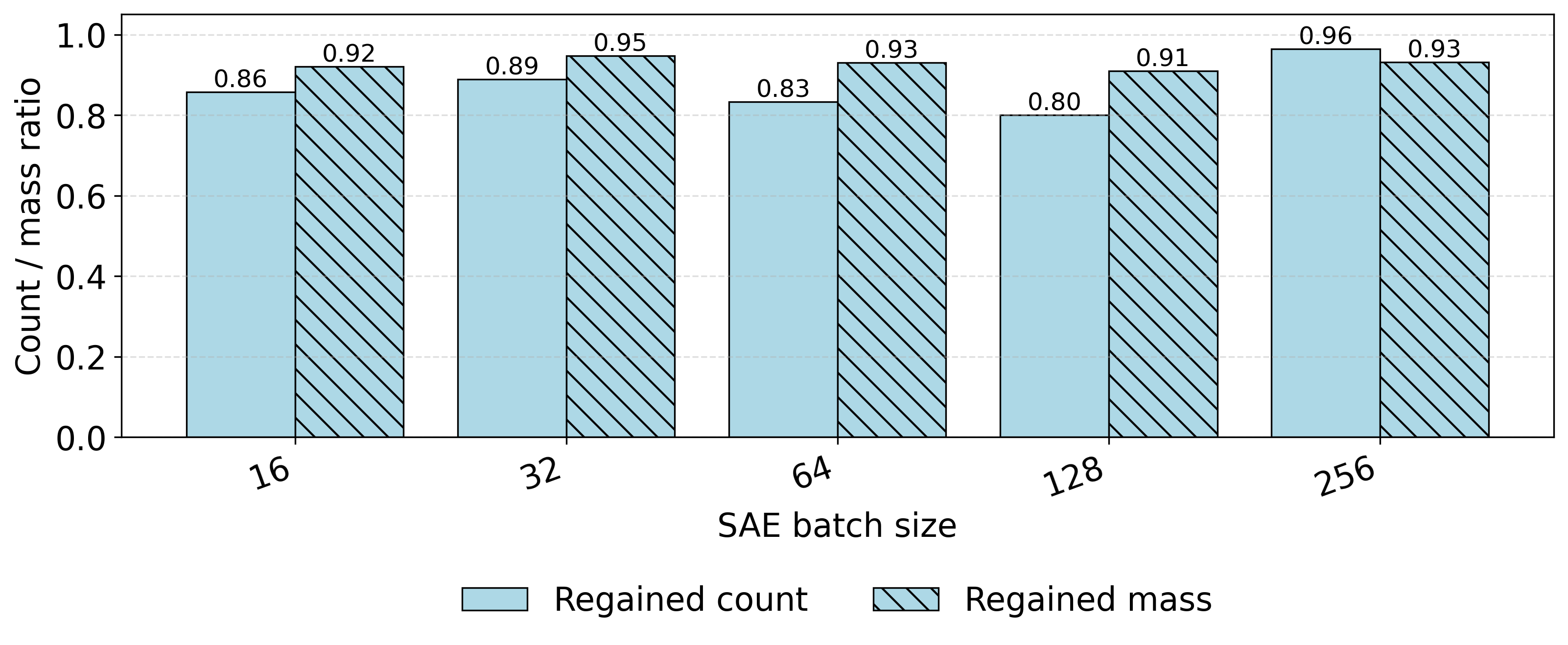}
        \label{fig:del_batch}}
        \hfill
    \subfloat[Representation-based probe accuracy]{
        \includegraphics[width=0.42\textwidth]{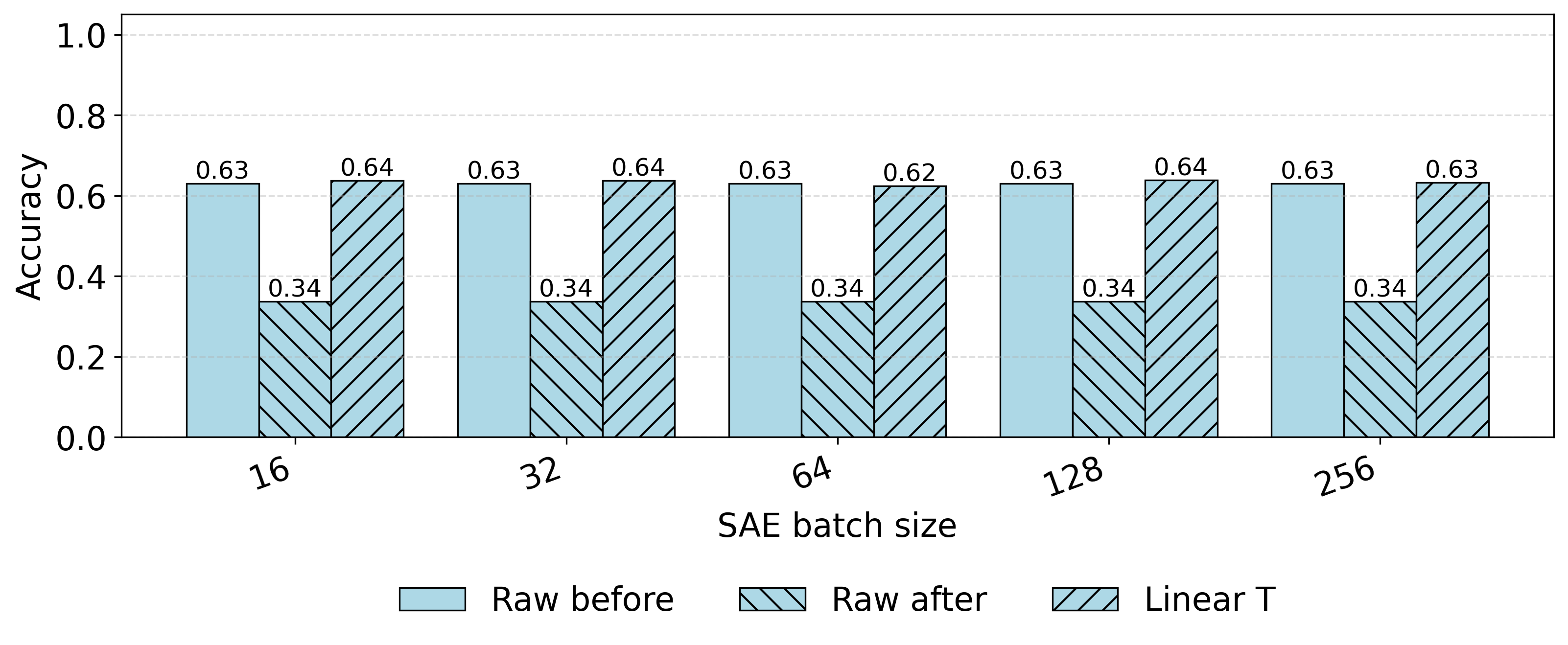}
        \label{fig:acc_batch}}
    
    \subfloat[Concept-based probe accuracy]{
        \includegraphics[width=0.42\textwidth]{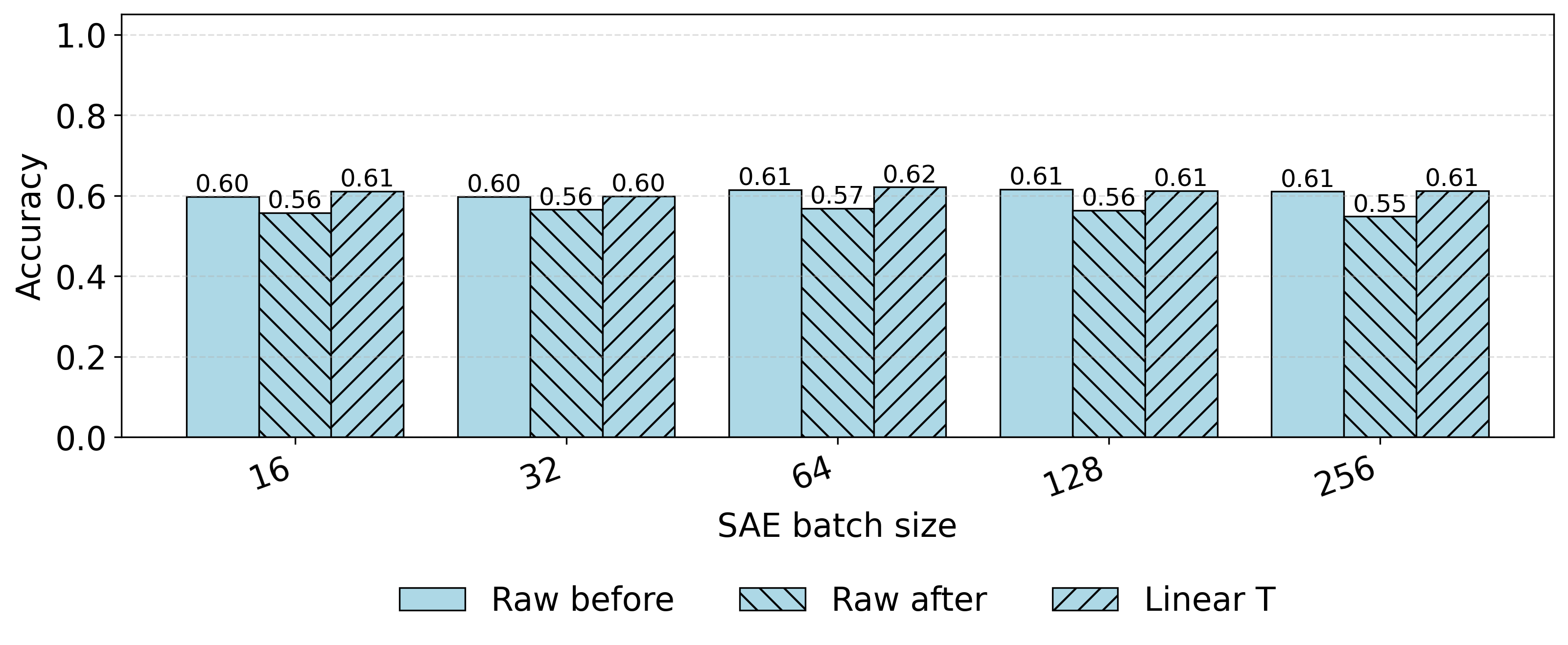}
        \label{fig:acc_sae_batch}}
        \hfill
    \subfloat[Mean balanced accuracy and F1]{
        \includegraphics[width=0.42\textwidth]{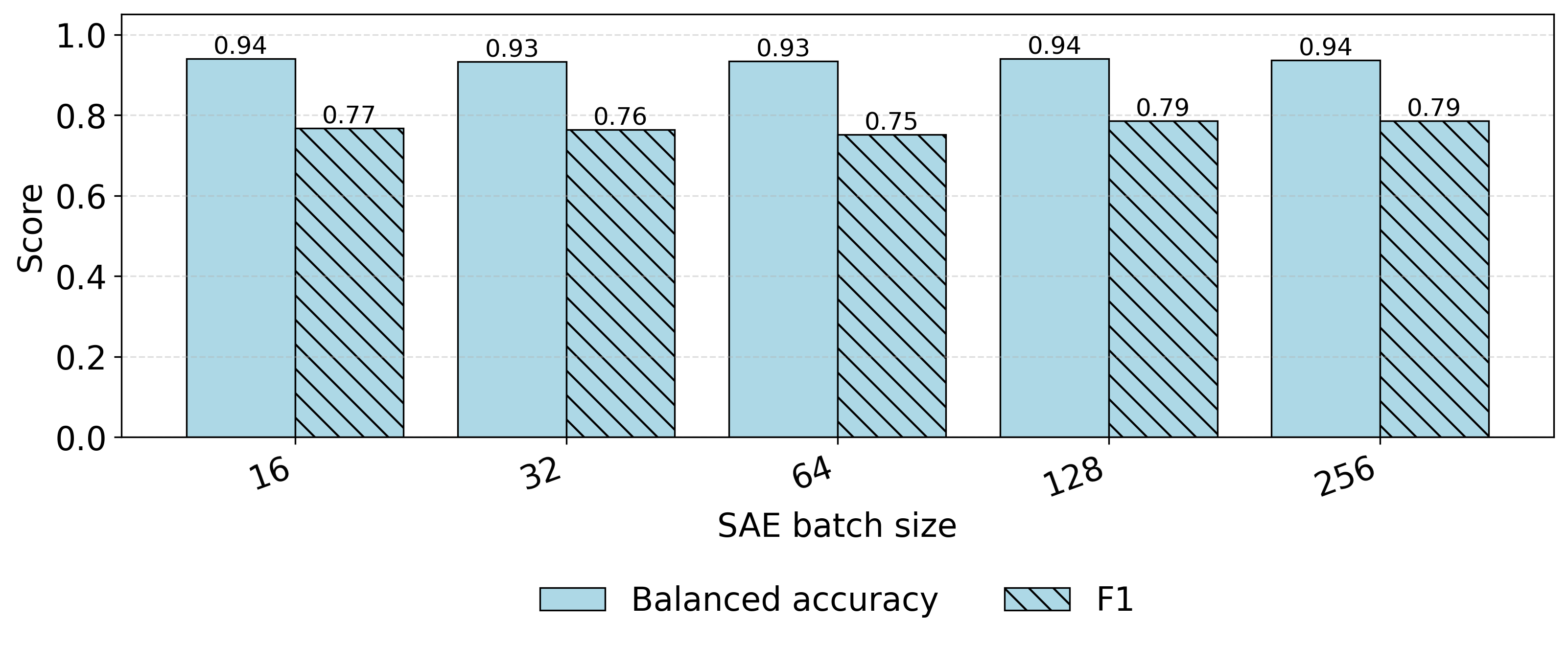}
        \label{fig:f1_batch}}
    \caption{The \textbf{impact of SAE batch size} on metrics for 2seq-tiny-ImageNet task 0 under LwF. }\label{fig:metrics_sae_batch}
\end{figure}

Fig. \ref{fig:metrics_sae_batch} shows the impact of the SAE training batch size on the key metrics for the example 2seq-tiny-ImageNet + LwF setup. Overall, the results are qualitatively stable across the tested batch sizes, indicating that our conclusions are not sensitive to this hyperparameter. Although small quantitative differences appear, the overall patterns remain unchanged: raw task-$t+1$ representations show concept deletion, linear alignment recovers a large fraction of the lost information, and both task- and concept-level probe performance stay close to the reference values. Similarly to the results for $K$, most values remain within the spread observed across repeated runs, suggesting that the effect of batch size is comparable to the run-to-run variability.

\begin{figure}[h]
    \centering
    \subfloat[Balanced accuracy]{
        \includegraphics[width=0.42\textwidth]{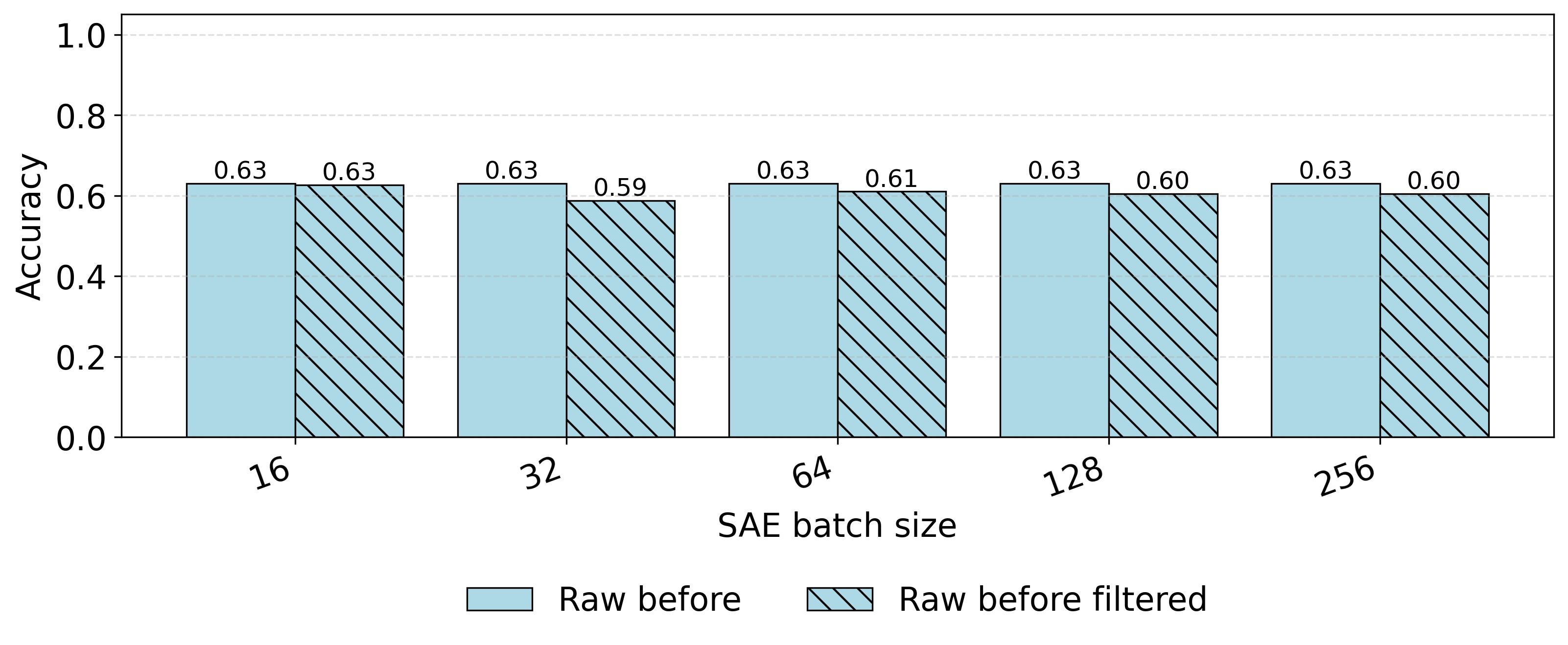}
        \label{fig:accuracy_sae_batch_raw_multiple}}
    \hfill
    \subfloat[F1 score]{
        \includegraphics[width=0.42\textwidth]{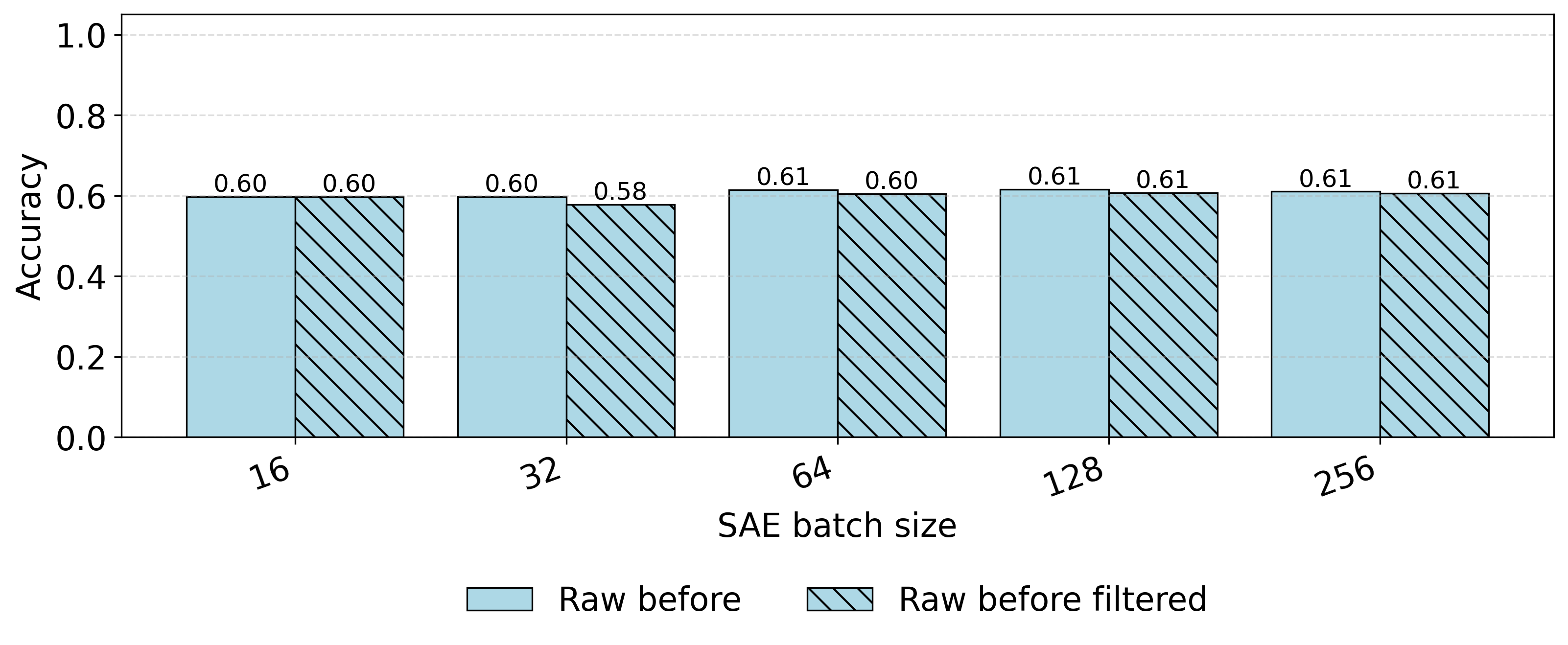}
        \label{fig:accuracy_sae_batch_sae_multiple}}
    \caption{The \textbf{impact of SAE batch size} on accuracy of probes built for prediction of 2seq-tiny-ImageNet task 0 classes under LwF based on the raw representations from the continual model (representation-level) and based on the SAE latent neuron activations (concept-level). All accuracy values are computed for the task 0 data after training on task 0, but for the "Raw before filtered" case, we use our SAE decoder to zero out the neurons that are not active binary under our binarization rule.}\label{fig:accuracy_sae_batch}
\end{figure}

To verify that the set of binary active neurons is functionally relevant under different SAE batch sizes, we additionally measure task-0 classification accuracy after zeroing out, through the SAE decoder, all latent neurons that are not marked as active under the given frequency-based rule. Figure~\ref{fig:accuracy_sae_batch} shows that the approach is robust to the chosen batch size for SAE training.

\section{Impact of using a non-linear translation model on concept decodability and accessibility}\label{app:nonlinear_translation}

\paragraph{Nonlinear recoverability setup.}
To test whether the information could be better restored by a low-capacity nonlinear mapping instead of a linear mapping, we additionally considered a nonlinear translation model. It was implemented as a \texttt{torch} model with three fully connected layers and GELU nonlinearities after the first two layers. The hidden dimensionality was set equal to the input dimensionality. The model was trained, analogously to the linear translator, to predict the task-$t$ representations from the task-$(t+s)$ representations by minimizing mean squared error. We trained it for $100$ epochs using AdamW, with learning rate $10^{-3}$, weight decay $10^{-4}$, batch size $128$, and a validation split of $0.2$.

\begin{figure}[h]
    \centering
    \subfloat[Active neurons]{
        \includegraphics[width=0.42\textwidth]{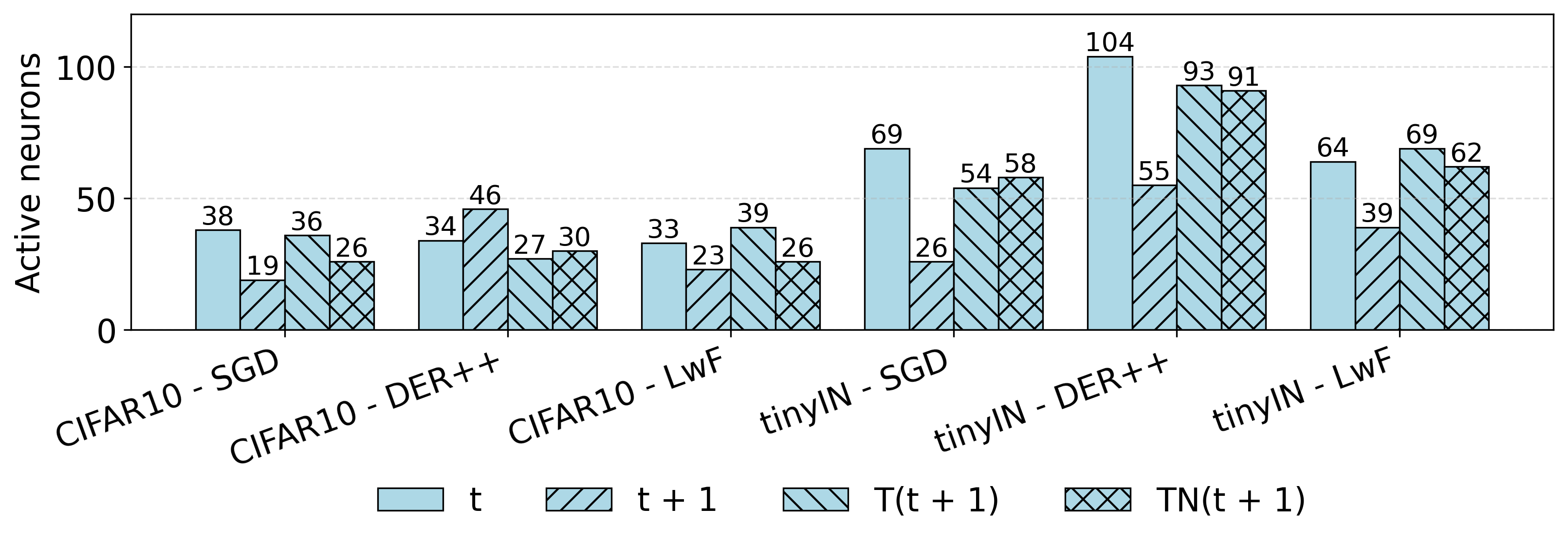}
        \label{fig:NONLINEAR_act}}
    \hfill
    \subfloat[Deletion ratio]{
        \includegraphics[width=0.42\textwidth]{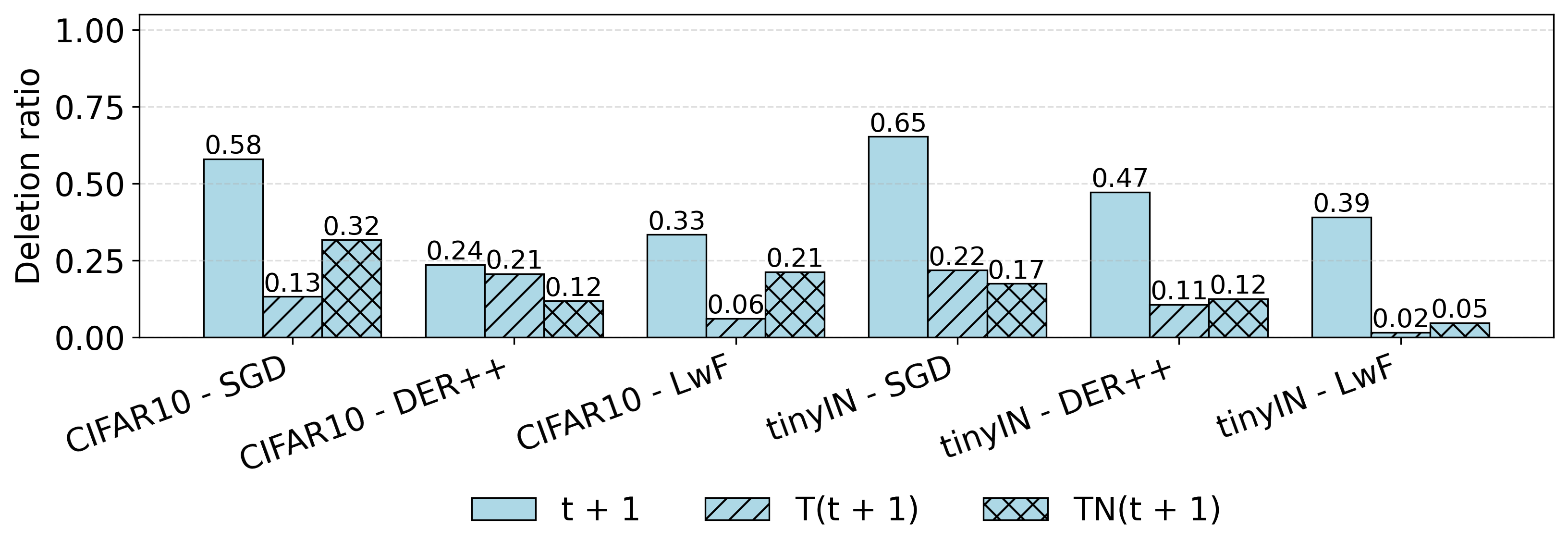}
        \label{fig:frec_NONLINEAR}}

    \subfloat[Regained count and mass ratio]{
        \includegraphics[width=0.42\textwidth]{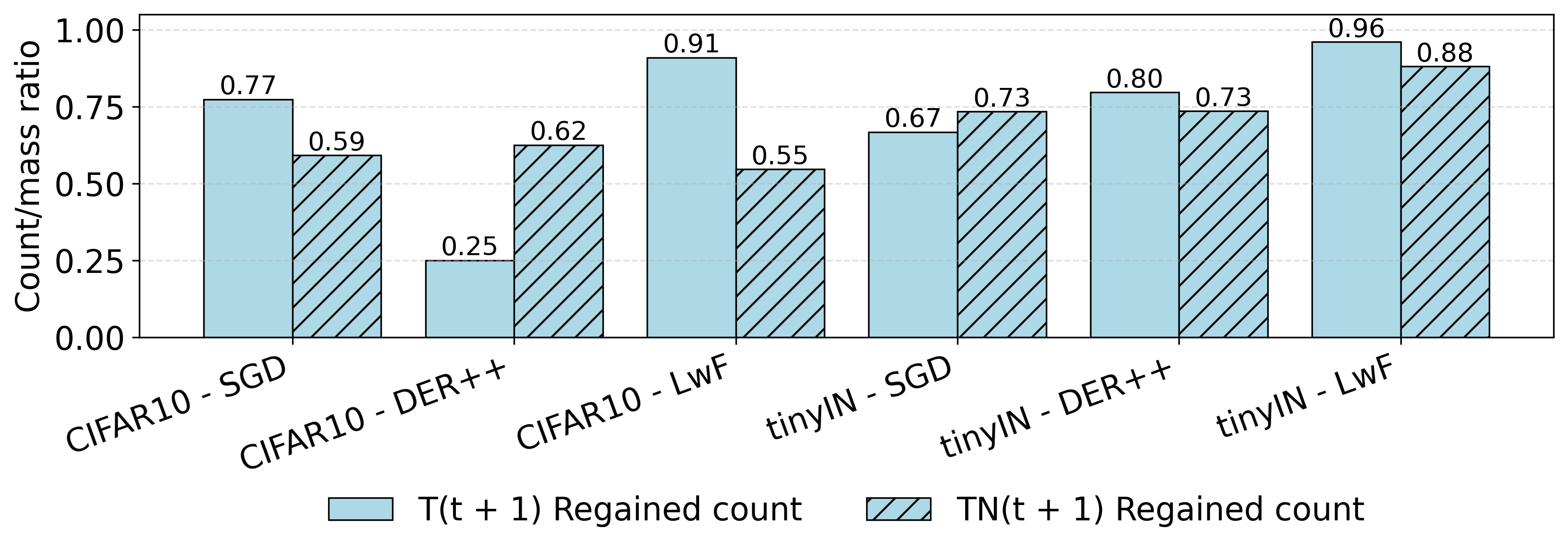}
        \label{fig:del_NONLINEAR}}
        \hfill
    \subfloat[Representation-based probe accuracy]{
        \includegraphics[width=0.42\textwidth]{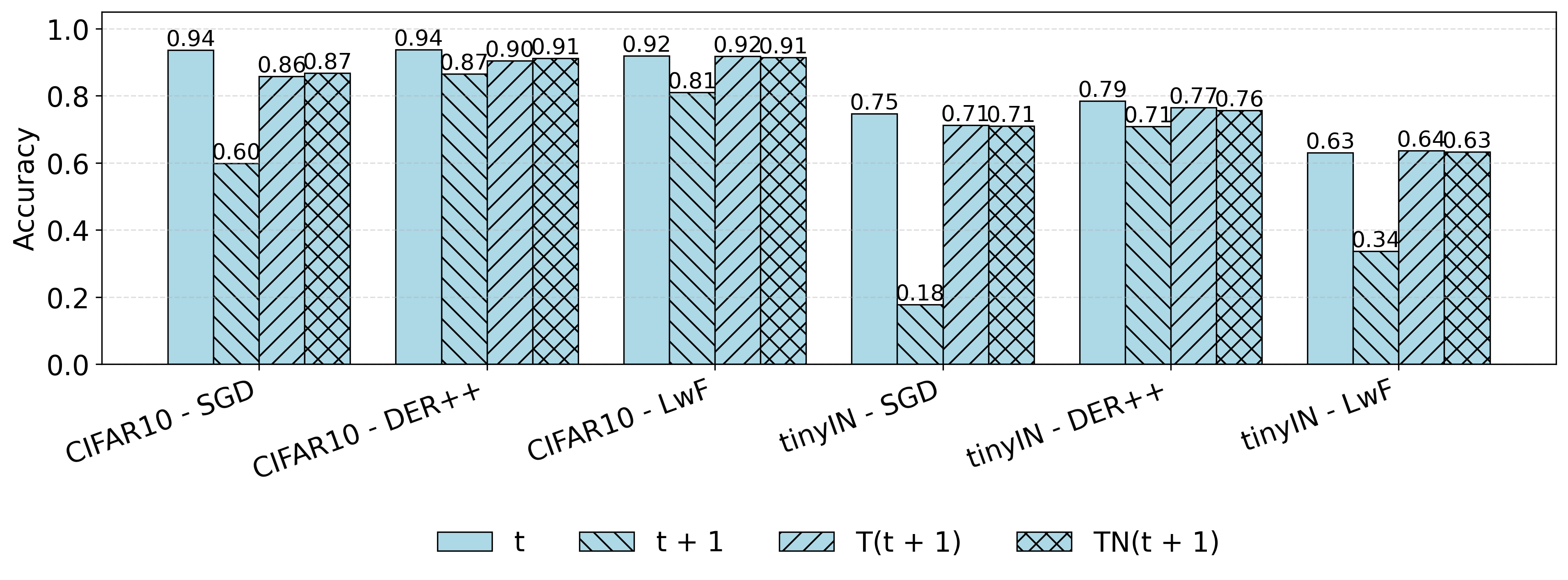}
        \label{fig:acc_NONLINEAR}}
    
    \subfloat[Concept-based probe accuracy]{
        \includegraphics[width=0.42\textwidth]{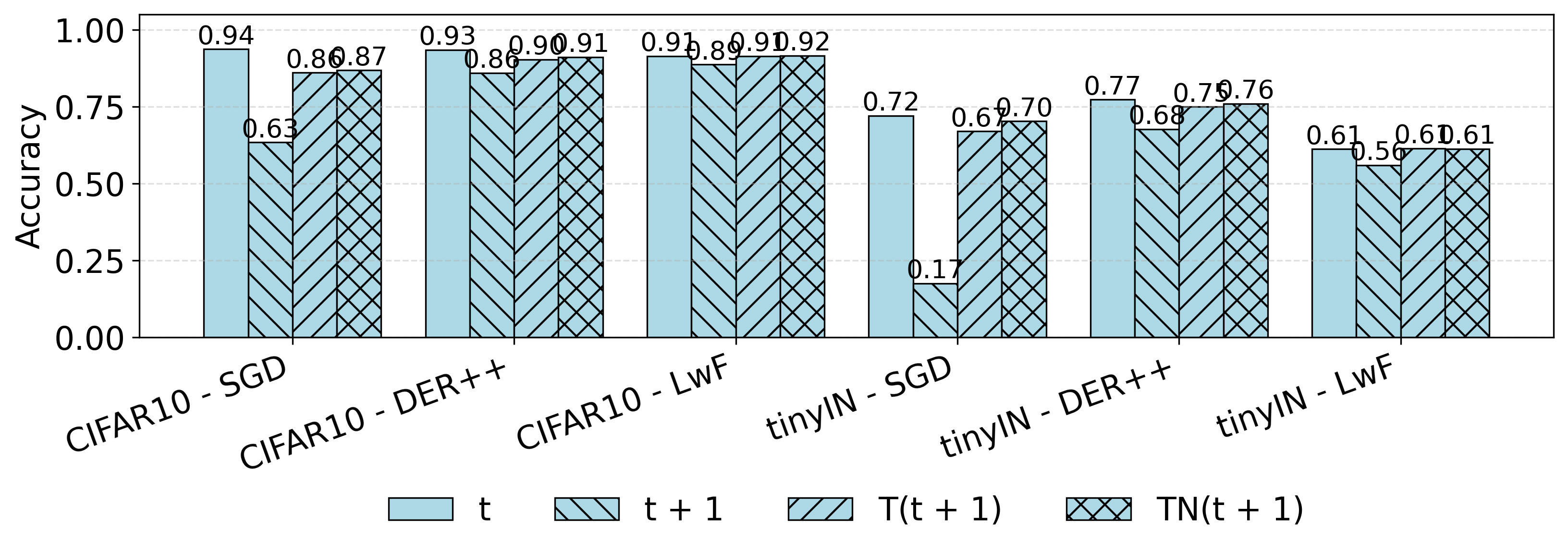}
        \label{fig:acc_sae_NONLINEAR}}
    \caption{The \textbf{impact of using a non-linear translation model} on metrics for 2seq-tiny-ImageNet compared to using linear translation. It is visible that the utilized model makes the recoverability and decodability worse in most cases, while in two improved cases, the improvement is marginal. }\label{fig:metrics_sae_NONLINEAR}
\end{figure}

Fig.~\ref{fig:metrics_sae_NONLINEAR} shows that the nonlinear translation model does not provide substantial improvements over the linear translation model. In most cases, the nonlinear model yields slightly worse performance across all considered metrics. Only two exceptions can be observed for deletion/regained count metrics, namely for 2seq-CIFAR10 with DER++ and 2seq-tiny-ImageNet with SGD, where the nonlinear model leads to minor improvements. However, these gains are marginal and correspond to restoring only a very small number of additional concepts (1-3 concepts), without meaningful impact on the overall metrics. Overall, both models achieve highly comparable performance, with the linear translation often performing slightly better. The overall results, therefore, indicate that the recoverable component of the representational drift is largely linear, both at the task level and at the level of individual concepts, and that simple nonlinear mappings do not provide additional explanatory power in this setting. While it is possible that more expressive nonlinear models could recover additional information, identifying such models and their configurations would require exploring a largely unconstrained architectural and hyperparameter space, which is beyond the scope of this work.

\section{Qualitative analysis of concepts}\label{app:concept_viz}

In this section, we provide a qualitative examination of SAE neurons across datasets and continual learning strategies. The goal of this analysis is to visually assess whether SAE latents after tasks $t$ and $t+1$ correspond to reasonably coherent and interpretable patterns. For data from task $t$, the visual coherence of the top-activating examples suggests that SAE latent neurons capture recurring patterns in the representation space and suggests that it may therefore serve as a useful proxy for concept detectors. Comparing task $t$ examples after training on task $t+1$ with those obtained immediately after task $t$ provides a qualitative indication of concept preservation. Visual similarity suggests that the concept remains at least partially encoded. Substantial differences, visible even in the most prototypical and central part of the neuron activation space, point either to information loss or to representational changes that prevent faithful recovery of the original concept. These results further support analysis of cases for which concept prediction $F1>0$ and can be used as a visual sanity check. In particular, we study the visual quality of neurons identified for task $t$ after $t$ and $t+1$ training, by inspecting their top-$9$ activating images, ordered by activation rank from $1$ to $9$. To make this comparison meaningful in the presence of representation drift, the maximally activating examples at $t+1$ are determined using the linearly translated feature vectors. This allows us to examine how the center of the concept changes after subsequent training, and whether the same visual motif can still be recovered despite the apparent shift in representation space.

\begin{figure*}[h]
    \centering

    \subfloat[After $t$ training]{
        \centering
        \par\medskip
        \foreach \row in {{01,02,03},{04,05,06},{07,08,09}}{%
            \foreach \img [count=\j from 1] in \row {%
                \includegraphics[width=0.11\linewidth]{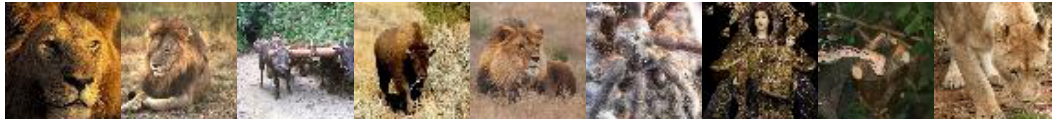}%
                \ifnum\j<3\hfill\fi
            }%
            \par\medskip
        }%
    }
    \hfill
     \subfloat[After $t+1$ training]{
        \centering
        \par\medskip
        \foreach \row in {{01,02,03},{04,05,06},{07,08,09}}{%
            \foreach \img [count=\j from 1] in \row {%
                \includegraphics[width=0.11\linewidth]{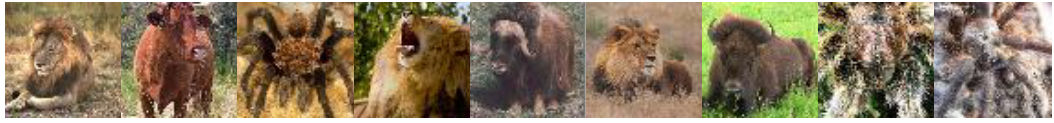}%
                \ifnum\j<3\hfill\fi
            }%
            \par\medskip
        }%
    }

    \caption{The example \textbf{retained} concept, which stayed active in raw vector after $t+1$.  Top-ranked examples for latent neuron 274 are presented at $t$ (top) and at $t+1$ (bottom). The setup was 2seq-tiny-ImageNet with DER++. The pictures show a \textbf{clear motif}: brown, furry objects, and the motif is shared between representations at $t$ and $t+1$.}
    \label{fig:neuron274_examples_retained}
\end{figure*}

\begin{figure*}[h]
    \centering

    \subfloat[After $t$ training]{
        \centering
        \par\medskip
        \foreach \row in {{01,02,03},{04,05,06},{07,08,09}}{%
            \foreach \img [count=\j from 1] in \row {%
                \includegraphics[width=0.11\linewidth]{FIGS/NEURON_IMAGES/370_active/before_rank_\img.png}%
                \ifnum\j<3\hfill\fi
            }%
            \par\medskip
        }%
    }
    \hfill
     \subfloat[After $t+1$ training]{
        \centering
        \par\medskip
        \foreach \row in {{01,02,03},{04,05,06},{07,08,09}}{%
            \foreach \img [count=\j from 1] in \row {%
                \includegraphics[width=0.11\linewidth]{FIGS/NEURON_IMAGES/370_active/after_rank_\img.png}%
                \ifnum\j<3\hfill\fi
            }%
            \par\medskip
        }%
    }

    \caption{The example \textbf{retained} concept, which stayed active in raw vector after $t+1$.  Top-ranked examples for latent neuron 370 are presented at $t$ (top) and at $t+1$ (bottom). The setup was 2seq-tiny-ImageNet with DER++. The pictures show a \textbf{clear motif}: penguins, and the motif is shared between representations at $t$ and $t+1$.}
    \label{fig:neuron370_examples_retained}
\end{figure*}

\begin{figure*}[h]
    \centering

    \subfloat[After $t$ training]{
        \centering
        \par\medskip
        \foreach \row in {{01,02,03},{04,05,06},{07,08,09}}{%
            \foreach \img [count=\j from 1] in \row {%
                \includegraphics[width=0.11\linewidth]{FIGS/NEURON_IMAGES/425_active/before_rank_\img.png}%
                \ifnum\j<3\hfill\fi
            }%
            \par\medskip
        }%
    }
    \hfill
     \subfloat[After $t+1$ training]{
        \centering
        \par\medskip
        \foreach \row in {{01,02,03},{04,05,06},{07,08,09}}{%
            \foreach \img [count=\j from 1] in \row {%
                \includegraphics[width=0.11\linewidth]{FIGS/NEURON_IMAGES/425_active/after_rank_\img.png}%
                \ifnum\j<3\hfill\fi
            }%
            \par\medskip
        }%
    }

    \caption{The example \textbf{retained} concept, which stayed active in raw vector after $t+1$.  Top-ranked examples for latent neuron 425 are presented at $t$ (top) and at $t+1$ (bottom). The setup was 2seq-tiny-ImageNet with DER++. The pictures show a \textbf{clear motif}: transport vehicles, and the motif is shared between representations at $t$ and $t+1$.}
    \label{fig:neuron425_examples_retained}
\end{figure*}

\begin{figure*}[h]
    \centering

    \subfloat[After $t$ training]{
        \centering
        \par\medskip
        \foreach \row in {{01,02,03},{04,05,06},{07,08,09}}{%
            \foreach \img [count=\j from 1] in \row {%
                \includegraphics[width=0.11\linewidth]{FIGS/NEURON_IMAGES/687_good_restoration/before_rank_\img.png}%
                \ifnum\j<3\hfill\fi
            }%
            \par\medskip
        }%
    }
    \hfill
     \subfloat[After $t+1$ training]{
        \centering
        \par\medskip
        \foreach \row in {{01,02,03},{04,05,06},{07,08,09}}{%
            \foreach \img [count=\j from 1] in \row {%
                \includegraphics[width=0.11\linewidth]{FIGS/NEURON_IMAGES/687_good_restoration/after_rank_\img.png}%
                \ifnum\j<3\hfill\fi
            }%
            \par\medskip
        }%
    }

    \caption{The example of a \textbf{lower quality concept}. The concept can be described as \textbf{seemingly deleted}, as it was absent in raw $t+1$ and \textbf{recovered}, because it reappeared in translated $t+1$. Top-ranked examples for latent neuron 687 are presented at $t$ (top) and at $t+1$ (bottom). The setup was 2seq-tiny-ImageNet with LwF. The concept obtained a relatively \textbf{high decodability level}: mean bacc was $0.97$ and F1 was $0.8$ for concept decodability. It had $76$ images with $>0$ activations at $t$ and $77$ at translated $t+1$.}
    \label{fig:neuron687_examples_deleted_recovered}
\end{figure*}

\begin{figure*}[h]
    \centering

    \subfloat[After $t$ training]{
        \centering
        \par\medskip
        \foreach \row in {{01,02,03},{04,05,06},{07,08,09}}{%
            \foreach \img [count=\j from 1] in \row {%
                \includegraphics[width=0.11\linewidth]{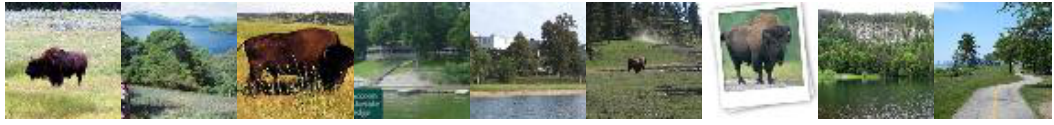}%
                \ifnum\j<3\hfill\fi
            }%
            \par\medskip
        }%
    }
    \hfill
     \subfloat[After $t+1$ training]{
        \centering
        \par\medskip
        \foreach \row in {{01,02,03},{04,05,06},{07,08,09}}{%
            \foreach \img [count=\j from 1] in \row {%
                \includegraphics[width=0.11\linewidth]{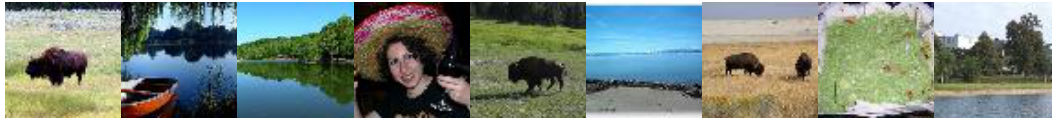}%
                \ifnum\j<3\hfill\fi
            }%
            \par\medskip
        }%
    }

    \caption{The example of a \textbf{lower quality concept}. The concept can be described as \textbf{seemingly deleted}, as it was absent in raw $t+1$ and \textbf{recovered}, because it reappeared in translated $t+1$. Top-ranked examples for latent neuron 358 are presented at $t$ (top) and at $t+1$ (bottom). The setup was 2seq-tiny-ImageNet with DER++. The concept obtained a \textbf{medium decodability level}: F1 ($0.6$) at mean bacc of $0.97$ for concept decodability.}
    \label{fig:neuron358_examples_deleted_recovered}
\end{figure*}

\begin{figure*}[h]
    \centering

    \subfloat[After $t$ training]{
        \centering
        \par\medskip
        \foreach \row in {{01,02,03},{04,05,06},{07,08,09}}{%
            \foreach \img [count=\j from 1] in \row {%
                \includegraphics[width=0.11\linewidth]{FIGS/NEURON_IMAGES/146_deleted/before_rank_\img.png}%
                \ifnum\j<3\hfill\fi
            }%
            \par\medskip
        }%
    }
    \hfill
     \subfloat[After $t+1$ training]{
        \centering
        \par\medskip
        \foreach \row in {{01,02,03},{04,05,06},{07,08,09}}{%
            \foreach \img [count=\j from 1] in \row {%
                \includegraphics[width=0.11\linewidth]{FIGS/NEURON_IMAGES/146_deleted/after_rank_\img.png}%
                \ifnum\j<3\hfill\fi
            }%
            \par\medskip
        }%
    }

    \caption{The example of a \textbf{lower quality concept}. The concept can be described as \textbf{seemingly deleted}, as it was absent in raw $t+1$ and \textbf{recovered}, because it reappeared in translated $t+1$. Top-ranked examples for latent neuron 146 are presented at $t$ (top) and at $t+1$ (bottom). The setup was 2seq-tiny-ImageNet with LwF. The concept obtained a relatively \textbf{high decodability level}: F1 of $0.71$ and mean bacc of $0.92$ for concept decodability.}
    \label{fig:neuron146_examples_deleted_recovered}
\end{figure*}

\begin{figure*}[h]
    \centering

    \subfloat[After $t$ training]{
        \centering
        \par\medskip
        \foreach \row in {{01,02,03},{04,05,06},{07,08,09}}{%
            \foreach \img [count=\j from 1] in \row {%
                \includegraphics[width=0.11\linewidth]{FIGS/NEURON_IMAGES/cifaar_14_deleted/before_rank_\img.png}%
                \ifnum\j<3\hfill\fi
            }%
            \par\medskip
        }%
    }
    \hfill
     \subfloat[After $t+1$ training]{
        \centering
        \par\medskip
        \foreach \row in {{01,02,03},{04,05,06},{07,08,09}}{%
            \foreach \img [count=\j from 1] in \row {%
                \includegraphics[width=0.11\linewidth]{FIGS/NEURON_IMAGES/cifaar_14_deleted/after_rank_\img.png}%
                \ifnum\j<3\hfill\fi
            }%
            \par\medskip
        }%
    }

    \caption{The example of a \textbf{good quality concept}. The concept can be described as \textbf{seemingly deleted}, as it was absent in raw $t+1$ and \textbf{decodable}, because it didn't reappear in translated $t+1$ but obtained concept decodability $F1 > 0$. Top-ranked examples for latent neuron 14 are presented at $t$ (top) and at $t+1$ (bottom). The setup was 2seq-CIFAR10 with SGD. The concept obtained a relatively \textbf{low concept decodability level}: F1 of $0.44$ and mean bacc of $0.88$.}
    \label{fig:neuron_14_cifar_examples_decodable}
\end{figure*}

In the qualitative analysis, we present representative examples of concepts drawn from the proposed taxonomy, illustrating different post-task outcomes such as \textbf{retained} (Figs. \ref{fig:neuron274_examples_retained}, \ref{fig:neuron370_examples_retained}, \ref{fig:neuron425_examples_retained}), \textbf{deleted} (Figs. \ref{fig:neuron687_examples_deleted_recovered}, \ref{fig:neuron146_examples_deleted_recovered}, \ref{fig:neuron_14_cifar_examples_decodable}, \ref{fig:neuron_677_cifar_examples_deleted_decodable}), \textbf{recovered} (Figs. \ref{fig:neuron687_examples_deleted_recovered}, \ref{fig:neuron146_examples_deleted_recovered}, \ref{fig:neuron358_examples_deleted_recovered}), and \textbf{decodable} (Figs. \ref{fig:neuron_14_cifar_examples_decodable}, \ref{fig:neuron_677_cifar_examples_deleted_decodable}) concepts. These examples complement the quantitative results by showing how the central visual structure of individual concepts expressed with top-9 activating images changes after subsequent task learning. Notably, across the cases examined in our analysis, we did not observe any concepts that would fall into the \textbf{Lost} category at the strictest possible F1 threshold, i.e., concepts that were neither recoverable through translation nor decodable from the representation in our setup.

\begin{figure*}[h]
    \centering

    \subfloat[After $t$ training]{
        \centering
        \par\medskip
        \foreach \row in {{01,02,03},{04,05,06},{07,08,09}}{%
            \foreach \img [count=\j from 1] in \row {%
                \includegraphics[width=0.11\linewidth]{FIGS/NEURON_IMAGES/cifar_677_deleted/before_rank_\img.png}%
                \ifnum\j<3\hfill\fi
            }%
            \par\medskip
        }%
    }
    \hfill
     \subfloat[After $t+1$ training]{
        \centering
        \par\medskip
        \foreach \row in {{01,02,03},{04,05,06},{07,08,09}}{%
            \foreach \img [count=\j from 1] in \row {%
                \includegraphics[width=0.11\linewidth]{FIGS/NEURON_IMAGES/cifar_677_deleted/after_rank_\img.png}%
                \ifnum\j<3\hfill\fi
            }%
            \par\medskip
        }%
    }

    \caption{The example of a \textbf{lower quality concept}. The concept can be described as \textbf{seemingly deleted}, as it was absent in raw $t+1$ and \textbf{decodable}, because it didn't reappear in translated $t+1$ but obtained concept decodability $F1 > 0$. Top-ranked examples for latent neuron 677 are presented at $t$ (top) and at $t+1$ (bottom). The setup was 2seq-CIFAR10 with SGD. The concept obtained a relatively \textbf{low decodability level}: F1 of $0.28$ and mean bacc of $0.75$.}
    \label{fig:neuron_677_cifar_examples_deleted_decodable}
\end{figure*}

In many of presented cases images representing the concepts at $t$ show a clear common motif, e.g. brown, furry objects (Fig. \ref{fig:neuron274_examples_retained}), penguins (Fig. \ref{fig:neuron370_examples_retained}), bisons (Fig. \ref{fig:neuron687_examples_deleted_recovered}) or birds (Fig. \ref{fig:neuron_14_cifar_examples_decodable}), regardless of the $t+1$ outcome of the concept transition, which was \textbf{retained} for the first two examples, \textbf{recovered} for the third example and \textbf{decodable} for the fourth example. Some lower-quality neurons can be showed as well, e.g. see Figs. \ref{fig:neuron146_examples_deleted_recovered} and \ref{fig:neuron_677_cifar_examples_deleted_decodable}. Nevertheless, even in these cases, some recurring objects or motifs can still be identified. For example, in Fig.~\ref{fig:neuron146_examples_deleted_recovered}, several highly activating examples contain backpacks or towers, while in Fig.~\ref{fig:neuron_677_cifar_examples_deleted_decodable}, recurring elements such as cats and cars can be observed. This behavior is consistent with standard SAEs, which typically extract features of varying quality and degrees of disentanglement. Overall, the presented examples still suggest that the SAE and our mean-based and frequency-based thresholding criteria capture at least partially disentangled features associated with recurring visual patterns, supporting our use of these latent units as proxies for concepts.

Across all presented examples, the top-$9$ activating images at task $t$ and after learning task $t+1$ remain highly aligned, indicating that the centers of the corresponding concepts' (latent features) activation spaces are largely preserved. This is reflected in shared visual motifs and, in many cases, even repeated images among the strongest activations. Importantly,  this observation is valid even for the seemingly deleted, recovered and decodable categories. Fig.~\ref{fig:neuron687_examples_deleted_recovered} shows that some contamination of the task $t$ concept center can occur at $t+1$. While at task $t$ the dominant motifs include green plants, water, and bisons on grass, at task $t+1$ an out-of-pattern image of a girl in a sombrero appears among the top activations. These qualitative observations provide graphical support for the quantitative results in the paper, which indicate that a large portion of concept-level knowledge is preserved during updates on subsequent tasks, although some information may become harder to read out or partially lost.

\section{Hyperparameters} \label{app:hyperparameters}
In our \textbf{concept-level analysis setup}, 
for each task, we train an SAE with \texttt{BatchTopKSAE} regime via the \texttt{overcomplete} \footnote{\url{https://github.com/KempnerInstitute/overcomplete}} on the same Mammoth's train/test splits. We use the unified setup: batch size $B=16$, top-$K=10$, $10$ epochs, learning rate $5 \cdot 10^{-3}$, expansion rate $2$, Mean Squared Error (MSE) loss augmented with an \texttt{overcomplete}'s dead-neuron reactivation loss weighted by $10^{-2}$. Across experiments, manual inspection showed reconstruction $R^2 > 0.6$ and close to 0\% dead-neuron rate, indicating meaningful and non-degenerate SAEs. As a qualitative complement, we examine top 9 activating images for selected SAE latents and find consistent shared visual motifs, indicating that they capture coherent patterns rather than arbitrary feature mixtures (Appendix \ref{app:concept_viz}). 
We use frequency-based binarization threshold $\tau=0.05$. We use a single linear layer as for translation, trained for $100$ epochs with \texttt{AdamW} using MSE, learning rate $10^{-3}$, weight decay $10^{-4}$, batch size $128$, and a validation split $0.2$. For task-level probes, we use Logistic Regression with \texttt{lbfgs} solver and 1500 maximum iterations, while for concept-level prediction we use Logistic Regression with \texttt{liblinear} solver, 1000 iterations, and \texttt{balanced} class weight. We verified the analysis robustness across 10 runs (App.~\ref{app:stability_across_runs}), frequency-based binarization thresholds (App.~\ref{app:stability_binarization}, \{0.00625, 0.0125, 0.025, 0.05, 0.1, 0.2\}), SAE batch size (App.~\ref{app:stability_sae_batch}, \(B \in \{16,32,64,128,256\}\)) and SAE $K$ (App.~\ref{app:stability_sae_K}, \(K \in \{10,16,32,64\}\)). The results show consistent behavior of key metrics, with variability largely within run-to-run fluctuations, and support robustness of our conclusions. We further validate that the selected active neurons retain most task-relevant information with a SAE-based experiments on deletion of non-active neurons under different values of parameters and under our binarization rule for all methods and datasets, further supporting the use of threshold \(0.05\). We used multiple NVIDIA GH200 GPUs for continual learning experiments and 1 NVIDIA Titan RTX GPU for SAE training and evaluation. Average execution time for an example task, including saving all the required and additional data, was $204.199\pm42.473$ s on the Titan RTX GPU.

\section{Monosemanticity of SAE concept proxies}\label{app:ms_analysis}

As an additional sanity check of the quality of the SAE latent features used as concept proxies in our framework, we evaluate their monosemanticity using the Monosemanticity Score (MS) introduced in work~\cite{pach2025sparse}. The aim of this experiment is not to claim that SAE latents correspond to ground-truth semantic concepts, but to quantitatively assess whether the selected latent features capture more coherent visual patterns than random associations.

MS measures how coherent are the images that strongly activate a given SAE latent neuron. Pairwise cosine similarities are computed between embeddings from a pretrained encoder. For a neuron $k$, representing a given concept proxy, latent activations are used to weight similarities between samples. The MS is then defined as the activation-weighted average of pairwise embedding similarities, assigning higher scores to neurons whose most ctivating examples correspond to coherent visual patterns, and lower scores to neurons activating for diverse unrelated images~\cite{pach2025sparse}.

We perform the analysis for an example configuration: 2seq-tiny-ImageNet under the LwF continual learning strategy. For each active SAE latent neuron identified under our frequency-based binarization criterion, we compute the MS using embeddings extracted from the continual model backbone used as an encoder. As a baseline, we additionally compute the same score after randomly permuting the embeddings from the encoder, which destroys the sample similarity structure while preserving the structure of the encoder embedding space. This lets us assess whether latent neurons capture more monosemantic patterns than expected by chance, disentangling embedding-space geometry from neuron-induced meaningful similarity.

\begin{figure}[h]
    \centering
    \includegraphics[width=0.45\textwidth]{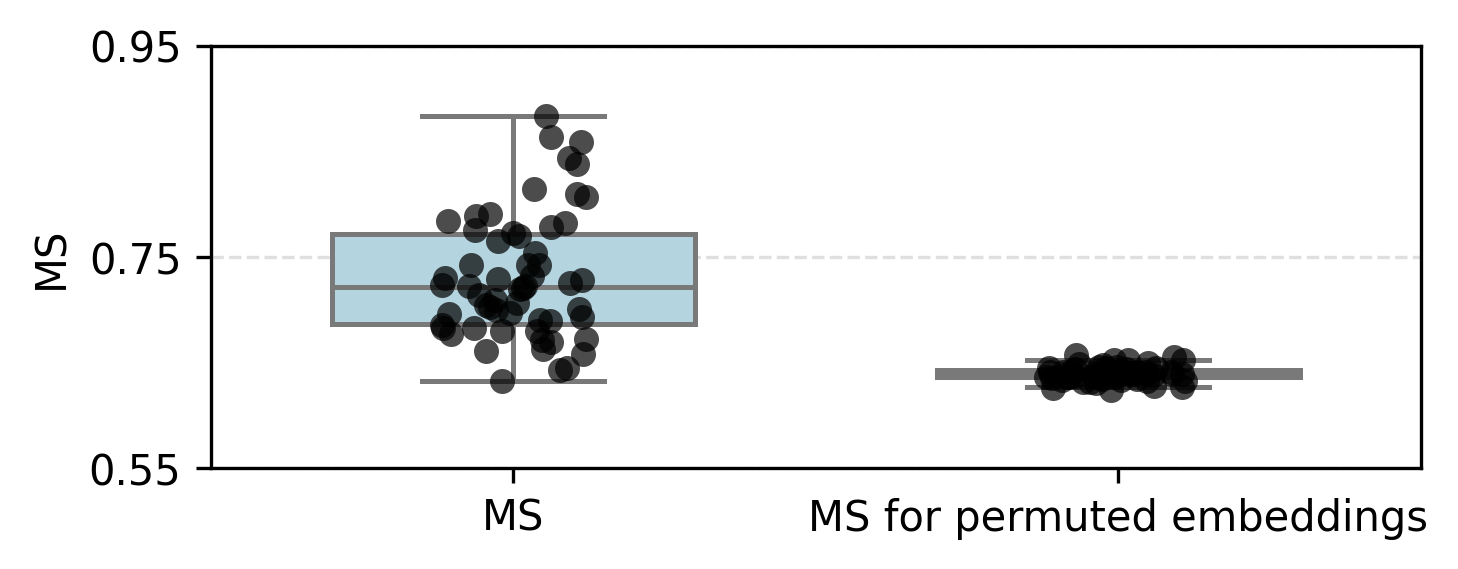}
    \label{fig:ms}
    \caption{Monosemanticity Score (MS) analysis for active SAE latent neurons in the example 2seq-tiny-ImageNet + LwF configuration. The box plots compare MS values of active SAE latent neurons with those obtained by permuting embeddings. Active SAE latent neurons achieve substantially higher MS values, indicating that they capture more coherent visual patterns than expected from random associations within the embedding space.}\label{fig:ms_Score}
\end{figure}
\FloatBarrier

Fig.~\ref{fig:ms_Score} shows that the SAE latents active under our binarization rule achieve substantially higher MS values than the permutation baseline. Although the scores vary across neurons, indicating different degrees of concept quality and disentanglement, which is expected from SAE latents, the majority of features are significantly more coherent than the baseline. In contrast, the baseline produces consistently lower and stable scores, as expected for randomly associated examples. The presented quantitative results support that the analyzed SAE latent features capture recurring and coherent visual patterns rather than arbitrary mixtures of activations, and thus can serve as concept proxies.

\raggedbottom

\FloatBarrier



\end{document}